\documentclass[runningheads]{llncs}

\usepackage{eccv}

\usepackage{eccvabbrv}

\usepackage{graphicx}
\usepackage{booktabs}

\usepackage[accsupp]{axessibility}  

\usepackage{hyperref}
\hypersetup{
    colorlinks=true,
    linkcolor=RoyalBlue,
    filecolor=RoyalBlue,
    urlcolor=RoyalBlue,
    citecolor=RoyalBlue,
}

\usepackage{orcidlink}

\usepackage{wrapfig}
\usepackage{pifont}
\usepackage{graphicx}
\usepackage{tabularx}
\usepackage{multirow}
\usepackage{booktabs}
\usepackage{amsmath}  
\usepackage{amssymb}  
\usepackage{mathrsfs} 
\usepackage[table,dvipsnames]{xcolor}
\usepackage{array}
\usepackage{float}
\usepackage{pifont} 
\usepackage{bbding} 
\usepackage{fontawesome} 
\usepackage[most]{tcolorbox}
\usepackage{listings}
\usepackage{graphicx}
\usepackage{etoolbox}
\usepackage{subcaption}

\def\ie{\emph{i.e.}}
\def\eg{\emph{e.g.}}

\newcommand{\mname}{\textsl{RoboTracer}}
\newcommand{\dname}{\textsl{TraceSpatial}}
\newcommand{\bname}{\textsl{TraceSpatial-Bench}}
\definecolor{ourpurple}{HTML}{7030A0} 
\definecolor{ouryellow}{HTML}{F4B402} 

\newcommand{\highlight}[1]{\textbf{\color{orange}#1}}

\definecolor{codeblue}{rgb}{0.25,0.5,0.5}
\definecolor{mylightblue}{rgb}{0.96, 0.995, 1}
\definecolor{myblue}{rgb}{0.88,0.98,1}
\definecolor{mygreen}{rgb}{0.92, 1.0, 0.92}
\definecolor{myred}{rgb}{1, 0.9, 0.9}
\definecolor{mygray}{gray}{0.95}
\definecolor{mydarkblue}{rgb}{0,0.08,1}
\definecolor{mydarkred}{rgb}{0.8,0.02,0.02}
\definecolor{mydarkorange}{rgb}{0.40,0.2,0.02}
\definecolor{mypurple}{RGB}{239,229,253}
\definecolor{mygold}{rgb}{0.75,0.6,0.12}
\definecolor{mydarkgray}{rgb}{0.66, 0.66, 0.66}
\definecolor{mydarkgreen}{rgb}{0.02,0.6,0.02}
\definecolor{mygray}{gray}{0.9}
\definecolor{tablegray}{gray}{0.95}

\begin{document}

\title{Towards Spatial Trace with Reasoning
in Vision-Language Models for Robotics} 

\titlerunning{{\mname}}

\author{
Enshen Zhou$^{1, 3\ast\ddagger}$,
Yibo Li$^{1,3\ast}$,
Jingkun An$^{1\ast}$,
Jiayuan Zhang$^{1\ast}$, \\
Shanyu Rong$^{2,3}$,
Mengzhen Liu$^{2,3}$,
Yi Han$^{1,3}$,
Yuheng Ji$^{3,5}$,\
Huajie Tan$^{2,3}$,
Jiawei He$^{3}$,
Pengwei Wang$^{3}$,
Zhongyuan Wang$^{3}$, 
Cheng Chi$^{3\dagger}$,\\
Lu Sheng$^{1, 4\dagger}$,
Shanghang Zhang$^{2,3\dagger}$
}
\institute{
$^{1}$School of Software, Beihang University, $^{2}$ State Key Laboratory of Multimedia Information Processing, School of Computer Science, Peking University, \\
$^{3}$ Beijing Academy of Artificial Intelligence, $^{4}$ Beijing Key Laboratory of Intelligent Creative Content Generation and Immersive Experience, $^{5}$ CASIA\\
\email{zhouenshen@buaa.edu.cn}, ~~\email{leeibo@buaa.edu.cn}, ~~\email{anjingkun02@gmail.com}\\
\email{chicheng@baai.ac.cn}, ~~\email{lsheng@buaa.edu.cn}, ~~\email{shanghang@pku.edu.cn}\\
}

\authorrunning{E.~Zhou et al.}


\maketitle

\let\thefootnote\relax\footnotetext{$^*$ Equal contribution\hspace{3pt} \hspace{5pt}$^\dagger$ Corresponding author\hspace{5pt} $^\ddagger$ Project leader
}

\begin{abstract}

%
%
%

Spatial tracing, as a fundamental embodied interaction ability for robots, is inherently challenging as it requires multi-step metric-grounded reasoning compounded with complex spatial referring and real-world metric measurement. 
However, existing methods struggle with this compositional task.
To this end, we propose RoboTracer, a 3D-aware VLM that first achieves both 3D spatial referring and measuring via a universal spatial encoder and a regression-supervised decoder to enhance scale awareness during supervised fine-tuning (SFT).
Moreover, RoboTracer advances multi-step metric-grounded reasoning via reinforcement fine-tuning (RFT) with metric-sensitive process rewards, supervising key intermediate perceptual cues to accurately generate spatial traces.
To support SFT and RFT training, we introduce TraceSpatial, a large-scale dataset of 30M QA pairs, spanning outdoor/indoor/tabletop scenes and supporting complex reasoning processes (up to 9 steps).
We further present TraceSpatial-Bench, a challenging benchmark filling the gap to evaluate spatial tracing.
Experimental results show that RoboTracer surpasses baselines in spatial understanding, measuring, and referring, with an average success rate of {79.1\%}, and also achieves SOTA performance on TraceSpatial-Bench by a large margin, exceeding Gemini-2.5-Pro by {36\%} accuracy. 
Notably, RoboTracer can be integrated with various control policies to execute long-horizon, dynamic tasks across diverse robots (UR5, G1 humanoid) in cluttered real-world scenes.
Please see the project page at \href{https://zhoues.github.io/RoboTracer/}{https://zhoues.github.io/RoboTracer}.

\keywords{Spatial tracing \and Spatial Reasoning \and Vision-Language Model}

\end{abstract}

\begin{figure}
    \centering
    \includegraphics[width=1\linewidth]{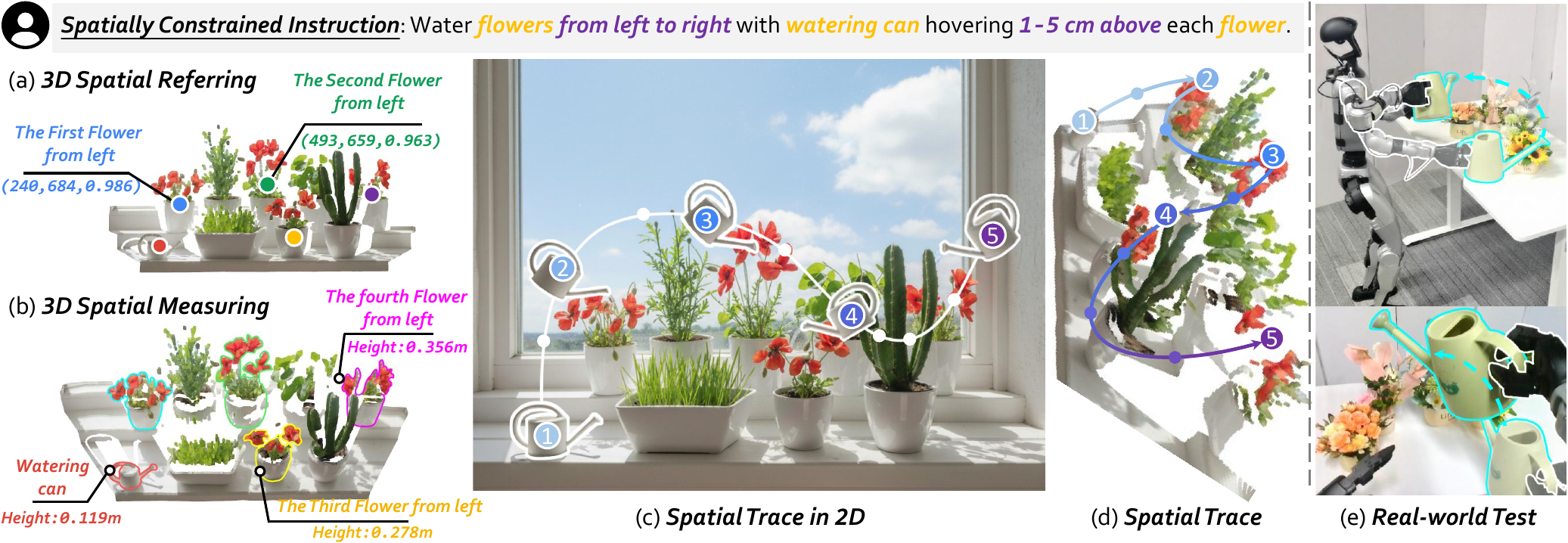}
    \vspace{-5mm}
    \captionof{figure}{    
    %
    Spatial tracing is pivotal for embodied robots to translate the spatially constrained instructions into 3D positional sequences (\ie, spatial traces) in complex 3D scenes.
    This task demands \textbf{(a)} 3D spatial referring to resolve spatial relations and locate relevant objects involved in the trace, and \textbf{(b)} 3D spatial measuring to understand absolute, real-world metric quantities related to the trace.
    For example, \textbf{(a)} 3D positions of the watering can, and each flower pot are localized from left to right, and \textbf{(b)} their corresponding heights in meters are measured.
    By performing multi-step, metric-grounded reasoning over the key information above, the generated spatial trace can support not only \textbf{(c)} multi-step manipulation, but also \textbf{(d)} collision-free motion, thereby \textbf{(e)} enabling efficient control of diverse robots across tasks in cluttered scenes.}
   \label{fig: motivation}
\end{figure}

\section{Introduction}
\label{sec: intro}

Embodied robots usually have to execute actions based on increasingly complex, spatially constrained instructions~\cite{zhou2025roborefer, team2025robobrain, team2025gemini, abdolmaleki2025gemini}, such as ``\textit{Water flowers from left to right with watering can hovering 1–5 cm above each one}'' in \cref{fig: motivation}, where recent data-scarce Vision-Language-Action (VLA) models fail to master.
%
%
In this case, it would be beneficial to generate a 3D positional sequence, named as \textit{spatial trace}, as an intuitive bridge to interpret the instruction following procedure in 3D space and guide the generation of actual action trajectories for robots. 
%
%
However, this surrogate task (\ie, \textit{spatial tracing}) is inherently challenging as it requires multi-step, metric-grounded reasoning in complex 3D scenes.
To be specific, each reasoning step requires two key components:
\textbf{(1)} \textit{3D spatial referring} to resolve spatial relationships and accurately localize objects involved in the trace generation (\eg, identifying flowers with their from left to right order and locating them).
\textbf{(2)} \textit{3D spatial measuring} to understand absolute, real-world metric quantities related to the trace in captured scene (\eg, quantifying each flower's physical height and 1–5 cm height above each).
%



While recent Vision-Language Models (VLMs)~\cite{liu2023visual, liu2024nvila, bai2025qwen2} can perform 2D spatial reasoning~\cite{zhou2025roborefer, cheng2024spatialrgpt, chen2024spatialvlm, yuan2024robopoint} and even 2D visual trace (\ie, 2D positional sequence) generation~\cite{ji2025robobrain, yuan2025embodied, li2025hamster}, they overlook the multi-step nature of this task, particularly the crucial participation of intermediate objects involved in the trace, resulting in suboptimal generation.
Moreover, their outputs are mainly in 2D space, lacking 3D space grounding and absolute metric understanding, creating a fundamental chasm between 2D visual trace and 3D \textit{spatial trace}.

To this end, we propose {\mname}, a 3D-aware reasoning VLM that not only acquires precise \textit{3D spatial referring and measuring} via Supervised Fine-tuning~(SFT), but also exhibits multi-step metric-grounded reasoning capabilities for \textit{spatial tracing} via reinforcement fine-tuning~(RFT).  
The core of our approach is to introduce a set of metric-sensitive reward functions (\eg, referring, measuring, scale) during RFT.
These rewards supervise the key perceptual objects involved in the trace and offer crucial intermediate evidence for accurate spatial trace generation.
In addition, as \textit{3D spatial referring and measuring} require better metric-grounded understanding, we introduce a scale decoder for VLM, supervised by a regression loss on the predicted metric scale factor to enhance metric perception, even from RGB inputs.
Moreover, we incorporate a universal spatial encoder with the architectural flexibility to integrate diverse geometric configurations (\eg, camera intrinsics, absolute depth) and further improve the precision of \textit{spatial trace} generation.

To support {\mname}'s training, we present {\dname}, a large-scale dataset with $4.5$M high-quality examples and $30$M QA pairs to first learn \textit{3D spatial referring and measuring} in SFT, and further achieve \textit{spatial tracing} with compositional reasoning on both in RFT. 
It spans diverse indoor, outdoor and tabletop scenes with fine-grained annotations (\eg, precise geometry, object-level spatial referents) and contains a greatly higher proportion ($48.2\%$) of absolute-scale data ($14$x prior~\cite{sun2025spacevista}) for metric-grounded understanding.
To advance metric-grounded multi-step reasoning capabilities, it provides step-wise annotations of the reasoning process (up to $9$ steps).
Moreover, it has object-centric and end-effector-centric spatial traces spanning $3$ single-arm and dual-arm robot configurations, enhancing generalization and real-world applicability.


We evaluate {\mname} on spatial understanding, measuring, and referring benchmarks, achieving SOTA average success rate of \textbf{79.1\%}, exceeding Gemini-2.5-Pro by \textbf{11\%}.
It also outperforms all baselines on 2D visual trace benchmarks.
To address the lack of benchmarks for spatial tracing, we introduce {\bname}, which contains 100 real-world images with manually annotated tasks involving object localization, movement, and placement.
Each sample requires metric-grounded, multi-step reasoning (up to $8$ steps), with precise start-point masks, end-point 3D bounding boxes, precise geometry annotation, and diverse metrics for fine-grained evaluation.
\mname{} still achieves best performance, surpassing Gemini-2.5-Pro by \textbf{36\%}.
Moreover, in \cref{fig: motivation} and \cref{subsec: sim and real world}, \mname{} can execute long-horizon, dynamic tasks in cluttered real-world scenes, showing strong generalization across robots (\eg, UR5, G1 humanoid).

Our contributions are summarized as follows: 
\textbf{(1)} We propose {\mname}, a 3D-aware VLM that accepts arbitrary geometric inputs, uses scale supervision, guided by metric-sensitive rewards to achieve spatial tracing. 
\textbf{(2)} We construct  {\dname}, a well-annotated dataset tailored for spatial tracing, enabling both SFT and RFT training, and {\bname}, a benchmark that fills the gap in evaluating it.
\textbf{(3)} Extensive experiments show that {\mname} generalizes well, surpasses baselines in spatial understanding, measuring, referring, tracing, and efficiently controls diverse robots across tasks in real world.



    

\section{Related work}
\label{sec: related work}

\noindent \textbf{Spatial Reasoning with VLMs.} 
Spatial reasoning refers to the ability to perceive and reason about 3D space, comprising metric-agnostic and -grounded types. 
Metric-agnostic reasoning~\cite{chen2024spatialvlm, cheng2024spatialrgpt, song2024robospatial, zhou2025roborefer, yang2025vlaser} captures object-centric properties (\eg, position, orientation) and inter-object relations (\eg, left/right, near/far), whereas metric-grounded reasoning involves precise, absolute-scale measurements~\cite{chen2025sd, cai2024spatialbot, sun2025spacevista} (\eg, distance, depth, size) in the physical world.
Compared to metric-agnostic one, metric-grounded reasoning requires better 3D understanding.
Despite using 3D modalities, either explicitly~\cite{chen2025sd, zhou2025roborefer, cheng20253d, cai2024spatialbot, mao2025spatiallm} (\eg, point clouds, depth maps) or implicitly~\cite{chen2025think, wu2025spatial, fan2025vlm, zheng2025learning} (\eg, VGGT~\cite{wang2025vggt}), existing methods still struggle to reason complex absolute-scale scenes for \textit{3D spatial referring and measuring}.
We thus propose a 3D-aware VLM that uses scale supervision and accepts arbitrary geometric inputs to address this gap.
%
%
%
%
%

\vspace{+1mm}
\noindent \textbf{Trace Generation with VLMs for Robotics.}
Trace enhances manipulation by capturing the spatio-temporal dynamics of objects.
Recent advances in VLMs~\cite{achiam2023gpt,team2023gemini,anthropic2024claude,bai2025qwen2,qin2024mp5, qin2024worldsimbench, tan2025roboos, tan2025roboosnext, zhou2025chain, zhou2024minedreamer, chen2024rh20t, ji2026prm, wang2025towards,li2025llavaonevision} focus on predicting 2D visual traces (\ie, 2D point sequences) and guiding robotic actions in two ways.
\textbf{(1)} Lift-to-3D~\cite{xu2025a0, ji2025robobrain, team2025robobrain, yuan2025embodied}: Projects 2D traces into 3D spatial trace using depth maps and camera intrinsics.
\textbf{(2)} Overlap-on-2D~\cite{lee2025molmoact, li2025hamster, zheng2024tracevla}: Renders the 2D trace onto the image by steering the control policies for action generation.
However, 2D traces struggle to fully capture object dynamics in 3D space. 
Moreover, existing works for 2D visual trace struggle to handle complex spatially constrained instructions and supervise the key perception steps, also required by multi-step 3D spatial tracing, largely due to the lack of datasets.
We thus propose a new dataset and benchmark tailored for spatial tracing.

\vspace{+1mm}
\noindent \textbf{Reinforcement Fine-tuning for VLMs.}
Reinforcement Fine-tuning~(RFT)~\cite{bai2022training, rafailov2023direct, shao2024deepseekmath} is a post-training strategy that aligns models with human preferences or specific goals via feedback, complementing SFT~\cite{wei2021finetuned, zhou2023lima}, which adapts pre-trained models using task-oriented data.
Recent advances in VLMs use RFT to improve visual/spatial reasoning~\cite{yang2025r1, zhang2025r1, tan2025reason, kang2025viki, an2025agfsync, tan2026robobrain}, grounding~\cite{liu2025visual, zhan2025vision, shen2025vlm, zhou2025roborefer}, segmentation~\cite{liu2025seg}, tool-use~\cite{han2025tiger, zhang2026think3d}, coding~\cite{zhang2026prune4web}, safety~\cite{lu2026homeguard,lu2026bench}, and 2D visual trace prediction~\cite{liu2025seg, huang2025thinkact, yuan2025embodied}.
However, most approaches remain confined to 2D-relative perception and outcome-based reward, limiting their performance on spatial tracing tasks requiring 3D multi-step metric-grounded reasoning.
Therefore, we design metric-sensitive process rewards to supervise key perception steps in the reasoning process to meet the above expectations.

\section{Method}
\label{sec: method}


\subsection{Problem Formulation and Representation Advantages}
\label{subsec: problem formulation}

%
%

\begin{figure*}[t]
\centering
\includegraphics[width=\linewidth]{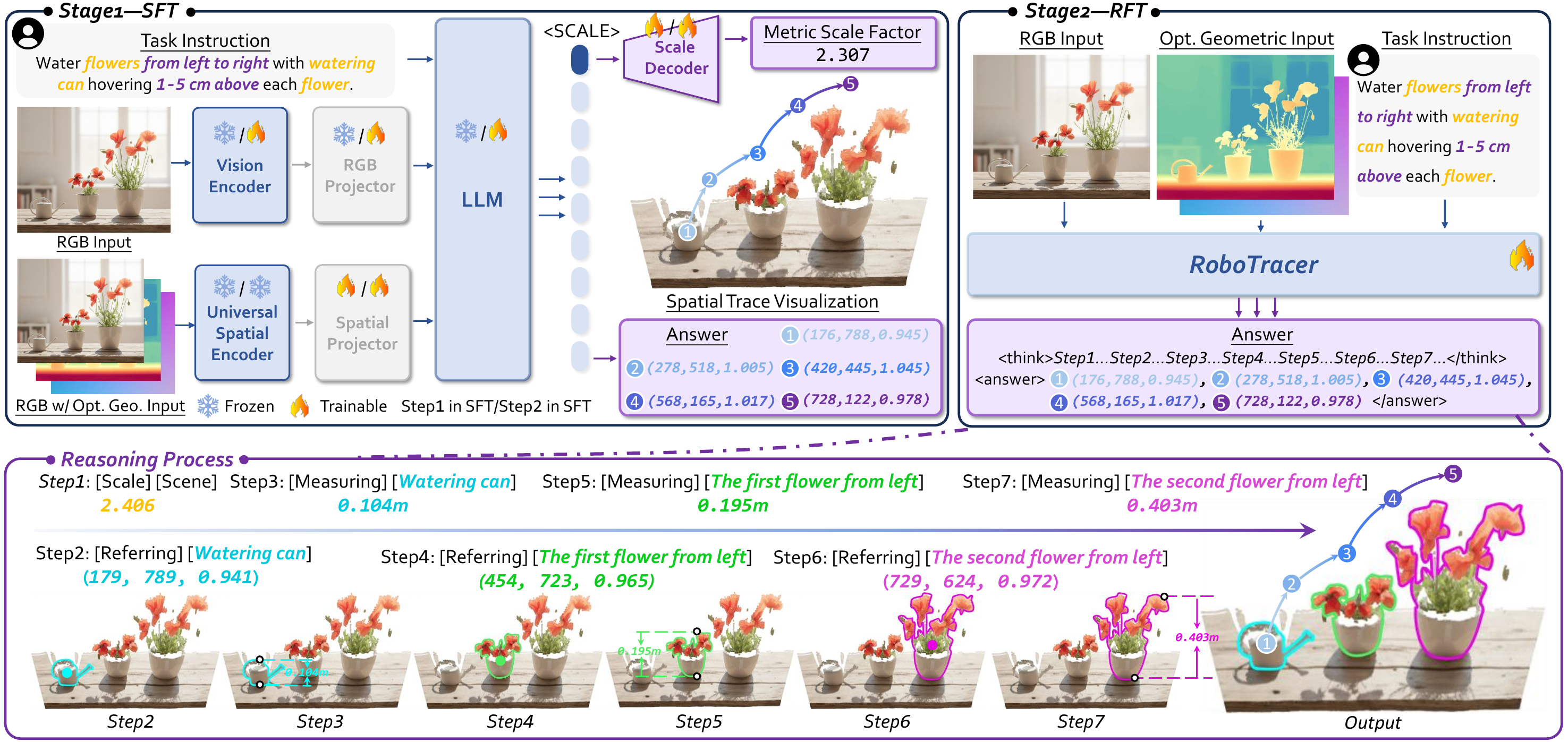}
\vspace{-5mm}
   \caption{
    Overview of RoboTracer.
    RoboTracer can process RGB images and task instructions, while flexibly integrating various geometric configurations when available to improve spatial precision (enabled by the integrated universal spatial encoder) with real-world scale awareness (powered by the scale decoder with regression loss).
    After SFT, metric-sensitive rewards in RFT further supervise the key perceptual steps in the trace and offer crucial intermediate evidence for accurate spatial trace generation.
    %
    %
   }
\label{fig: pipeline}
\vspace{-5mm}
\end{figure*}

We formulate \textit{spatial tracing} as predicting an ordered sequence of $6-12$ 3D sparse and discrete keypoints $\tau = \{p_t\}_{t=1}^{T}$ from visual inputs $\mathcal{O}$ (\eg, RGB, RGB-D), optional camera geometry $\mathcal{G}$ (\eg, intrinsics) and a textual instruction $\mathcal{T}$ via a VLM in a open-loop manner.
Each point $p_t = (u_t, v_t, d_t)$ comprises a image-plane coordinate $(u_t, v_t)$ and corresponding absolute depth $d_t$, encoding essential object-related and collision-critical waypoints.
The resulting trace \( \tau \) serves as a spatial plan for entities (\eg, a robot end-effector or object), and can be integrated with motion planning to predict physically feasible actions via adherence to the trace to follow the instruction.
%
%
Notably, the instruction encodes both \textit{3D spatial referring and measuring}, often requiring multi-step compositional reasoning.
For example, in \cref{fig: pipeline}, to accomplish the task of ``\textit{Water flowers from left to right with watering can hovering 1-5 cm above each flower}'', requires inferring the 3D positions and heights of all flowers in 3D scenes.
These spatial cues provide crucial intermediate evidence for multi-step reasoning, enabling accurate trace generation under spatial constraints at the start, end, and along the path.

Instead of predicting 3D \((x,y,z)\) coordinates directly, we adopt a decoupled \((u,v,d)\) formulation that is trivially convertible to 3D via camera intrinsics. 
This circumvents the need for VLMs to implicitly learn camera geometry, thereby simplifying training and improving accuracy. 
Moreover, \((u,v,d)\) points project into lower-dimensional spaces easily: omitting \(d\) yields 2D visual traces, and retaining only start/end points yields 3D or 2D spatial referring data. 
This formulation enhances data reusability, aligns seamlessly with existing 2D datasets~\cite{deitke2025molmo, lee2025molmoact, niu2024llarva} for co-training, thus bolstering multi-task learning performance.

Our spatial trace, combined with downstream motion planning for robotic control, offers the following advantages:
\textbf{(1)} Compared to end-to-end VLAs~\cite{liu2026sapave, Pi0, lee2025molmoact, li2026you, li2025robomirror, bai2026latent} that directly predict 6D dense actions, our approach can achieve superior zero-shot generalization without task-specific training, enabled by our generalizable formulation and less-constrained 3D space of this trace.
\textbf{(2)} As an intermediate representation that delegates spatial reasoning and metric perception from instructions to spatial trace, this trace can handle more spatially and metrically constrained tasks compared to direct VLAs.
\textbf{(3)} As a sequence of keypoints, spatial trace captures the intermediate execution process more effectively than start or end keypoint-based planning methods.

%
%
%

\subsection{RoboTracer}
\label{subsec: vlm}

\noindent\textbf{VLM Architecture.}
Spatial tracing requires a metric-grounded 3D understanding for referring and measuring, yet simply fine-tuning existing VLMs encounters two hurdles:
\textbf{(1)} insufficient absolute-scale supervision, especially with RGB-only data;
\textbf{(2)} underuse of absolute-scale geometric cues (\eg, camera intrinsics, depth) to enhance precision.
In \cref{fig: pipeline}, we address these issues by introducing a scale decoder and a universal spatial encoder, each aligned with the LLM via projectors, akin to the existing RGB encoder.
The scale decoder maps the \texttt{<SCALE>} token embedding into a numeric factor, linking scale-invariant representations to absolute metric scales.
Rather than classification loss, we use regression loss to supervise it to heighten real-world 3D scale awareness.
Moreover, we find that leveraging strong priors from a powerful feed-forward metric 3D geometry model~\cite{keetha2025mapanything} greatly enhances spatial and scale understanding. 
Building on this model, our universal spatial encoder flexibly integrates optional geometric cues (\eg, camera intrinsics, poses, depth) to refine spatial representations as more geometry becomes available.
This modular design enables:
\textbf{(1)} Flexible training, leveraging diverse scale-aware annotations in datasets via flexible input augmentation to enrich spatial learning;
\textbf{(2)} Geometry-adaptive inference, integrating available geometric cues without retraining or architectural changes, and more cues for better performance.
%
%
See Supp.~\ref{suppsubsec: architecture} for details.

%
%
%
%
%

\vspace{+1mm}
\noindent\textbf{Supervised Fine-tuning.}
Since general VLMs’ 2D-only pretraining limits 3D metric-grounded understanding, we propose a two-step SFT.
\textbf{(1)} Metric Alignment.
In \cref{fig: pipeline}, we align the spatial encoder and scale decoder with LLM by using geometric annotations (\eg, depth, scale) from the {\dname} dataset (see \cref{subsec: dataset}). 
During this stage, only the projector and scale decoder are updated.
\textbf{(2)} Metric Enhancement.
We freeze the spatial encoder and finetune all other components on {\dname} and additional instruction-following datasets~\cite{liu2024mitigating, liu2024improved, yu2016modeling}.
Crucially, we train on both RGB-only and RGB+$\mathcal{X}$ inputs, where $\mathcal{X}$ indicates arbitrary combinations of geometric annotations.
This preserves image encoder's general VQA ability while flexibly accommodating diverse geometric configurations without retraining during inference.
The SFT loss is defined as:
%

{
\begin{equation}
\mathcal{L} = \mathcal{L}_{ntp} + 0.1 \|\log (\hat{s})-\operatorname{stopgrad}(\log(s^{*})\|_{2}^{2}
\tag{1}
\end{equation}
}

\noindent where $\mathcal{L}_{\text{ntp}}$ is next-token prediction loss, $\log (\hat{s})$ is predicted scale in logarithmic space, \(s^{*}\) is ground-truth scale.
Moreover, {\dname} contains multi-step data with reasoning processes, providing a  ``cold start'' for the subsequent RFT stage.
%
%
Please check Supp.~\ref{suppsubsec: training data} for details about SFT training configuration.

\vspace{+1mm}
\noindent\textbf{Reinforcement Fine-tuning.}
We use RFT after SFT to improve compositional metric-grounded reasoning using GRPO~\cite{shao2024deepseekmath} with multi-step reasoning data from {\dname}.
We first define outcome-based rewards:
\textbf{(1)} Outcome Format Reward ($R_\text{OF}$) for structured outputs;
\textbf{(2)} Point Reward ($R_P$) for start/end point consistency between the predicted trajectory \((p_1, p_T)\) in \(\hat{\tau} = \{\hat{p_t}\}_{t=1}^{T}\) vs. \(\tau\). Formally,  \(R_P = \tfrac{1}{2}\bigl[f(p_1, \hat{p}_1) + f(p_T, \hat{p}_T)\bigr]\),\(f(p, p') = \max\bigl(0,\; 1 - \|p - p'\|_2^2\bigr). \)  
%
%
\textbf{(3)} Trace Reward ($R_\text{T}$) for trajectory-level alignment using a distance metric \(d(\tau, \hat{\tau})\):  \(R_{\text{T}} = \max\bigl(0,\; 1 - d(\tau, \hat{\tau})\bigr).\)  
All \((u,v,d)\) values are normalized to \([0,1]\), with depth scaled by the scene’s maximum depth.
%
%
%
%
%
However, the outcome-based rewards described above are metric-agnostic and provide no supervision over the key perceptual objects involved in trace generation (\eg, \textit{3D spatial measuring and referring}).
To address this, we introduce metric-sensitive process rewards that leverage key-step perception annotations from {\dname}:
\textbf{(1)} Process Format Reward ($R_\text{PF}$), enforcing the format ``[Perception Type] [Target Object]:'';
\textbf{(2)} Accuracy Reward ($R_\text{Acc}$), which applies to steps included in the key-step perception annotations.
For each relevant step, we measure the prediction error using a specific metric, according to the perception type (\eg, L1 distance for referring).
Notably, this design is order-invariant, allowing flexible step ordering.
The final reward sums both outcome-/process-based terms, with process-based terms scaled by \(0.25\).
%
%
\cref{fig: pipeline} shows that RFT-trained model generalizes well to 7-step spatial tracing, progressively resolving complex spatial relations and producing accurate trace.
Please check Supp.~\ref{suppsubsec: SFT training details} for more details.

\begin{figure*}[t]
\centering
\includegraphics[width=\linewidth]{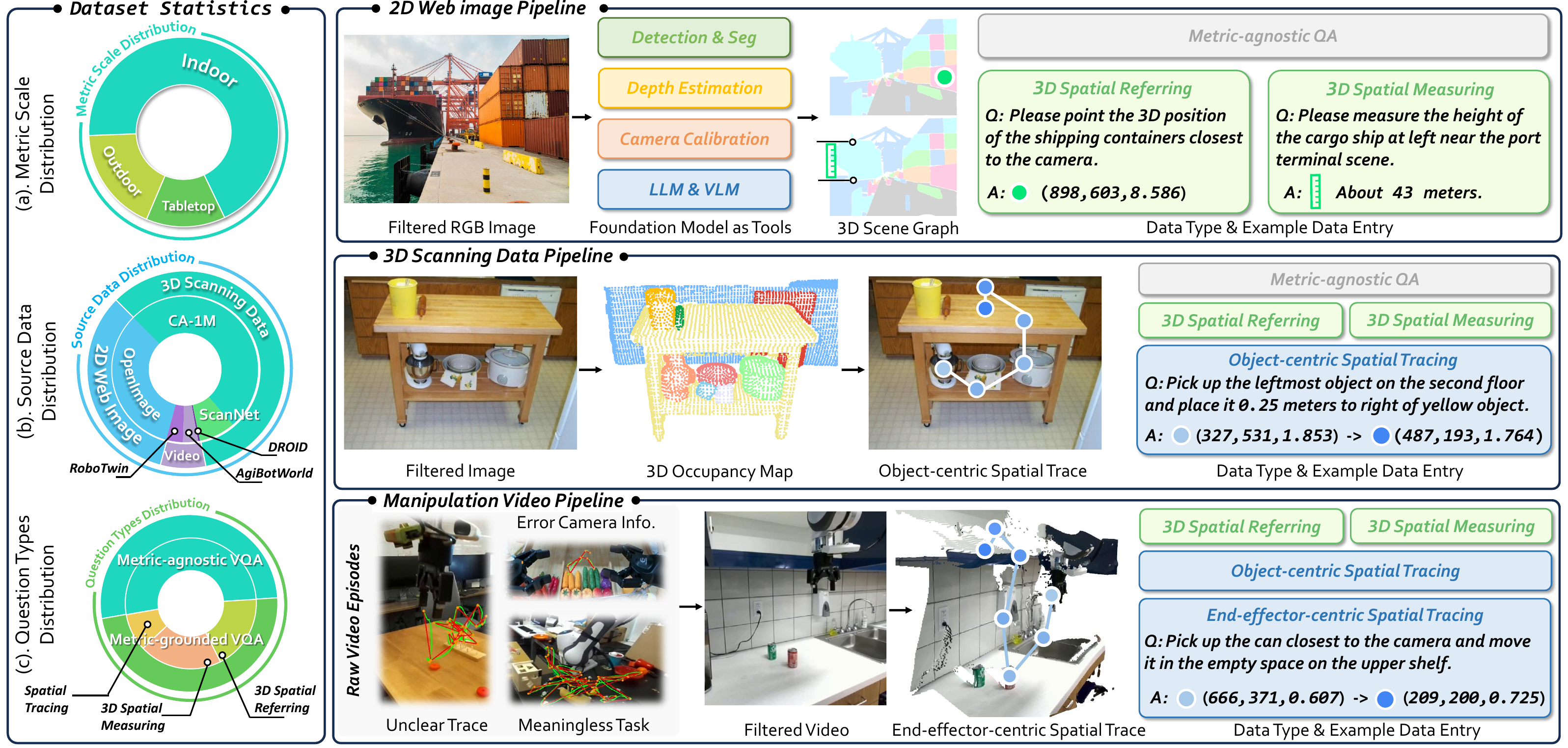}
\vspace{-5mm}
   \caption{{\dname} has 4.5M data samples from 2D, 3D, and video sources, covering outdoor, indoor, and tabletop scenes.
   It contains metric-agnostic/-grounded QA pairs for 3D spatial referring, measuring, tracing with a detailed reasoning process.}
\label{fig: dataset}
\vspace{-3mm}

\end{figure*}

\subsection{TraceSpatial Dataset}
\label{subsec: dataset}

%

\noindent \textbf{Overview.}
Key features are:
\textbf{(1) Rich Diversity.} {\dname} spans outdoor, indoor, and tabletop scenes (\cref{fig: dataset}~(a)) and includes both object-centric and end-effector-centric spatial traces; the latter captures gripper motions across $3$ single-arm and dual-arm robot configurations, fostering model generalization in open-world scenarios.
\textbf{(2) Multi-Dimensionality.} Beyond metric-agnostic spatial concepts and relations, the dataset includes $48.2\%$ metric-grouded QA (\cref{fig: dataset}~(b)).
These samples cover 3D spatial measuring and referring, and support multi-step spatial tracing by providing detailed annotations of reasoning process, addressing limitations in existing datasets~\cite{ji2025robobrain, yuan2025embodied}.
\textbf{(3) Large Scale.} With $4.5$M samples and $30$M QA pairs, {\dname} is the largest dataset for 3D spatial reasoning, supporting bottom-up spatial tracing learning.
\textbf{(4) Fine-Grained.} Hierarchical object captions, from coarse categories (\eg, “flower”) to fine-grained spatial referents (\eg, “the first flower from the left”), enable 3D spatial measuring, referring, and tracing in cluttered scenes.
Absolute-scale geometry (\eg, intrinsics, depth) further enriches spatial learning and flexible input augmentation.
\textbf{(5) High Quality.}  Rigorous filtering preserves spatial relevance. From $1.7$M OpenImages~\cite{kuznetsova2020open}, $466$k images remain; from CA-1M~\cite{lazarow2024cubify}($2$M) and ScanNet~\cite{dai2017scannet}($190$k), $100$k and $12$k frames are retained based on text-identifiable 3D boxes, respectively. From DROID~\cite{khazatsky2024droid}($116$k) and AgiBot~\cite{contributors2024agibotworldrepo}($167$k) videos, $2$0k and $59$k episodes are preserved after verifying valid camera poses, coherent tasks, and clean trajectories, respectively.
We also manually inspect a subset to verify data quality and iteratively refined the dataset.
\textbf{(6) Easy Scalability.} Our pipeline scalably integrates 2D images, 3D scans with bounding boxes, and manipulation videos. 
Even faced with RGB-only datasets in new domains, we can still scale up easily and acquire multi-step reasoning annotations automatically.

\vspace{+1mm}
\noindent \textbf{Data Recipe.}
In Fig.~\ref{fig: dataset}, we propose a data pipeline that progressively integrates 2D/3D/video sources for general VLMs to perform 3D spatial referring/measuring/tracing.
\textbf{(1) 2D Web Images} aim to provide basic spatial concepts and broad-scale perception across indoor and outdoor scenes.
We filter $1.7$M images from OpenImage~\cite{kuznetsova2020open} down to $466$K and employ VLM~\cite{bai2025qwen2} with hierarchical region-captioning to produce fine-grained spatial references, surpassing previous approaches~\cite{song2024robospatial, daxberger2025mm}.
Object captions serve as nodes for 3D scene graphs, where edges depict spatial relations inferred via object detection, depth, and camera estimation.
Using template-/LLM-based methods, we generate metric-agnostic and 3D spatial QA grounded in these captions.
\textbf{(2) 3D Scanning Datasets} want to arm the model with a focused metric-grounded spatial reasoning of indoor scenes.
We thus leverage the richly annotated CA-1M~\cite{lazarow2024cubify} and ScanNet~\cite{dai2017scannet}. 
After fine-grained filtering, we construct 3D scene graphs with more diverse spatial relations, enabled by precise 3D bounding boxes compared to 2D approaches. 
Moreover, we generate 3D occupancy maps that encode positions, orientations, and metric distances (\eg, ``35cm right of the toy'') for accurate object-centric spatial trace generation.
\textbf{(3) Manipulation Videos} provide spatial traces aligned with the embodied manipulation in tabletop settings.
While 3D scans enable object-centric tracing, they lack physically plausible manipulations.
Hence, we curate real and simulated~\cite{chen2025robotwin} tabletop videos (from $167$k to $59$k for AgiBot~\cite{contributors2024agibotworldrepo}, and from $116$k to $24$k for DROID~\cite{khazatsky2024droid}) with calibrated cameras, accurate task execution, and clear trajectories. 
We further leverage VLM~\cite{bai2025qwen2} to decompose these tasks into subgoals, enabling precise multi-step spatial tracing for single-/dual-arm across $3$ robots.
Notably, as 3D datasets and simulation videos are all-seeing, we construct multi-step, metric-grounded spatial tracing data, under the assumption that the generated code reflects optimal reasoning, with each line translated into textual form and intermediate results filled into structured formats (\eg, coordinates, distances).
%
%
%
%
See Supp.~\ref{suppsec: dataset} for details.


\section{Experiments}
\label{sec: experiments}



\noindent \textbf{Model Configuration.}
We adopt NVILA~\cite{liu2024nvila} (2B/8B) as base model and apply SFT. 
Due to computational limits, we only perform RFT on 2B model. 
Since the model can accept arbitrary geometric inputs, it defaults to using only images when input types are unspecified.
%
%
%
%
Please check Supp.~\ref{suppsec: implementation details} for details.

\noindent \textbf{Robotic Evaluation Configuration.}
We ensure physical feasibility and smoothness by motion planning via adherence to the spatial trace, collision avoidance, and joint limits with a smoothness cost.
We also correct VLM-generated spatial traces before use to satisfy physical constraints.
Please see Supp.~\ref{suppsubsec: real-world evaluation} for details.

\begin{table}[t]
\caption{Performance on spatial understanding benchmarks. Rel., Dep., Dist. denote relation, depth, distance.
Top-1/-2 success rate are indicated by \textbf{bold}/\underline{underlined} text.
%
}
\vspace{-2.5mm}
\centering 
\tiny
\setlength{\tabcolsep}{2pt}
\begin{tabular}{ll|ccc|cc|cc}
\toprule
\multirow{2}{*}{Method}  & \multirow{2}{*}{Input}
& \multicolumn{3}{c|}{CV-Bench~\cite{tong2024cambrian}}
& \multicolumn{2}{c|}{$\text{BLINK}_{val}$~\cite{fu2024blink}}
& \multirow{2}{*}{RoboSpatial~\cite{song2024robospatial}}
& \multirow{2}{*}{EmbSpatial~\cite{du2024embspatial}} \\

& 
& 2D-Rel. & 3D-Dep. & 3D-Dist.
& 2D-Rel. & 3D-Dep.
& & \\
\midrule
GPT-4o~\cite{achiam2023gpt}              & RGB   & 84.62 & 86.50 & 83.33 & 82.52 & 78.23 & 77.20 & 63.38 \\
Gemini-2.5-Pro~\cite{comanici2025gemini} & RGB   & 93.54 & 91.00 & \underline{90.67} & \textbf{91.61} & 87.90 & 77.24 & 76.67 \\
\midrule
NVILA-2B~\cite{liu2024nvila}            & RGB   & 70.15 & 79.67 & 60.00 & 67.83 & 62.10 & 51.79 & 47.34 \\
NVILA-8B~\cite{liu2024nvila}            & RGB   & 91.54 & 91.83 & \underline{90.67} & 76.92 & 76.61 & 59.35 & 67.72 \\
Qwen-3-VL-4B~\cite{qwen3technicalreport} & RGB  & 92.31 & 94.67 & 87.50 & \underline{87.71} & 85.48 & 79.67 & 77.01 \\
Qwen-3-VL-8B~\cite{qwen3technicalreport} & RGB  & 93.85 & 94.50 & 90.33 & \underline{87.41} & 85.48 & 77.24 & \underline{77.86} \\
\midrule
Molmo 7B-D~\cite{deitke2025molmo}       & RGB   & 87.69 & 66.00 & 61.83 & 59.44 & 77.42 & 58.60 & 58.74 \\
SpaceVLM-13B~\cite{chen2024spatialvlm}  & RGB   & 63.69 & 66.83 & 70.17 & 72.73 & 62.90 & 61.00 & 49.40 \\
RoboBrain 2.0-7B~\cite{team2025robobrain} & RGB & 96.00 & 94.83 & 90.00 & 79.72 & 85.48 & 74.80 & 74.78 \\
SpatialBot-3B~\cite{cai2024spatialbot}  & RGB-D & 69.38 & 77.33 & 60.83 & 67.83 & 67.74 & 72.36 & 50.66 \\
\midrule
\cellcolor{mylightblue}\mname-2B-SFT  & \cellcolor{mylightblue}RGB  
& \cellcolor{mylightblue}\underline{96.62}
& \cellcolor{mylightblue}\underline{96.00}
& \cellcolor{mylightblue}89.83
& \cellcolor{mylightblue}83.22
& \cellcolor{mylightblue}\underline{91.94}
& \cellcolor{mylightblue}\underline{82.93}
& \cellcolor{mylightblue}70.66 \\
\cellcolor{myblue}\mname-8B-SFT   & \cellcolor{myblue}RGB  
& \cellcolor{myblue}\textbf{97.08}
& \cellcolor{myblue}\textbf{97.17}
& \cellcolor{myblue}\textbf{93.50}
& \cellcolor{myblue}\textbf{91.61}
& \cellcolor{myblue}\textbf{92.74}
& \cellcolor{myblue}\textbf{83.74}
& \cellcolor{myblue}\textbf{81.75} \\
\bottomrule[1pt]
\end{tabular}
\label{tab: spatial_understanding}
\vspace{-3mm}
\end{table}

\begin{table}[t]
\centering
\tiny

\begin{minipage}[t]{0.49\textwidth}
\captionsetup{type=table}
\caption{Performance on spatial measuring benchmarks. S.N., S.E., R.E. denote Plus, Scannet, Scale Estimation, Refer Estimation.
Top-1/-2 success rate (\%) are indicated by \textbf{bold}/\underline{underlined} text.}
\vspace{-1.1mm}
\centering
\setlength{\tabcolsep}{1pt}
\begin{tabular}{ll|cc|cc}
\toprule
\multirow{2}{*}{Method} & \multirow{2}{*}{Input}
& \multicolumn{2}{c|}{Q-spatial~\cite{liao2024reasoningpathsreferenceobjects}}
& \multicolumn{2}{c}{MSMU~\cite{chen2025sdvlm}} \\
& & Plus & S.N. & S.E. & R.E. \\
\midrule
GPT-4o~\cite{achiam2023gpt}              & RGB   & 31.68 & 37.06 & 3.86  & 2.09 \\
Gemini-2.5-Pro~\cite{comanici2025gemini} & RGB   & 56.44 & 70.00 & 64.86 & 48.42 \\
\midrule
NVILA-2B~\cite{liu2024nvila}             & RGB   & 36.90 & 40.59 & 40.15 & 37.37 \\
NVILA-8B~\cite{liu2024nvila}             & RGB   & 46.87 & 44.71 & 33.98 & 38.95 \\
Qwen-3-VL-4B~\cite{qwen3technicalreport} & RGB   & 53.47 & 70.00 & 63.71 & 42.11 \\
Qwen-3-VL-8B~\cite{qwen3technicalreport} & RGB   & 29.70 & 56.47 & 63.32 & 52.63 \\
\midrule
Molmo 7B-D~\cite{deitke2025molmo}        & RGB   & 51.49 & 63.53 & 59.85 & 43.68 \\
SpaceVLM-13B~\cite{chen2024spatialvlm}   & RGB   & 25.74 & 45.29 & 32.42 & 31.58 \\
RoboBrain 2.0-7B& RGB   & 44.55 & 55.88 & 69.11 & 48.42 \\
SpatialBot-3B~\cite{cai2024spatialbot}   & RGB-D & 20.79 & 29.41 & 16.60 & 22.62 \\
\midrule
\cellcolor{mylightblue}\mname-2B-SFT & \cellcolor{mylightblue}RGB
& \cellcolor{mylightblue}\underline{68.32}
& \cellcolor{mylightblue}\underline{70.59}
& \cellcolor{mylightblue}\underline{78.38}
& \cellcolor{mylightblue}\underline{60.00} \\
\cellcolor{myblue}\mname-8B-SFT & \cellcolor{myblue}RGB
& \cellcolor{myblue}\textbf{73.27}
& \cellcolor{myblue}\textbf{78.82}
& \cellcolor{myblue}\textbf{83.01}
& \cellcolor{myblue}\textbf{70.00} \\
\bottomrule[1pt]
\end{tabular}
\label{tab: spatial_measuring}
\end{minipage}
\hfill
\begin{minipage}[t]{0.49\textwidth}
\captionsetup{type=table}
\caption{
Performance on 2D spatial referring benchmarks.
W.2.P., RoboS., RefS.-L. and RefS.-P. denote Where2Place~\cite{yuan2024robopoint}, RoboSpatial~\cite{song2024robospatial}, and Location and Placement parts of RefSpatial-Bench~\cite{zhou2025roborefer}, respectively.
Top-1/-2 success rate (\%) are indicated by \textbf{bold}/\underline{underlined} text.
}
\vspace{-2.5mm}
\centering
\setlength{\tabcolsep}{1pt}
\begin{tabular}{l|ccccc}
\toprule
\textbf{Model} & W.2.P. & RoboS. & RefS.-L. & RefS.-P. \\
\midrule
Gemini-2.5-Pro~\cite{comanici2025gemini} & 61.90 & 40.20 & 46.96 & 24.21 \\
\midrule
Qwen3-VL-4B  & \underline{66.0} & 63.11 & 43.00 & \textbf{54.55} \\
Qwen3-VL-8B ~\cite{qwen3technicalreport} & \underline{64.00} & 61.48 & \underline{51.00} & 42.00 \\
\midrule
SpaceLLaVA~\cite{chen2024spatialvlm} & 11.8 & 16.0 & 5.82 & 4.31 \\
RoboPoint~\cite{yuan2024robopoint} & 46.80 & 41.30 & 22.87 & 9.27 \\
Molmo-7B~\cite{deitke2025molmo}  & 45.00 & 38.00 & 21.91 & 12.85 \\
Molmo-72B~\cite{deitke2025molmo} & 63.80 & 40.90 & 45.77 & 14.74 \\
RoboBrain 2.0-7B ~\cite{team2025robobrain}& 63.59 & 54.87 & 36.00 & 29.00 \\
\midrule
\cellcolor{mylightblue}{\mname}-2B-SFT & \cellcolor{mylightblue}63.00 & \cellcolor{mylightblue}\underline{62.30} & \cellcolor{mylightblue}49.00 & \cellcolor{mylightblue}{45.00} \\
\cellcolor{myblue}{\mname}-8B-SFT & \cellcolor{myblue}\textbf{69.00} & \cellcolor{myblue}\textbf{66.40}& \cellcolor{myblue}\textbf{55.00} & \cellcolor{myblue}\underline{53.00} \\
\bottomrule[1pt]
\end{tabular}
\label{tab: referring}
\end{minipage}
\vspace{-6mm}
\end{table}

\begin{table}[t]
\caption{
Performance on the 2D visual trace benchmarks. RMSE denotes Root Mean Square Error.
Top-1/-2 scores are shown by \textbf{bold}/\underline{underlined} text.
}
\vspace{-2.5mm}
\centering
\scriptsize
\begin{tabular}{l|ccc|ccc}
\toprule
\multirow{3}{*}{Model} & \multicolumn{3}{c|}{ShareRobot-Bench~\cite{ji2025robobrain}} & \multicolumn{3}{c}{VABench-V~\cite{yuan2025seeing}} \\
 & Discrete Fréchet  & Hausdorff  & \multirow{2}{*}{RMSE $\downarrow$} & Discrete Fréchet  & Hausdorff  & \multirow{2}{*}{RMSE $\downarrow$} \\
 & Distance $\downarrow$ & Distance $\downarrow$ &  & Distance $\downarrow$ & Distance $\downarrow$ & \\
\midrule
Qwen3-VL-4B ~\cite{qwen3technicalreport} & 0.3808 & 0.3294 & 0.2204  & 0.2792 & 0.2528 & 0.2037  \\
Qwen3-VL-8B ~\cite{qwen3technicalreport} & 0.3922 & 0.3411 & 0.2328  & 0.2741 & 0.2549 & 0.2021 \\
\midrule
MolmoAct ~\cite{lee2025molmoact} & 0.7764 & 0.7764 & 0.6771  & 0.8136 & 0.8136 & 0.6877  \\
HAMSTER ~\cite{li2025hamster} & 0.4365 & 0.3919 & 0.3554  & 0.2124 & 0.2045 & 0.1825  \\
RoboBrain 2.0-3B  ~\cite{team2025robobrain}& 0.1974 & 0.1853 & 0.1380  & 0.3642 & 0.2820 & 0.2445 \\
RoboBrain 2.0-7B  ~\cite{team2025robobrain}& 0.1669 & 0.1575 & 0.1250  & 0.3289 & 0.2604 & 0.2237 \\
Embodied-R1-3B  ~\cite{yuan2025embodiedr1reinforcedembodiedreasoning}& 0.3426 & 0.3002 & 0.2388  & 0.3028 & 0.2588 & 0.2129 \\

\midrule
 \cellcolor{mylightblue}{\mname}-2B-SFT  & \cellcolor{mylightblue}\underline{0.1605} & \cellcolor{mylightblue}\underline{0.1544} & \cellcolor{mylightblue}\underline{0.1114}  & \cellcolor{mylightblue}\underline{0.1664} & \cellcolor{mylightblue}\underline{0.1551} & \cellcolor{mylightblue}\underline{0.1237}  \\
 \cellcolor{myblue}{\mname}-8B-SFT & \cellcolor{myblue}\textbf{0.1449} & \cellcolor{myblue}\textbf{0.1384} & \cellcolor{myblue}\textbf{0.0966}  & \cellcolor{myblue}\textbf{0.1494} & \cellcolor{myblue}\textbf{0.1367} & \cellcolor{myblue}\textbf{0.1065}  \\
\bottomrule
\end{tabular}
\label{tab: 2D_trace}
\end{table}

\begin{figure*}[t]
\centering
\includegraphics[width=\linewidth]{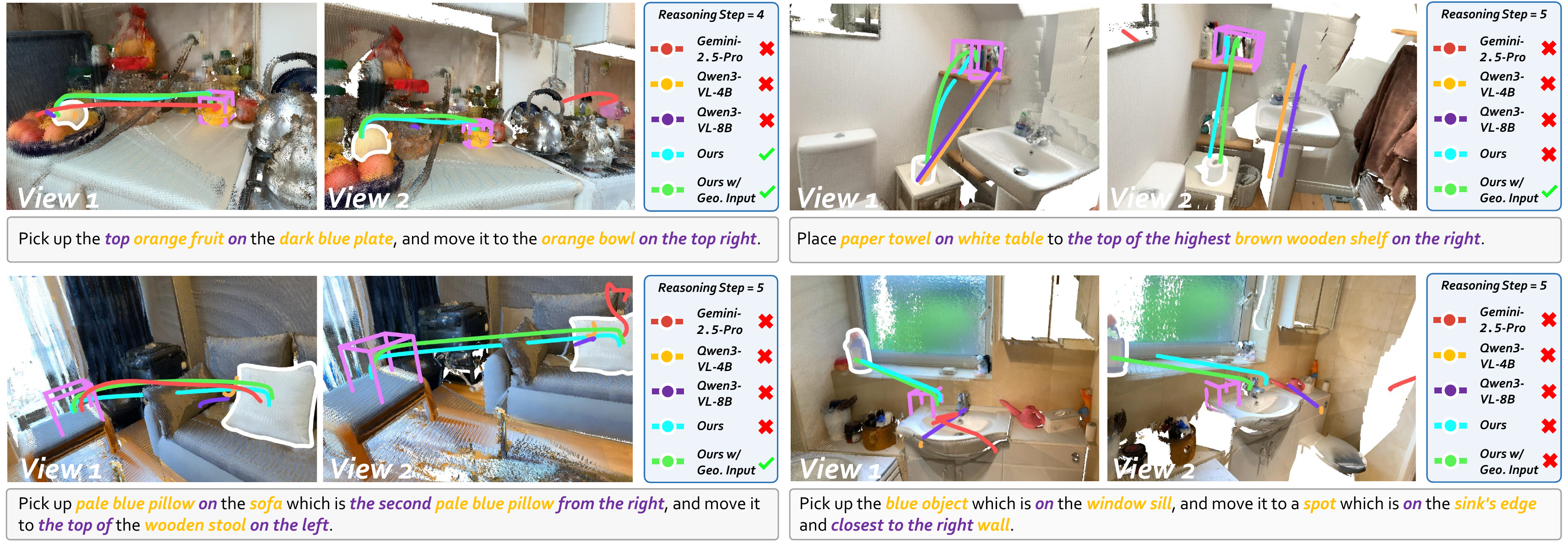}
\vspace{-6mm}
   \caption{
   {\bname} Results.
   White masks denote ground-truth 3D starting region; pink 3D boxes mark correct end regions.
   Despite similar 2D projections (left views of each case), our model yields more accurate spatial traces than strong general VLMs, which often produce floating or colliding traces due to inaccurate depth estimation.
   Leveraging richer geometric cues further improves performance.
   }
\label{fig: benchmark}
\vspace{-3mm}

\end{figure*}

\subsection{Spatial Understanding and Measuring}
\label{subsec: spatial understanding and measuring}

We evaluate our SFT model on public spatial understanding benchmarks, including CV-Bench~\cite{tong2024cambrian}, BLINK~\cite{fu2024blink} (validation set), RoboSpatial~\cite{song2024robospatial} (configuration subset), and EmbSpatial~\cite{du2024embspatial}.
We also assess spatial measuring performance on Q-Spatial~\cite{liao2024reasoningpathsreferenceobjects} and MSMU~\cite{chen2025sdvlm}.
Please refer to Supp.~\ref{suppsubsec: spatial understanding benchmarks} and \ref{suppsubsec: spatial measuring benchmarks} for details.
%

\vspace{+1mm}
\noindent \textbf{SFT learns strong spatial understanding and metric measuring.}
In \cref{tab: spatial_understanding} and \cref{tab: spatial_measuring}, {\mname}-8B-SFT trained solely on {\dname} surpasses all baselines with an average success rate of {$85.7\%$}, even outperforming Gemini-2.5-Pro by {$8.58\%$} and base model NVILA-8B by {$20.3\%$}.
Notably, we find that its improvements are more pronounced in 3D-related and measurement tasks compared to 2D tasks ({$23.6\%$} \textit{vs.}{$14.7\%$}), showing that our architectural design and curated dataset during SFT enhance the model’s 3D spatial and scale awareness.

\subsection{2D Spatial Referring and Visual Trace}
\label{subsec: spatial referring and tracing}

We evaluate 2D spatial referring (Where2Place~\cite{yuan2024robopoint}, RoboSpatial~\cite{song2024robospatial}, RefSpatial-Bench~\cite{zhou2025roborefer}) and visual trace benchmarks (ShareRobot-Bench~\cite{ji2025robobrain}~(end-effector-centric), VABench-V~\cite{yuan2025seeing})~(object-centric).
See Supp.~\ref{suppsubsec: spatial referring benchmarks} and \ref{2D visual tracing benchmarks} for details.
%

\vspace{+1mm}
\noindent \textbf{Decoupled formulation makes multi-task learning easy.}
Our {\mname} achieves best performance on both 2D spatial referring (see \cref{tab: referring}) and visual trace (see \cref{tab: 2D_trace}) benchmarks.
%
%
While {\dname} focuses on 3D metric-grounded data without explicitly targeting 2D tasks, our decoupled point formulation enables seamless projection of {\dname} into the data formats required for such 2D tasks (see \cref{subsec: problem formulation}).
%
%
%
Moreover, in \cref{tab: ablation} (ID G \& H), modeling $(u, v, d)$ surpasses direct 3D $(x, y, z)$ modeling.
We attribute this to \textbf{(1)} our dataset reuse via dimensionality reduction at various levels (\eg, 3D to 2D, point sequences to individual points), \textbf{(2)} stronger alignment with existing 2D datasets for co-training, thereby improving multi-task learning performance.
%



\subsection{Multi-Step Spatial Tracing}
\label{subsec: multi-step spatial tracing}

To evaluate complex multi-step, metric-grounded spatial tracing, we introduce {\bname}, a challenging real-world benchmark focusing on cluttered indoor/tabletop scenes. 
The dataset comprises $100$ images with precise geometric annotations (\eg, camera geometry, absolute depth, and 3D occupancy). 
Each sample requires $3–8$ reasoning steps, specifying both the manipulated object’s start mask and its end 3D box.
We assess performance in 2D and 3D; a trajectory succeeds if its start and end positions are correct and all intermediate paths remain collision-free.
See Supp.~\ref{suppsubsec: spatial tracing benchmarks} for details.
We present our analyses below.

\begin{table}[t]
\caption{
Performance on {\bname}. 
%
%
%
R., I., D. denote RGB image, intrinsics, absolute depth. N. means adding noise to depth input.
We report success rate (SR).
%
}
\vspace{-2.5mm}
\centering
\scriptsize
\setlength{\tabcolsep}{0.5pt}
\begin{tabular}{l|l|c|cc|cc|c}
\toprule
\multirow{2}{*}{Model} & \multirow{2}{*}{Input}  & Process  & 2D Start & 2D End &3D Start & 3D End & Overall\\
    &   &  Reward & SR (\%) $\uparrow$ & SR (\%) $\uparrow$ & SR (\%) $\uparrow$ & SR (\%) $\uparrow$ & SR (\%) $\uparrow$ \\
\midrule
Gemini-2.5-Pro ~\cite{comanici2025gemini}& RGB & - & 31 & 33 &9 & 16 & 3  \\
Qwen3-VL-4B~\cite{qwen3technicalreport} & RGB  & - & \underline{64} & 16 & 43 & 24 & 4\\
Qwen3-VL-8B ~\cite{qwen3technicalreport}& RGB  & - & 60 & 21 & 47 & 22 & 6\\
\midrule
MolmoAct ~\cite{lee2025molmoact} & RGB& -   & 4 & 15 & - & - & - \\
Magma~\cite{yang2025magma}      & RGB & -   & 28 & 17 & - & - & - \\
RoboBrain 2.0-7B  ~\cite{team2025robobrain} & RGB& -  & 46 & 23  & - & - & -\\
Embodied-R1-3B ~\cite{yuan2025embodiedr1reinforcedembodiedreasoning} & RGB& -  & 63 & 13 & - & - 
& - \\
\midrule
RoboRefer-2B (finetuned)  & R.D. & -& 48 & 43 & 60 & 51  & 28 \\
LEO-7B (finetuned)        & PointCloud & - & 46 & 39 & 58 & 48 & 27 \\
\midrule
\cellcolor{mylightblue}{\mname}-2B-SFT & \cellcolor{mylightblue}RGB&\cellcolor{mylightblue}-  & \cellcolor{mylightblue}56 & \cellcolor{mylightblue}44 & \cellcolor{mylightblue}63 & \cellcolor{mylightblue}52 & \cellcolor{mylightblue}31\\
\cellcolor{mylightblue}{\mname}-2B-SFT & \cellcolor{mylightblue}R.I.D.&\cellcolor{mylightblue}-  & \cellcolor{mylightblue}59 & \cellcolor{mylightblue}45 & \cellcolor{mylightblue}\underline{75} & \cellcolor{mylightblue}\underline{56} & \cellcolor{mylightblue}38\\
\midrule
\cellcolor{mylightblue}{\mname}-RFT & \cellcolor{mylightblue}RGB&\cellcolor{mylightblue}\ding{55} & \cellcolor{mylightblue}61 & \cellcolor{mylightblue}46 & \cellcolor{mylightblue}64 & \cellcolor{mylightblue}54 & \cellcolor{mylightblue}33 \\
\midrule

\cellcolor{myblue}{\mname}-RFT & \cellcolor{myblue}RGB&\cellcolor{myblue}\checkmark & \cellcolor{myblue}63 & \cellcolor{myblue}47 & \cellcolor{myblue}73 & \cellcolor{myblue}55 & \cellcolor{myblue}39 \\
\cellcolor{myblue}{\mname}-RFT & \cellcolor{myblue}R.I.&\cellcolor{myblue}\checkmark & \cellcolor{myblue}{64} & \cellcolor{myblue}{48} & \cellcolor{myblue}73 & \cellcolor{myblue}{56} & \cellcolor{myblue}{40} \\
\cellcolor{myblue}{\mname}-RFT (2D pipeline) & \cellcolor{myblue}R.I.D.&\cellcolor{myblue}\checkmark  & \cellcolor{myblue}66 & \cellcolor{myblue}51 & \cellcolor{myblue}75 & \cellcolor{myblue}58 & \cellcolor{myblue}42\\
\cellcolor{myblue}{\mname}-RFT & \cellcolor{myblue}R.I.D. w/ N.&\cellcolor{myblue}\checkmark & \cellcolor{myblue}68 & \cellcolor{myblue}53 & \cellcolor{myblue}76 & \cellcolor{myblue}59 & \cellcolor{myblue}44 \\
\cellcolor{myblue}{\mname}-RFT & \cellcolor{myblue}R.I.D.&\cellcolor{myblue}\checkmark & \cellcolor{myblue}\textbf{69} & \cellcolor{myblue}\textbf{53} & \cellcolor{myblue}\textbf{78} & \cellcolor{myblue}\textbf{61} & \cellcolor{myblue}\textbf{45} \\
\bottomrule
\end{tabular}
\label{tab: spatial trace}
\vspace{-3mm}
\end{table}

\vspace{+1mm}
\noindent \textbf{RFT enables better multi-step metric-guided reasoning.}
In \cref{tab: spatial trace}, {\mname}-RFT outperforms all baselines on 3D metrics, exceeding Gemini-2.5-Pro by {$36\%$}.
We find that while these VLMs perform well on 2D referring and tracing, they fall short in 3D spatial tracing due to their limited metric depth understanding, often producing traces that float or collide with objects.
%
Moreover, {\mname}-RFT also outperforms the baselines (\eg, LEO, RoboRefer) with 3D input after being finetuned by {\dname}.
This shows that {\mname}-RFT leverages learned 3D spatial knowledge and compositional reasoning to generate more accurate spatial traces.
\cref{fig: benchmark} further shows complex spatial tracing cases from {\bname} with model comparisons.

\vspace{+1mm}
\noindent \textbf{Multi-step processes enable better reasoning with robustness.}
In \cref{tab: spatial trace} and \cref{tab: ablation} (ID D), we find that metric-grounded reasoning relies more on multi-step process RFT training than high-quality external 3D inputs (\eg, depth map) from spatial encoder equipped in SFT.
Specifically, RFT without spatial encoder, initialized from SFT, slightly underperforms RFT with spatial encoder but largely exceeds SFT.
Moreover, RFT model remains robust under noisy geometric input, showing its robustness without the high-quality external features.

\begin{table}[t]
\caption{
Performance on general benchmarks.
Top-1 scores are shown by \textbf{bold} text.
}
\vspace{-2.5mm}
\centering
\scriptsize
\begin{tabular}{l|cccc}
\toprule
Model & $\text{MME}_{test}$  & $\text{MMBench}_{dev}$ & OK-VQA &POPE  \\
\midrule
NVILA-2B~\cite{liu2024nvila} & 1547 & \textbf{78.63} & 64.9 & 81.96 \\
\cellcolor{myblue}{\mname}-2B-SFT & \cellcolor{myblue}\textbf{1751} & \cellcolor{myblue}77.62 & \cellcolor{myblue}\textbf{65.22} & \cellcolor{myblue}\textbf{82.52} \\
\bottomrule
\end{tabular}
\label{tab: general}
\end{table}

\begin{table}[t]
\caption{Performance on RoboTwin 2.0~\cite{chen2025robotwin} hard tasks. We report the success rate (\%) compared to end-to-end VLA and VLM-based models. Gray rows indicate unseen tasks not present in {\dname}. {\mname} demonstrates strong generalization.}
\vspace{-2.5mm}
\centering
\scriptsize
\setlength{\tabcolsep}{1pt}
\begin{tabular}{l|ccccc|cc|c}
\toprule
\multirow{2}{*}{Task} & \multicolumn{5}{c|}{\cellcolor{myred}End-to-End Policy} & \multicolumn{2}{c|}{\cellcolor{mygreen}Vision-Language Model} & \cellcolor{myblue}Ours \\ 
& \cellcolor{myred}ACT & \cellcolor{myred}DP & \cellcolor{myred}DP3 & \cellcolor{myred}RDT & \cellcolor{myred}$\pi_0$ & \cellcolor{mygreen}Qwen3-VL-8B & \cellcolor{mygreen}Gemini-2.5-Pro & \cellcolor{myblue}RoboTracer-2B \\ 
\midrule
Place A2B Left        & 0 & 0 & 2  & 1  & 1  & 0 & 2  & \textbf{84} \\
Move Playingcard Away & 0 & 0 & 3  & 11 & 22 & 0 & 5 & \textbf{94} \\
Click Alarmclock      & 4 & 5 & 14 & 12 & 11 & 0 & 0  & \textbf{79} \\
Place Burger Fries    & 0 & 0 & 18 & 27 & 4  & 0 & 0  & \textbf{99} \\
$\cdots$              &   &   &    &    &    &   &   &             \\
\rowcolor{gray!10}
Place Container Plate & 1 & 0 & 1  & 17 & 45 & 0 & 0  & \textbf{52} \\  
\rowcolor{gray!10}
Stack Blocks Two      & 0 & 0 & 0  & 2  & 1  & 0 & 0  & \textbf{33} \\
\rowcolor{gray!10}
Place Empty Cup       & 0 & 0 & 1  & 7  & 11 & 0 & 0  & \textbf{85} \\
\rowcolor{gray!10}
Place Object Stand    & 0 & 0 & 0  & 5  & 11 & 0 & 0  & \textbf{38} \\

\midrule
\textbf{Seen Avg. Success Rate}   & 0.9 & 0.4 & 3.7 & 6.3 & 6.6 & 0 & 0.7 & \textbf{75.4} \\
\rowcolor{gray!10}
\textbf{Unseen Avg. Success Rate} & 0.1 & 0 & 0.6 & 4.7 & 11.6 & 0 & 0.3 & \textbf{44.4} \\
\textbf{Total Avg. Success Rate}  & 0.6 & 0.2 & 2.5 & 5.7 & 8.6 & 0 & 0.5 & \textbf{64.0} \\
\bottomrule[1pt]
\end{tabular}
\label{tab: robotwin}
\end{table}

\vspace{+1mm}
\noindent \textbf{Accurate geometry refines metric reasoning.}
In \cref{tab: spatial trace}, using more precise geometric cues greatly improves performance, yield up to {$6\%$} absolute gain.
This suggests that explicit, readily available geometry in embodied settings can further improve metric reasoning, rather than relying solely on implicit learning of VLM to understand them via RGB-only input.
Moreover, our model supports more geometry input to enhance itself without retraining, broadening its applicability.
\cref{fig: benchmark} shows the generated traces with different geometric inputs.

\subsection{Public vision-language Benchmarks}
\label{subsec: public vision-language benchmarks}

\noindent \textbf{Joint training preserves common-sense knowledge.}
In \cref{tab: general}, our model performs on par with or slightly surpasses the base model.
This benefit stems from our use of {\dname} for RGB and RGB+$\mathcal{X}$ (See \cref{subsec: vlm}) joint training, enriched by general visual instruction datasets. 

\begin{table}[t]
\caption{ Real-world robot evaluation of spatially and metrically constrained tasks.
MolmoAct is the end-to-end VLA baseline. RoboRefer is the keypoint-based baseline.
%
%
%
}
\vspace{-3mm}
\centering
\tiny
\begin{tabular}{l|cc|c|c}
\bottomrule[1pt]
\multirow{2}{*}{Spatially and Metrically Constrained Tasks}                  & \multicolumn{3}{c}{Success Rate(\%) $\uparrow$} \\
                                        & MolmoAct~\cite{lee2025molmoact} & RoboRefer~\cite{zhou2025roborefer} & MolmoAct (finetuned) & \textbf{Ours} \\
\midrule
Pick the rightmost hamburger and place it on the &  \multirow{2}{*}{0.00}  & \multirow{2}{*}{0.00} &\multirow{2}{*}{40.00}  & \multirow{2}{*}{\textbf{60.00}} \\      
keyboard in front of the laptop without collisions  &   &  &  \\     
\hline
Water flowers from right to left with watering   & \multirow{2}{*}{0.00} & \multirow{2}{*}{0.00} &\multirow{2}{*}{10.00}  & \multirow{2}{*}{\textbf{30.00}} \\      
can hovering 1-5 cm above each flower.  &  &                  &                  \\

\bottomrule[1pt]
\end{tabular}
\label{tab: real world}

\end{table}

\begin{figure*}[t]
\centering
\includegraphics[width=\linewidth]{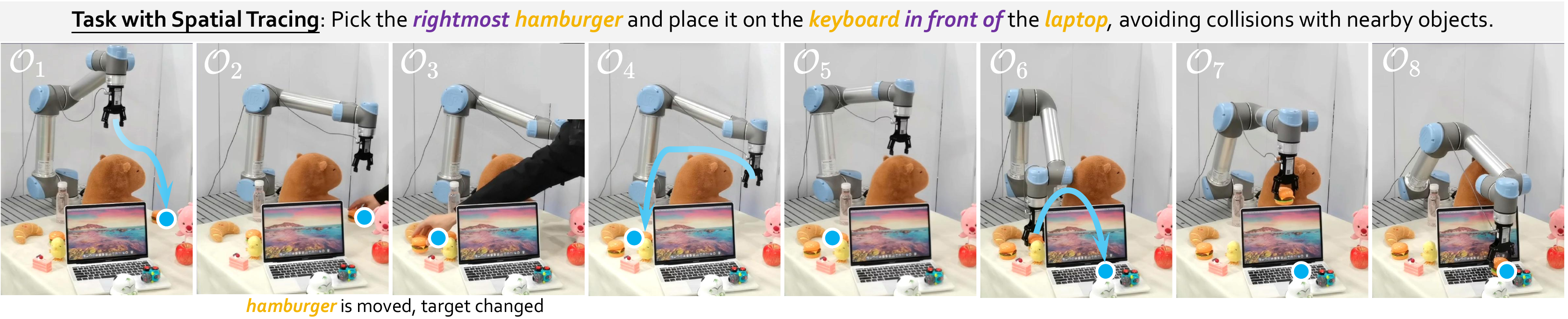}
\vspace{-7mm}
   \caption{Real-World Evaluation. The blue trace denotes predicted spatial trace in 2D, and the blue dot marks current target.
   {\mname} can generate spatial traces whose start and end points all satisfy spatial constraints in cluttered and dynamic scenes.
   %
   %
   }
\label{fig: real_exp}

\end{figure*}

\subsection{Simulator and Real-world Evaluation}
\label{subsec: sim and real world}

\noindent \textbf{Spatial trace offers superior zero-shot generalization.}
In \cref{tab: robotwin}, we evaluate our model on $19$ Robotwin 2.0~\cite{chen2025robotwin} hard tasks involving clustered scenes.
%
Among them, $12$ tasks are seen and $7$ are unseen in the {\dname} dataset.
%
Unlike task-specific end-to-end VLA baselines trained on each task, our model, without any task-specific training, outperforms the best baseline by $32.8\%$ in unseen tasks, showing strong generalization.
%
Moreover, current powerful VLMs struggle to generate spatial traces, highlighting the importance of our model design and dataset contribution.
Please check Supp.~\ref{suppsubsec: simulation evaluation} for details.

\vspace{+1mm}
\noindent \textbf{Spatial tracing is critical for spatially and metrically constrained tasks.}
In \cref{tab: real world}, only our method can handle long-horizon tasks requiring multi-step, metric-grounded spatial tracing in cluttered, dynamic scenes compared to VLA-based (\eg, MolmoAct~\cite{lee2025molmoact}) and keypoint-based (\eg, RoboRefer~\cite{zhou2025roborefer}) method. 
These tasks demand precise identification and placement of objects under evolving spatial constraints, while continuously avoiding collisions.
In \cref{fig: real_exp}, integrating {\mname} with motion planning and Code-as-Monitor~\cite{zhou2024code} enables rapid updates at 1.5 Hz.
Thus, when the rightmost hamburger is moved, the robot adapts by grasping the newly rightmost one, reaching over the large doll and the computer screen to place it on the keyboard.
Notably, this embodiment-agnostic spatial trace can also be executed by G1 humanoid, enabling even more complex, long-horizon tasks such as flower-watering (\cref{fig: motivation}).
See Supp.~\ref{suppsubsec: real-world evaluation} for details.

\vspace{+1mm}
\noindent \textbf{RoboTracer serves as a practical tool to generate action data.}
{\mname} equipped with motion planning and Code-as-Monitor~\cite{zhou2024code} can generate action data for spatially and metrically tasks.
In \cref{tab: real world}, after finetuning with successful episode and employing generated spatial traces as auxiliary supervision, we improve MolmoAct’s 3D reasoning and spatial performance (success rate raised from $0\%$ to $25\%$), showing {\mname}'s application for data generation.



\subsection{Ablation Study}
\label{subsec: ablation}


\noindent \textbf{Data recipe is crucial for SFT training.}
\cref{tab: ablation} shows that combining 2D, 3D, and video data yields optimal performance.
2D/3D data offer critical scale supervision for both indoor and outdoor scenes; their absence degrades metric-grounded Q-spatial~\cite{liao2024reasoningpathsreferenceobjects} accuracy. 
Video data enables gripper motion learning across diverse robot configurations, improving end-effector-centric performance on ShareRobotBench~\cite{ji2025robobrain}.
Moreover, 3D data and simulated videos increase spatial instruction diversity and metric-grounded reasoning, crucial for {\bname}. 
This three-way data synergy is thus key to SFT training.

\vspace{+1mm}
\noindent \textbf{Dataset construction pipeline has scalability.}
Our dataset pipeline scales easily to RGB-only datasets in new domains, enabling automatic multi-step reasoning annotations.
We re-annotate {\dname} via 2D web image pipeline and a gripper detector~\cite{niu2024llarva} without ground-truth metadata.
\cref{tab: spatial trace} shows comparable performance of newly RFT-trained model, demonstrating its scalability.

\vspace{+1mm}
\noindent \textbf{Universal spatial encoder improves 3D reasoning.}
We fine-tune NVILA-2B~\cite{liu2024nvila} on {\dname} without universal spatial encoder, followed by RFT.
In \cref{tab: ablation} (ID D \& I), we find that the spatial encoder enhances metric-grounded reasoning and greatly improves multi-step spatial tracing.
This is mainly due to:
\textbf{(1)} encoder's prior 3D knowledge that facilitates implicit 3D learning, 
%
%
\textbf{(2)} cumulative reasoning across steps, amplifying the utility of spatial cues.

\begin{table}[t]
\caption{Ablation Study on data recipe, spatial encoder, scale supervision  with regression (Reg.) and next-token prediction (N.T.P) loss, point formulation in Q-Spatial, ShareRobotBench(S.R.B.) and {\bname}(T.S.B). 
%
%
We report success rate (SR)/Discrete Fréchet Distance (DFD) of 2B model.
%
%
}
\vspace{-2mm}
\scriptsize
\centering
\begin{tabular}{l|ccc|c|c|c|c|ccc}
\bottomrule[1pt]
\multirow{2}{*}{ID} & \multicolumn{3}{c|}{Data Recipe} & Spatial & Scale  & Point  & Model & Q-Spatial  &S.R.B & T.S.B.\\   
\cmidrule(lr){2-4}\cmidrule(lr){9-11}
& 2D & 3D & Video & Encoder & Supervision & Formulation & Type &   SR(\%) $\uparrow$ & DFD $\downarrow$ & SR (\%) $\uparrow$  \\
\hline
A & \ding{55} & \checkmark & \checkmark & \checkmark & Reg. & $(u, v, d)$ & SFT & 51.49 & 0.1723  & 27\\
B & \checkmark & \ding{55} & \checkmark & \checkmark & Reg. & $(u, v, d)$ & SFT & 33.52 & 0.1809  & 19\\ 
C & \checkmark & \checkmark & \ding{55} & \checkmark & Reg. & $(u, v, d)$ & SFT & 63.40 & 0.4376  & 24\\
\midrule
D & \checkmark & \checkmark & \checkmark & \ding{55} & Reg. & $(u, v, d)$ & RFT & - & - & 36\\
\midrule
E & \checkmark & \checkmark & \checkmark & \checkmark & \ding{55} & $(u, v, d)$ & SFT & 53.47 & 0.1693 & 24\\
F & \checkmark & \checkmark & \checkmark & \checkmark & N.T.P. & $(u, v, d)$ & SFT & 57.43 & 0.1671 & 26\\
\midrule
G & \checkmark & \checkmark & \checkmark & \checkmark & Reg. & $(x, y, z)$ & SFT & 68.32  & 0.2426 & 30\\
\midrule
\cellcolor{myblue}H & \cellcolor{myblue} \checkmark & \cellcolor{myblue} \checkmark & \cellcolor{myblue} \checkmark & \cellcolor{myblue} \checkmark& \cellcolor{myblue}Reg. & \cellcolor{myblue}$(u, v, d)$ & \cellcolor{myblue} SFT & \cellcolor{myblue}\textbf{69.61} & \cellcolor{myblue}\textbf{0.1605} & \cellcolor{myblue}31\\

\cellcolor{myblue} I & \cellcolor{myblue} \checkmark & \cellcolor{myblue} \checkmark & \cellcolor{myblue} \checkmark & \cellcolor{myblue} \checkmark& \cellcolor{myblue}Reg. & \cellcolor{myblue}$(u, v, d)$ & \cellcolor{myblue} RFT & \cellcolor{myblue}- & \cellcolor{myblue}- & \cellcolor{myblue}\textbf{39}\\

\bottomrule[1pt]
\end{tabular}
\label{tab: ablation}
\vspace{-3mm}
\end{table}

\vspace{+1mm}
\noindent \textbf{Scale regression supervision boosts metric awareness.}
In \cref{tab: ablation} (ID E \& F \& H), we compare supervising the metric scale factor using regression loss, next-token prediction loss, and no supervision.
The regression-based model performs best.
While textual (next-token) supervision of offers slight gains, it remains inferior to regression.
We observe that: 
\textbf{(1)} pure next-token prediction supervision—whether or not the scale factor is included as output—demands extensive data to improve scale awareness as shown in recent work~\cite{cai2025depthlm};
\textbf{(2)} explicitly predicting the scale factor, particularly for RGB-only inputs (as in our RGB/RGB+$\mathcal{X}$ mixed training), forces models to learn scale information without relying on auxiliary geometry, thereby bolstering capability.

\vspace{+1mm}
\noindent \textbf{Metric-sensitive reward advances the accuracy.}
%
In \cref{tab: spatial trace}, integrating process rewards boosts overall success rate by $4\%$ compared to purely outcome-based methods.
Notably, using only these metric-agnostic outcome-based rewards yields smaller improvements on 3D metrics and overall success rates compared to the gains observed on 2D metrics, relative to SFT.
%
%
This highlights that leveraging {\dname}’s key step annotations to supervise metric-grounded step-wise perception enables more accurate traces in complex spatial relations.


%

%


\section{Conclusion}
\label{sec: conclusion}


In this paper, we propose {\mname}, a novel 3D-aware VLM that addresses spatial tracing via multi-step, metric-grounded reasoning on both 3D spatial referring and measuring. 
In detail, we empower the model to flexibly process arbitrary geometric inputs for precision, employ scale supervision to enhance scale awareness, and guide it with metric-sensitive rewards to improve its reasoning.
%
%
We also present {\dname}, a large-scale well-designed dataset for SFT and RFT training, with {\bname}, a benchmark tailored to evaluate spatial tracing.
Extensive experiments show the effectiveness of {\mname} and highlight its potential for a broad range of robotic applications.

\vspace{+1mm}
\noindent\textbf{Limitation and Future Work.}
Despite the strong performance of our 3D spatial-trace prediction, we rely on motion planning to recover the 6D end-effector pose for robotic control, which is less effective for rotation-rich manipulation where orientation constraints are critical.
Dense spatial traces may handle such tasks, which is a potential area for further research to explore.

\section*{Acknowledgement}
This work was supported by the National Natural Science
Foundation of China (62132001, 62476011), the Beijing Natural Science Foundation (L252218, L252060), and the Fundamental Research Funds for the Central Universities.
%
%


\clearpage
\setcounter{page}{1}

\appendix
\setcounter{section}{0}

\begin{center}
    {\Large \bfseries Supplementary Material of {\mname}}
\end{center}

\noindent The supplementary document is organized as follows:
\newline
\begin{itemize}

    \item Sec.~\ref{suppsec: more discussion}: More Discussions like Technical Details, Evaluation.
    \newline
    \item Sec.~\ref{suppsec: dataset}: Implementation Details of {\dname}, including data filtering, collection, and QA generation. 
    \newline
    \item Sec.~\ref{suppsec: implementation_of_benchmark}: Implementation details of {\bname}, including benchmark annotation, metrics, and statistics.
    \newline
    \item Sec.~\ref{suppsec: implementation details}: Implementation details of {\mname}, including architecture and training details of each stage.
    \newline
    \item Sec.~\ref{suppsec: experimental setting and details}: More Details on experimental settings, including VQA benchmarks, simulation, and real-world evaluation.
    \newline
    \item Sec.~\ref{suppsec: more demonstrations}: More Demonstrations of {\mname}.
    \newline
    \item Sec.~\ref{suppsec: limitation}: More Discussion on Limitations and Future Work.
\end{itemize}

\section{More Discussions}
\label{suppsec: more discussion}

\vspace{+1mm}
\noindent\textbf{{\dname} Benefit.}
In Tab.~\ref{tab: rebuttal vqa} (ID ABCD), finetuning the same base model on our dataset surpasses specialized models on SpatialRGPT-Bench/SAT, even without benchmark-specific tuning, showing the effectiveness of data.

\vspace{+1mm}
\noindent\textbf{Larger {\bname} Evaluation and Risk of Benchmark-Model Alignment.}
In Tab.~\ref{tab: rebuttal spatial trace}, two checks rule out benchmark-model alignment.
\textbf{(1)} Independent annotators relabeled all {\bname} samples, yet our model still surpasses the baselines. 
\textbf{(2)} Scaling the benchmark to 800 samples under the same relabeled protocol further increases our margin over Gemini-2.5-Pro.

\vspace{+1mm}
\noindent\textbf{More Real-world Evaluation.}
In Tab.~\ref{tab: rebuttal real world evaluation}, we add 3 real-robot tasks.
Each task uses 20 trials with human-annotated success.
MolmoAct is fine-tuned using 50 successful rollouts.
%

\vspace{+1mm}
\noindent\textbf{8B-RFT Results.}
In Tab.~\ref{tab: rebuttal spatial trace}, RFT yields a larger gain on the 8B model than on the 2B model ($12\%$ vs. $8\%$), further showing both its effectiveness and scalability.

\vspace{+1mm}
\noindent\textbf{More Details about Metric-grounded Outdoor Data.}
Outdoor data constitutes $30.23\%$ of the metric-grounded data.
Tab~\ref{tab: rebuttal real world evaluation} shows that the outdoor part has a negligible effect on tabletop manipulation.

\begin{table}[t]
\caption{
{\bname}' results w/ only RGB as model input.
%
%
%
%
}
\vspace{-3mm}
\centering
\tiny
\setlength{\tabcolsep}{3pt}
\begin{tabular}{l|l|l|l|l|c|c|c}
\toprule
ID & Model & Input  & Lift 3D & Training & 100-Original & 100-Relabeled & 800-Relabeled\\
\midrule
A & Gemini=2.5-Pro & RGB & - & - & 3 & 2 & 3.375\\

\midrule

B & Two-stage baseline 1 & RGB & Model-lift & SFT+RFT & 31 & 30 & 34.125\\
C & Two-stage baseline 2 & RGB & Oracle-lift & SFT+RFT & 34 & 33 & 35.625\\
\midrule
D & {\mname}-2B  & RGB & Native & SFT & 31 & 30 & 33.250\\
E & {\mname}-2B  & RGB & Native & SFT+RFT & 39 & 39 & 41.750\\
\midrule
F & {\mname}-8B  & RGB & Native & SFT & 42 & 40 & 44.375\\
G & {\mname}-8B  & RGB & Native & SFT+RFT & \textbf{54} & \textbf{53} & \textbf{57.125}\\

\bottomrule
\end{tabular}
\label{tab: rebuttal spatial trace}
\end{table}

\noindent\textbf{Lift 3D Comparison.}
Tab.~\ref{tab: rebuttal spatial trace} (ID BCE) compares ours (lift label to 3D) with two baselines (lift output to 3D). 
\textbf{(1)} This baseline predicts $(u,v)$ via the first model and lift trace with depth $d$ via the second model, both of which use the same data, model backbone, and training schedule as ours.
\textbf{(2)} We replace the second model with oracle-lift.
Our model surpasses them, showing the superiority of 3D-native formulation.
Overall, under same training data, lifting to $(u,v,d)$ 3D label (39\%) $>$ direct output lifting with oracle-lift (34\%) $>$  lifting to $(x,y,z)$ 3D label (30\%). %

\vspace{+1mm}
\noindent\textbf{Formulation Generality.}
Our $(u,v,d)$ naturally supports navigation: $(u,v)$ offers explicit 2D object referring or waypoints, coupled with SLAM for downstream navigation, as in recent work (\eg, RoboPoint~\cite{yuan2024robopoint}).

\vspace{+1mm}
\noindent\textbf{Compared to RoboRefer~\cite{zhou2025roborefer}.}
\textbf{(1)} Task setting:
We address a more challenging 3D spatial tracing task: predicting a 3D trace rather than a 2D keypoint.
We formulate tracing in $(u,v,d)$ space, extending beyond RoboRefer’s 2D keypoint setting.
\textbf{(2)} Metric features.
3D tracing requires explicit metric reasoning. Our framework incorporates metric awareness at every stage—input, output, supervision, and training—beyond RoboRefer’s qualitative spatial referring.
\textbf{(3)} Temporal modeling.
Our trace data is task-level and temporally structured.
Using manipulation and simulation videos, we collect full-task-execution 3D keypoint sequences that capture intermediate dynamics and temporal consistency.
RoboRefer’s single-image, single-point feature cannot support such 3D tracing.

\vspace{+1mm}
\noindent\textbf{Stronger Evaluation of spatial reasoning.}
We add SpatialRGPT-Bench/SAT results in Tab.~\ref{tab: rebuttal vqa} (ID ABE), where we still surpass specialized models.

\begin{table}[t]
\caption{
Performance on VQA of SpatialRGPT-Bench and SAT. 
}
\vspace{-3mm}
\centering
\tiny
\setlength{\tabcolsep}{4pt}
\begin{tabular}{l|l|l|l|c|c}
\toprule
ID & Model & Base Model & Dataset & SpatialRGPT-Bench & SAT\\


\midrule
A & SpatialRGPT-8B & VILA-1.5-8B & OSD   &  92.69 & - \\
B & SAT-7B & LLaVA-Video-7B & SAT  & -  & 63.4 \\
\midrule
C & Our finetuned VILA-8B & VILA-1.5-8B & {\dname}   & 94.82 & - \\
D & Our finetuned LLaVA-Video-7B & LLaVA-Video-7B & {\dname}  & -  & 82.00\\
\midrule
E & {\mname}-8B & NVILA-8B & {\dname}   & \textbf{97.41} & \textbf{86.67} \\

\bottomrule
\end{tabular}
\label{tab: rebuttal vqa}
\end{table}

\begin{table}[t]
\caption{
More evaluation. F. O. means Finetuned \& Outdoor data.
}
\vspace{-3mm}
\centering
\tiny
\setlength{\tabcolsep}{1.5pt}
\begin{tabular}{l|c|c|c|c}
\toprule
Spatially and Metrically Constrained Tasks & MolmoAct & MolmoAct (F.) & Ours (w/o O.) & Ours\\
\midrule
Place the hot dog \textbf{0.1 m} to the \underline{right} of the \underline{leftmost} peach. & 0 & 40 & 55 & \textbf{55}\\
\midrule
Place the \underline{nearest} blue object to the \underline{right} of the \underline{tallest} object, & \multirow{2}{*}{0} & \multirow{2}{*}{30} & \multirow{2}{*}{50}& \multirow{2}{*}{\textbf{50}}\\
\textbf{5 cm} \underline{above} the bowl center and release it. &  & & &  \\
\midrule
Move the green block \underline{behind} the \underline{tallest} central obstacle and & \multirow{2}{*}{0} & \multirow{2}{*}{15} & \multirow{2}{*}{30} & \multirow{2}{*}{\textbf{35}}\\
\textbf{5 cm} to the \underline{left} of the yellow block via a \textbf{collision-free} motion. &  & & & \\
\bottomrule
\end{tabular}
\label{tab: rebuttal real world evaluation}
\end{table}

\section{{\dname} Details}
\label{suppsec: dataset}

In this section, we provide a detailed overview of the implementation procedures and representative data samples, highlighting the construction of the \dname{} dataset.
The dataset is specifically designed to empower general VLMs with a step-by-step capacity to: 
\textbf{(1)} adapt to \textit{3D spatial referring and measuring} tasks, and 
\textbf{(2)} subsequently progress toward \textit{spatial tracing} tasks in a bottom-up manner. 
To achieve this, we meticulously establish a multi-data-source generation pipeline. 
In the following, we describe the three fundamental components of this pipeline in detail:

\begin{itemize}
    \item \textbf{2D Web Image~(Supp.~\ref{suppsubsec: 2D web image})}: We present a 2D data pipeline comprising image filtering, pseudo-3D scene graph construction, hierarchical referential description generation—from coarse categories to fine-grained spatial referents—and diverse QA pair creation. 
    Since part of our pipeline builds on prior great work~\cite{zhou2025roborefer}, we focus here on the key modifications we introduce and the motivations behind them.

    \item \textbf{3D Scanning Data~(Supp.~\ref{suppsubsec: 3D scanning data})}: This section outlines the 3D data selection process from CA-1M~\cite{lazarow2024cubify} and ScanNet~\cite{dai2017scannet} and presents methods for enriched scene graph construction compared to the 2D data source. We further describe a QA generation framework that leverages detailed 3D annotations (\eg, depth maps, oriented 3D bounding boxes) to capture richer spatial relations for \textit{3D spatial referring and measuring}. 

    \item \textbf{Object-centric Spatial Tracing from 3D scanning data~(Supp.~\ref{suppsubsec: object_centric_tracing})}:
    To bridge the gap between static 3D scanning scenes and manipulation, we introduce a simulation-based generation pipeline. This section details our hierarchical framework (\eg, RRT* with escape mechanisms), the taxonomy of five manipulation primitives including active obstacle bypass, and the rigorous spatial trace refinement process. We further explain how we synthesize metric-grounded tracing data with diverse types (\eg, 2D, 3D, and Lifting 2D to 3D).

    \item \textbf{Manipulation Video~(Supp.~\ref{suppsubsec: manipulation video})}: We describe how to construct end-effector-centric spatial tracing data, including cleaning real-world manipulation video datasets and generating large-scale data in simulation. This enables data collection across three different robot embodiments and both single-arm and dual-arm configurations.
\end{itemize}


\subsection{2D Web Image}
\label{suppsubsec: 2D web image}

2D Web Images aim to provide basic spatial concepts and broad-scale perception across indoor and outdoor scenes. Here we use OpenImage\footnote{\href{https://storage.googleapis.com/openimages/web/index.html}{https://storage.googleapis.com/openimages/web/index.html}}~\cite{kuznetsova2020open} as 2D data source.

\subsubsection{Dataset Construction}

Inspired by recent notable work~\cite{zhou2025roborefer, cheng2024spatialrgpt, chen2024spatialvlm}, we adopt a multi-step pipeline for data cleaning and construction: \textbf{(1)} multi-stage image filtering, \textbf{(2)} pseudo-3D scene graph construction, \textbf{(3)} hierarchical object description generation, and \textbf{(4) }diverse QA pair generation based on the scene graphs. 
While our overall approach aligns with prior work~\cite{zhou2025roborefer}, we further refine Step 2 by developing a more accurate method for constructing pseudo-3D scene graphs.

\subsubsection{Multi-Stage Image Filtering.}

The OpenImages~\cite{kuznetsova2020open} dataset covers 1.7M training images with extensive visual diversity. 
However, a large portion (\eg, text-only graphics, QR codes, medical scans, abstract art) is not well-suited for spatial reasoning, especially for \textit{3D spatial referring and measuring}. 
To curate a subset amenable to referential and reasoning tasks, we adopt a two-stage filtering approach.
We detail below.

\vspace{+1mm}
\noindent \textbf{(1) Stage 1: Coarse Filtering.}
\label{par: siglip2_filtering_stage}
We quickly eliminate off-theme or low-quality imagery lacking multiple everyday objects by using the ``siglip2-giant-opt-patch16-384'' model~\cite{tschannen2025siglip}. 
We define positive labels (desired content) and negative labels (undesired content), then compute the cosine similarity between each image embedding and all label embeddings. 
Images closest to positive labels are retained; otherwise, they are discarded. 
Iterative refinement of label sets balances recall and precision. 
From the original 1.7M images, 934k remain after this step (see Listings~\ref{lst:siglip_positive_labels} and~\ref{lst:siglip_negative_labels} for label details).

\begin{lstlisting}[basicstyle=\ttfamily\footnotesize, backgroundcolor=\color{myblue!50}, caption={Positive Labels used during SigLIP2 filtering.}, captionpos=t, breaklines=true, breakatwhitespace=true, columns=fullflexible, label={lst:siglip_positive_labels}]
Positive Labels = [
    "Mid-distance observation of some objects on a table",
    "Some objects on the desktop",
    "Distant view of some animals",
    "Mid-distance observation of some animals",
    "Distant view of one object",
    "Mid-distance observation of one object",
    "Distant view of some objects",
    "Mid-distance observation of some objects",
    "Distant view of a person",
    "Mid-distance observation of a person",
    "Distant view of some people",
    "Mid-distance observation of some people",
    "Distant view of indoor scene",
    "Distant view of outdoor scene",
    "Distant view of traffic",
    "Distant view of Urban architecture"
]
\end{lstlisting}

\begin{lstlisting}[basicstyle=\ttfamily\footnotesize, backgroundcolor=\color{myblue!50}, caption={Negative Labels used during SigLIP2 filtering.}, captionpos=t, breaklines=true, breakatwhitespace=true, columns=fullflexible, label={lst:siglip_negative_labels}]
Negative Labels = [
    "Macro shot of an animal",
    "Macro shot of one object",
    "Macro shot of a person",
    "Macro shot of flowers",
    "A piece of text",
    "A person displayed in front of a white background",
    "A product displayed in front of a white background",
    "A screenshot of the graphics user interface",
    "A dimly lit environment"
]
\end{lstlisting}

\noindent SigLIP2 preserves images enriched with diverse objects, robust depth cues, and contextual variety (indoor/outdoor) through the aforementioned labeling. However, it struggles with certain image categories: 
\textbf{(1)} paintings/artworks with visible brushstrokes or textures, 
\textbf{(2)} low-light scenes with minimal illumination or heavy shadows, 
\textbf{(3)} grayscale photographs lacking color cues, 
\textbf{(4)} distorted images exhibiting geometric anomalies, and 
\textbf{(5)} multi-scene collages containing three or more distinct segments. 
These types hinder reliable detection and interpretation, underscoring the need for a secondary, fine-grained filtering stage.

\vspace{+1mm}
\noindent \textbf{Stage 2: Fine-grained Filtering}
\label{par: qwen_vl_filtering_stage}
Due to SigLIP2’s limited capacity for certain visual content, we introduce a fine-grained filtering stage using the Qwen2.5-VL-7B~\cite{bai2025qwen2} model to improve dataset quality. 
This ensures that remaining images are clear, authentic, and suitable for the spatial understanding and reasoning essential to \textit{3D spatial referring and measuring} tasks. 
In total, Qwen2.5-VL processed 934k images pre-filtered by SigLIP2, retaining 846k. While Qwen2.5-VL offers superior precision, its slower speed necessitated SigLIP2 for rapid initial filtering, thereby increasing overall efficiency.
For accurate and consistent fine-grained filtering, we employ a structured prompt engineering approach. 
A system prompt (see Listing~\ref{lst:qwen_system_prompt}) designates Qwen2.5-VL as an image analysis expert, specifying key visual attributes to verify and negative categories to detect, with a strict workflow. For each image, a user prompt (see Listing~\ref{lst:qwen_user_prompt}) instructs the model to determine whether it falls into any predefined negative category. 
The model’s response follows a fixed format: if the text after the pipe symbol (|) is ''Yes'', the image is discarded; otherwise, it is retained. 
This scheme enforces a consistent output format and enhances the reliability of filtering.

\begin{lstlisting}[basicstyle=\ttfamily\footnotesize, backgroundcolor=\color{myblue!50}, caption={System Prompt for Qwen2.5-VL-7B filtering.}, captionpos=t, breaklines=true, label={lst:qwen_system_prompt}]
system_prompt = """
You are an image analysis expert. Follow this workflow rigidly:

1. **Content Analysis**:
   - Inspect: Main subjects, artistic style, visual characteristics
   - Check: Lighting intensity, color channels, geometric integrity, composition structure

2. **Category Verification** (YES if matches ANY):
   a) Painting/Artwork - Visible brushstrokes/canvas texture
   b) Dim Lighting - Very low brightness, heavy shadows
   c) B&W Photo - Grayscale only (0 color channels)
   d) Distorted Image - Warping/mirroring anomalies
   e) Multi-image Collage - >=3 distinct scenes with hard borders

3. **Structured Response**:
   Output EXACTLY in this format:
   "[Analysis sentence]. | Yes/No"
   - Analysis must contain observable evidence
   - Final answer MUST use pipe separator

Examples of VALID responses:
    "This image is a composite created by stitching together multiple smaller images, with distinct white borders visible between the individual components. | Yes"
    "This image features vibrant colors, is neither an artistic painting nor a composite of multiple images, and does not conform to any of the specified categories. | No"
"""
\end{lstlisting}

\begin{lstlisting}[basicstyle=\ttfamily\footnotesize, backgroundcolor=\color{myblue!50}, caption={User Prompt for Qwen2.5-VL-7B filtering.}, captionpos=t, breaklines=true, label={lst:qwen_user_prompt}]
user_prompt = """
Analyze if this image belongs to ANY of these categories:
1. Painting/artwork
2. Dim lighting
3. Black-and-white
4. Geometric distortion
5. Multi-image collage

Respond EXACTLY FORMATTED as:
"[Your evidence-based analysis]. | Yes/No"
"""
\end{lstlisting}

\begin{figure*}[t]
  \centering
  \includegraphics[width=\textwidth]{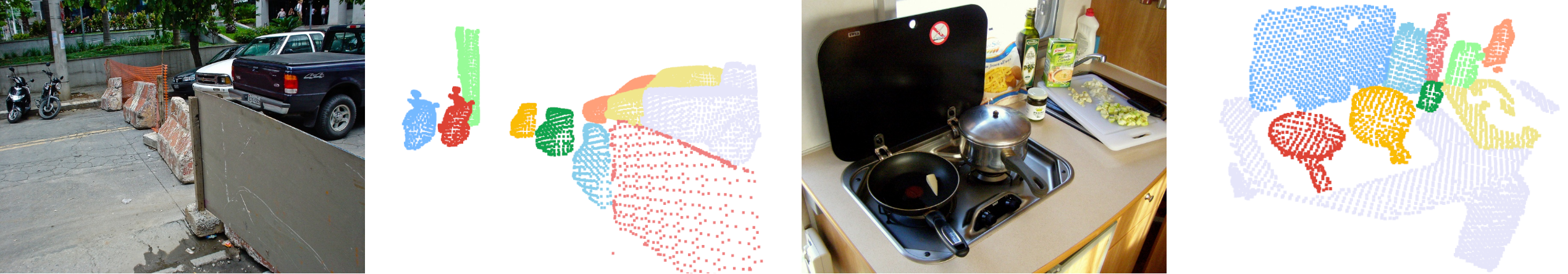}
  \caption{The visualization of pseudo-3D scene graphs generated from 2D images with detected objects, and corresponding point clouds.}
  \label{suppfig: scene_graph_visualization}
\end{figure*}

\subsubsection{Pseudo-3D Scene Graphs Construction}
\label{suppsubsubsec: 3D Scene Graph Construction}

Although 2D images provide limited spatial cues, deriving rich 3D spatial information (\eg, near vs. far, distances), especially for 3D spatial referring and measuring, directly from these images remains challenging. 
Inspired by prior work~\cite{zhou2025roborefer,cheng2024spatialrgpt, chen2024spatialvlm}, we construct pseudo-3D scene graphs from 2D images to enhance QA pairs with robust 3D spatial semantics. 
In these graphs, nodes represent object attributes, while edges denote inter-object spatial relationships. The following sections detail this conversion process.

\vspace{+1mm}
\noindent \textbf{Object Detection and Annotation.}
\label{par: detection_annotation_groundingdino_ram}
While the OpenImages dataset provides annotations, its limited vocabulary and coarse labeling hamper open-world applications. 
To remedy this, we integrate state-of-the-art foundation models for refined object detection and labeling. 
Specifically, the Recognize Anything Model (RAM++)~\cite{zhang2024recognize} and GroundingDINO~\cite{liu2024grounding} jointly assign semantic labels and bounding boxes to key objects in filtered 2D images.
The pipeline proceeds as follows:

\begin{enumerate}
    \item \textbf{Semantic Labeling via RAM++}: RAM++ analyzes each image to generate comprehensive category labels, ensuring broad semantic coverage.
    \item \textbf{Bounding Box Localization via GroundingDINO}: Label outputs from RAM++ serve as text prompts for GroundingDINO, which localizes objects and produces precise bounding boxes.
\end{enumerate}

\noindent \textbf{3D-aware information Extraction.}
To further extract 3D-aware information from 2D images, we need to convert the 2D images into 3D representations. 
Unlike prior works (\eg, RoboRefer~\cite{zhou2025roborefer}) that rely on multiple foundation models for metric depth estimation (\eg, UniDepth V2~\cite{piccinelli2025unidepthv2}) and camera intrinsics prediction (\eg, WildeCamera~\cite{zhu2023tame}), we adopt MoGe-2~\cite{wang2025moge}, a single accurate monocular geometry reconstruction model with metric scale and sharp details.
Compared to existing methods such as MoGe, DepthPro, UniDepth V2, Metric3D V2, and Depth Anything V2, MoGe-2 provides more accurate relative geometry, precise metric depth, and sharper detail recovery. 
It reconstructs scaled point clouds and estimates camera intrinsics from a single image, enabling robust 3D scene reconstruction.
we adopt  for metric depth estimation due to its recent state-of-the-art performance. 
Based on previously annotated object bounding boxes, we apply SAM 2.1~\cite{ravi2024sam} to generate instance masks. 
Each resulting Pseudo-3D scene graph comprises object labels (via RAM++), 2D bounding boxes (via GroundingDINO), instance masks (via SAM 2.1), and object-level point clouds (via MoGe-2).
Some visualizations are provided in Fig.~\ref{suppfig: scene_graph_visualization}.

\begin{figure*}[t]
  \centering
  \includegraphics[width=\linewidth]{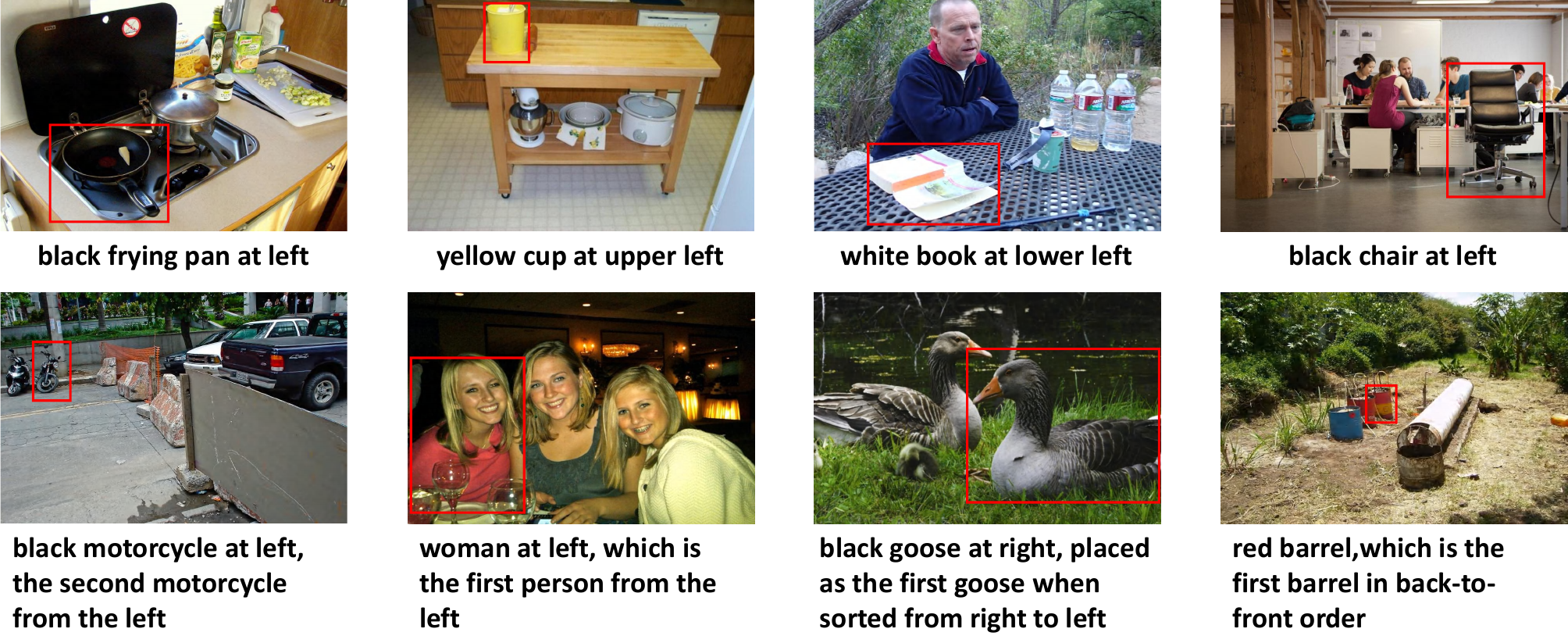} 
  \caption{Generated object descriptions. Top: Unique-category captions. Bottom: Spatially-aware captions for the same categories. Red boxes indicate referenced objects.}
  \label{suppfig: caption_visualization_example}
\end{figure*}

\subsubsection{Hierarchical Object Description Generation}
\label{suppsubsubsec: object_description_generation}
While 3D scene graphs commonly encode broad object categories, real-world scenes frequently contain multiple instances of each category. 
To distinguish among these instances, we augment object descriptions with attributes and spatial relations, yielding fine-grained disambiguation. 
Below, we outline our two-stage generation pipeline.

\vspace{+1mm}
\noindent \textbf{Stage 1: Generating Object Descriptions in image space.}
We employ the Qwen2.5-VL-7B model to generate detailed object- and image-level captions. 
These global captions provide contextual grounding for the QwQ-32B model during LLM QA generation (see Supp.~\ref{par: qwq_diversification_2d}), thereby enhancing relevance and accuracy. 
Prompt templates can be found in Listings~\ref{lst:qwen_image_capt_prompts} and ~\ref{lst:qwen_object_capt_prompts}. 
Notably, the \texttt{object\_caption\_user\_text\_prompt} includes a dynamic placeholder \texttt{[class\_name]}, populated with categories predicted by the RAM++ model (see Supp.~\ref{suppsubsubsec: 3D Scene Graph Construction}).
%

\begin{lstlisting}[basicstyle=\ttfamily\footnotesize, backgroundcolor=\color{myblue!50}, caption={Prompts for Image Caption Generation with Qwen-VL.}, captionpos=t, breaklines=true, label={lst:qwen_image_capt_prompts}, escapeinside={(*@}{@*)}]
image_caption_system_text_prompt =  """
    You are an expert image analysis assistant. Your task is to generate a detailed and comprehensive description of the image.
    Please focus on accurately capturing all visual elements present in the image, including objects, scenery, colors, shapes, textures, and lighting.
    Your description should be clear, precise, and professional. Additionally, ensure that your description begins with either `this image' or `the image'.
"""

image_caption_user_text_prompt = """
    Please carefully examine the provided image and generate a detailed description.
    Include all visible elements such as objects, scenery, colors, shapes, textures, and lighting.
    Ensure that your description is thorough, accurate, and complete, and that it starts with either `this image' or `the image'.
"""
\end{lstlisting}

\begin{lstlisting}[basicstyle=\ttfamily\footnotesize, backgroundcolor=\color{myblue!50}, caption={Prompts for Object Caption Generation with Qwen-VL.}, captionpos=t, breaklines=true, label={lst:qwen_object_capt_prompts}]
object_caption_system_text_prompt = """
    You are a visual localization analyzer working with TWO distinct images:
    1. [POSITION-REFERENCE] (First Image):
    - Contains ONLY location clues with background
    - Strictly use ONLY for determining spatial position (left/right/upper/lower/center)
    - Ignore all visual features except object placement

    2. [DETAIL-SOURCE] (Second Image):
    - Shows the object's TRUE APPEARANCE without background
    - Extract EXCLUSIVELY from this: color, texture, material, shape
    - Never infer details from the first image

    Generate phrase in pattern: [Color][Material][Object] at [Position]
    Example: "Matte black laptop on the left" NOT "Red-boxed laptop"
"""

object_caption_user_text_prompt = """
    For the [class_name] marked by red box in FIRST image and fully shown in SECOND image:
    -> COLOR/MATERIAL: Must come from SECOND image
    -> POSITION: Only from FIRST image's placement
    Forbidden actions:
    x Mention 'red box' or background elements
    x Use location terms in second image
    x Combine features across images

    Describe the [class_name] marked by red box in FIRST image and fully shown in SECOND image with this format:
    [Color][Material/Texture][Object] at [Position]
    Samples:
    - "Brushed metal water at bottle left"
    - "Glossy ceramic mug at upper center"
    - "Faded denim jacket at lower right"
"""
\end{lstlisting}

\begin{figure*}[t]
  \centering
  \includegraphics[width=\linewidth]{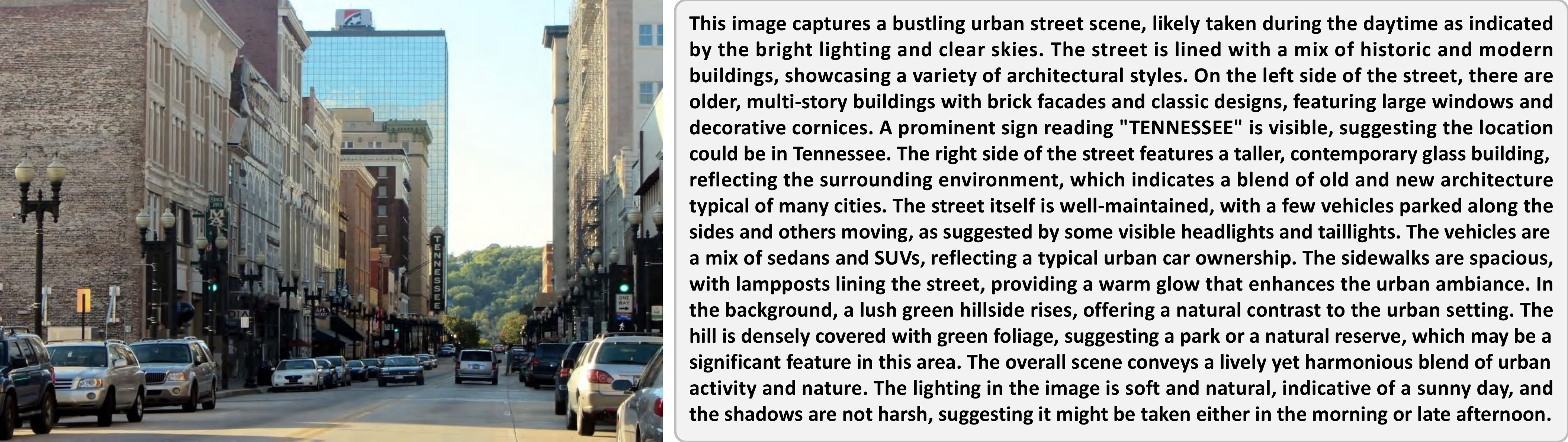}
  \caption{The visualization of generated image detailed descriptions from VLM (\eg, Qwen2.5-VL).}
  \label{suppfig: image_caption_visualization_example}
\end{figure*}

\vspace{+1mm}
\noindent\textbf{Stage 2: Generating Object Description with Spatial Cues.}
\label{par:heuristic_obj_description}
To enhance referential specificity when multiple same-category objects coexist, we adopt a heuristic that appends spatially oriented references (\eg, ``the third chair from the front''). 
This strategy leverages 3D object coordinates from the scene graph (see Supp.~\ref{suppsubsubsec: 3D Scene Graph Construction}). 
By comparing same-category objects along the three principal axes (front–back, left–right, top–bottom), we identify the axis with maximal spatial variation as the primary direction for generating relative spatial descriptions.
After identifying the main sorting axis, we retrieve suitable templates from a predefined library (see Listing~\ref{lst:spatial_order_templates}) to refine the initial object descriptions. 
These templates capture diverse natural language patterns. 
For instance, if a row of chairs is arranged left to right, potential templates include ``\{dense\_caption\}, which is the \{ordinal\} \{class\_name\} from left to right,'' or ``\{dense\_caption\}, the \{ordinal\} \{class\_name\} in the left-to-right sequence''.
Here, \texttt{{dense\_caption}} is the Qwen2.5-VL–generated description, \texttt{ordinal} indicates the object’s position in the sequence, and \texttt{{class\_name}} is the category label output by RAM++. 
This spatially aware enhancement applies only when multiple instances of the same category appear, preventing redundancy; otherwise, the original dense caption is used. 
To ensure spatial diversity, we enforce a variance threshold across the three principal axes, discarding images with multiple same-category instances but low variance. This process yields a final set of $466$k images. 
By integrating spatial ordering with visual descriptions, our heuristic produces precise, discriminative referential expressions, which are essential for generating high-quality, unambiguous question-answer pairs (\eg, \textit{3D spatial referring and measuring}).

\begin{lstlisting}[basicstyle=\ttfamily\footnotesize, backgroundcolor=\color{myblue!50}, caption={Templates for Spatial Order Description Enhancement.}, captionpos=t, breaklines=true, label={lst:spatial_order_templates}]
TEMPLATES = {
    "left_to_right": [
        "{dense_caption}, which is the {ordinal} {class_name} from left to right",
        "{dense_caption}, marked as the {ordinal} {class_name} in a left-to-right arrangement",
    ],
    "right_to_left": [
        "{dense_caption}, the {ordinal} {class_name} viewed from the right",
        "{dense_caption}, the {ordinal} {class_name} from the right",
    ],
    "front_to_back": [
        "{dense_caption}, which appears as the {ordinal} {class_name} when viewed from the front",
        "{dense_caption}, positioned as the {ordinal} {class_name} in front-to-back order",
    ],
    "back_to_front": [
        "{dense_caption}, which is counted as the {ordinal} {class_name}, starting from the back",
        "{dense_caption}, the {ordinal} {class_name} in the back-to-front sequence",
    ],
    "top_to_bottom": [
        "{dense_caption}, the {ordinal} {class_name} viewed from the top",
        "{dense_caption}, placed as the {ordinal} {class_name} when sorted from top to bottom",
    ],
    "bottom_to_top": [
        "{dense_caption}, which ranks as the {ordinal} {class_name} in bottom-to-top order",
        "{dense_caption}, arranged as the {ordinal} {class_name} when ordered from the bottom",
    ]
}
\end{lstlisting}

\vspace{+1mm}
\noindent\textbf{Examples of Object and Image Descriptions}
\label{par:2d_description_visable}
This part qualitatively show representative examples with generated descriptions. 
As shown in Fig.~\ref{suppfig: caption_visualization_example}, we present two types of object captions: the top row displays simple Qwen2.5-VL captions for single-instance categories where spatial ordering is unnecessary, while the bottom row includes captions enriched with spatial information to distinguish multiple instances of the same category.
Additionally, Fig.~\ref{suppfig: image_caption_visualization_example} shows Qwen2.5-VL’s ability to generate detailed global descriptions of entire images used in the following.



\subsubsection{Generating Diverse QA Pairs via 3D Scene Graphs}
\label{suppsubsubsec: qa_generation_2d}

After constructing scene graphs and generating hierarchical object descriptions, we can leverage this information to generate diverse QA pairs from pseudo-3D scene graphs to support SFT training for improved spatial understanding, especially for \textit{3D spatial referring and measuring}.

\vspace{+1mm}
\noindent\textbf{Template, Choice and Fact QA Generation.}
\label{par: template_qa_fact_2d}
We first adopt a template-based approach to generate structured QA pairs, multiple-choice questions, and factual statements. 
These templates are derived from scene graph information (\eg, object attributes for \textit{3D spatial measuring}, 3D positions for \textit{3D spatial referring}) and refined hierarchical object descriptions. 
The spatial concepts addressed by our QA templates encompass the following categories:

\begin{enumerate}
\item \textbf{Relative position relations}: capture spatial layouts (left/right, above/below, front/behind).
\item \textbf{Relative size comparisons}: describe object attributes (\eg, bigger/smaller, taller/shorter, wider/thinner) often inferred from image-plane projections.
\item \textbf{Quantitative information from 2D or pseudo-3D label}: include spatial reasoning based on estimated depth maps, 2D object coordinates, 3D object coordinates (\ie, \textit{3D spatial referring}), coarse depth approximations, and coarse metric estimation for 3D spatial measuring (\eg, object's length, width, height, distance).
\end{enumerate}


    


\noindent We further design fact templates to generate declarative statements, forming a structured basis for prompting Reasoning LLM to produce richer and more natural QA pairs. 
%

\vspace{+1mm}
\noindent\textbf{Reasoning QA Generation.}
\label{par: qwq_diversification_2d}
To produce more diverse, complex, and natural QA pairs beyond templated formats, we employ QwQ-32B~\cite{qwq32b}, a powerful reasoning LLM. 
Given factual statements, initial QA pairs, multiple-choice questions (if available), as well as global image captions and precise object descriptions, QwQ-32B generates more challenging and conversational spatial reasoning QA.

\subsection{3D Scanning Data}
\label{suppsubsec: 3D scanning data}

3D scanning data typically provides richer and more accurate 3D information (\eg, camera geometry, oriented 3D bounding boxes), whereas 2D web images rely entirely on foundation models to infer 3D structure.
\highlight{However, a major challenge in utilizing 3D datasets lies in the lack of comprehensive annotations that support full 2D-3D understanding. This requires, at a minimum, accurate camera parameters (\ie, intrinsic and extrinsic), as well as complete per-object annotations in both 2D (\eg, bounding boxes, masks, categories, spatial captions) and 3D (\eg, oriented 3D bounding boxes).}
While prior work~\cite{zhou2025roborefer} addresses several problems of specific dataset, such as resolving the missing object category labels in CA-1M \cite{lazarow2024cubify}, we note that a key resource, ScanNet~\cite{dai2017scannet}, is omitted, due to its lack of 2D bounding box annotations.
To this end, \highlight{we leverage the well-processed and comprehensive versions of the CA-1M and ScanNet datasets}—the former processed following the approach of prior work~\cite{zhou2025roborefer}, and the latter completed by filling in missing information as detailed below—to construct spatially related QA pairs.
These QA pairs focus on 3D spaital reasoning, especialy for \textit{3D spatial referring and measuring}, and more importantly, on building object-centric, multi-step, metric-grounded \textit{spatial tracing} data enabled by complete 3D information.
In the following, we first detail the comprehensive annotation pipeline for ScanNet, followed by how CA-1M and ScanNet are used to construct the desired data (\eg, 3D spatial referring, measuring and tracing).


\subsubsection{ScanNet Data Processing}

We leverage the ScanNet dataset, utilizing the original 3D bounding boxes, object labels, depth maps, and camera intrinsic and extrinsic parameters provided by EmbodiedScan. 
However, EmbodiedScan does not provide 2D bounding box annotations, which are necessary for our processing pipeline.
Therefore, we implement a procedure to automatically generate these 2D boxes from the 3D data.
Our extraction process is as follows:

\vspace{+1mm}
\noindent \textbf{Step 1: 3D Point Sampling.}
For a given 3D bounding box, we first sample 5,000 points uniformly from its surface.

\vspace{+1mm}
\noindent \textbf{Step 2: Projection and Filtering.}
We first project these 3D points into the 2D image plane using the provided camera parameters. We then retain valid points based on two criteria:
\textbf{(1)} Depth Consistency Check: We compare the depth of the projected point ($Z_{proj}$) with the depth in the corresponding pixel of the depth map ($D_{map}$). A point is retained only if: $D_{map} - 5\text{cm} < Z_{proj} < D_{map} + 5\text{cm}$.
\textbf{(2)} Image Boundary Check: We discard any point that projects to a 2D coordinate $(x, y)$ outside the image boundaries (\ie, $x < 0$, $x \ge \text{width}$, $y < 0$, or $y \ge \text{height}$).

\vspace{+1mm}
\noindent \textbf{Step 3: 2D Bounding Box Generation.}
After filtering, we compute the 2D bounding box by finding the minimum and maximum coordinates ($x_{min}, y_{min}, x_{max}, y_{max}$) from the set of all remaining valid 2D points.

    

    

\begin{figure*}[t]
    \centering
    \includegraphics[width=\linewidth]{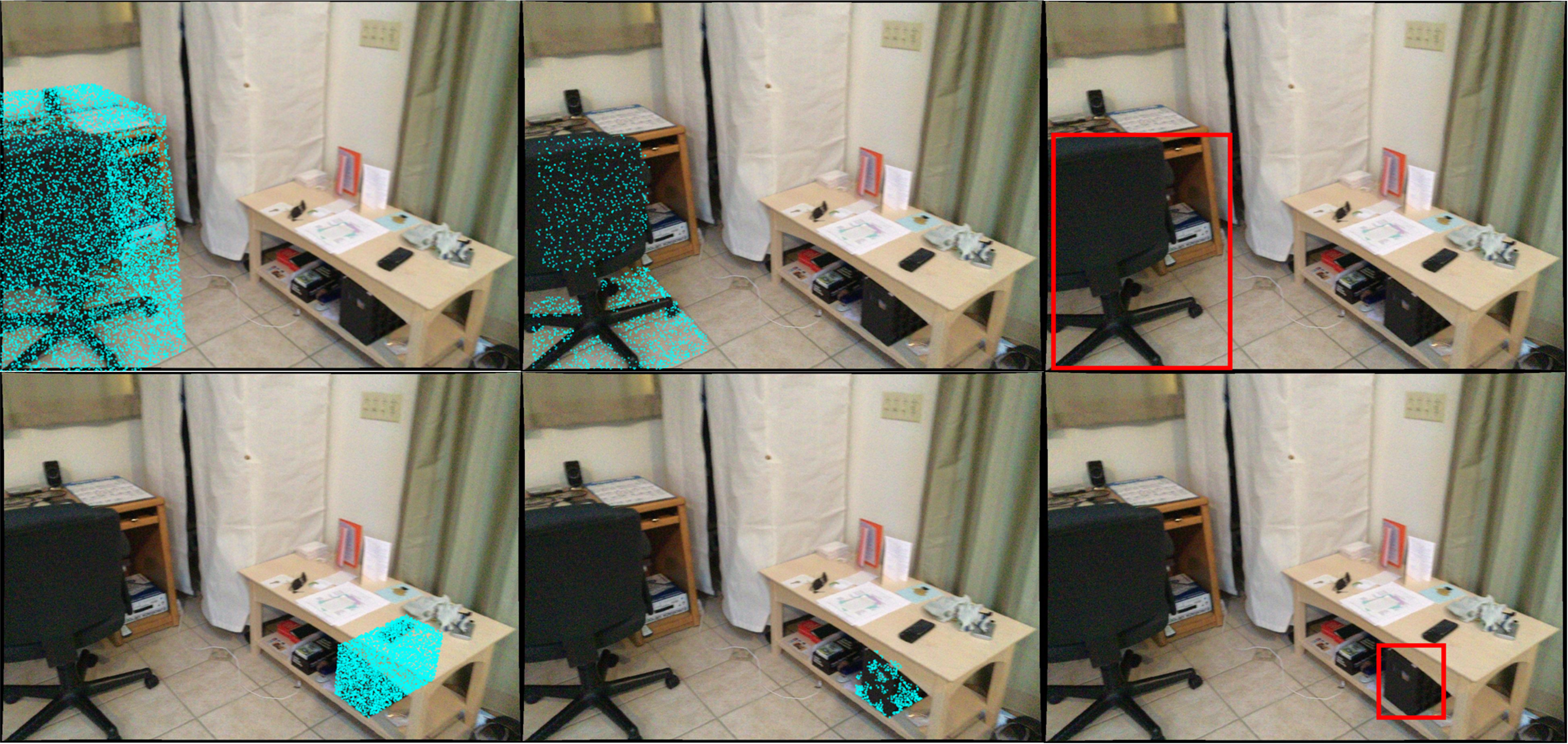}
    \caption{The visualization of the 2D bounding box generation pipeline for the ScanNet dataset.}
    \label{suppfig: scannet_2d_bbox}
\end{figure*}

\vspace{+1mm}
\noindent The visualization of this process is shown in Fig.~\ref{suppfig: scannet_2d_bbox}.

\subsubsection{3D Object Description Generation and Scene Graph Construction for 3D Scanning data}
Following the 2D scene graph pipeline (Supp.~\ref{suppsubsubsec: 3D Scene Graph Construction}) and the object description generation approach (Supp.~\ref{suppsubsubsec: object_description_generation}), we build 3D scene graphs that structurally resemble those in OpenImages (Fig.~\ref{suppfig: scene_graph_visualization}).
Unlike purely 2D counterparts, these 3D graphs emphasize indoor settings and incorporate precise geometric cues (\eg, ground-truth depth, camera intrinsics/extrinsics, 3D oriented bounding boxes). 
\highlight{The enhanced 3D scene graph outperforms 2D-based counterparts in object localization and spatial relation accuracy, enabling structurally rich and quantitatively grounded 3D QA data across $28$ spatial relation types}.
Similar to the 2D pipeline, we construct template-based, choice-based, and fact-based QA pairs from the 3D scene graph. 
Fact QA pairs are further paraphrased and enhanced using a reasoning LLM to generate reasoning questions. 
Moreover, due to the richer and more precise information in 3D scanning data, we additionally introduce methods for constructing fine-grained 3D spatial measuring, referring, and tracing data, as detailed below.

\begin{figure*}[t]
    \centering
    \includegraphics[width=\linewidth]{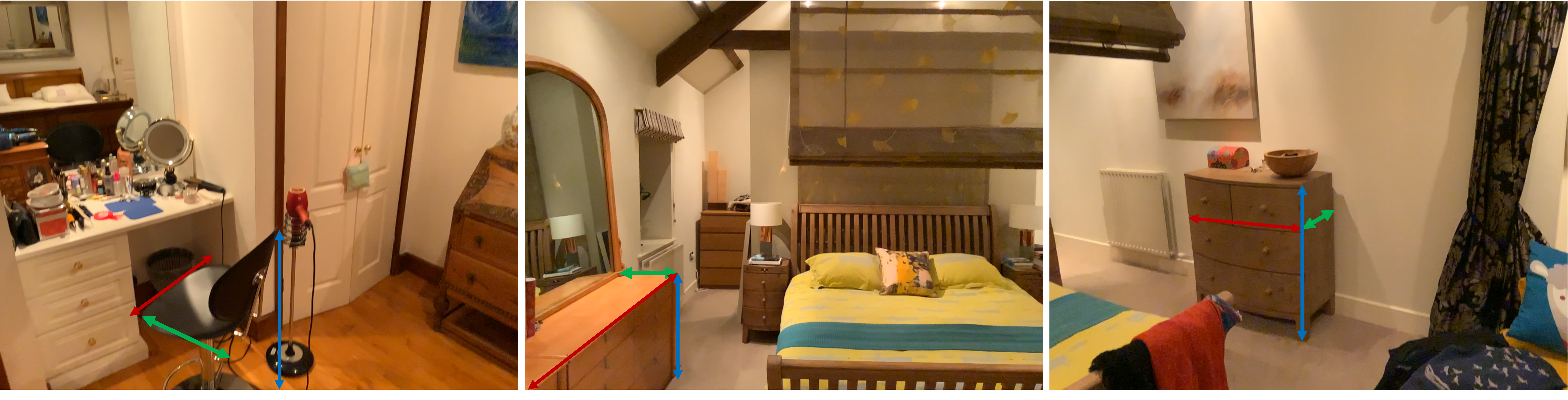} 
    \caption{The visualization of our semantic dimension definitions. \textbf{Length (red)} is defined as the intersection of the front and bottom faces. \textbf{Width (green)} is the intersection of the side and bottom faces. \textbf{Height (blue)} is the intersection of the front and side faces.}
    \label{suppfig: dimension_definition}
\end{figure*}

\subsubsection{Human-Like Measuring Descriptions Generation}
\label{suppsubsubsec: Human-Like Measuring Descriptions Generation}

During the generation of ground-truth text related to distances and dimensional measurements (\ie, length, width, height), \highlight{we make two critical observations regarding model performance: 
(1) we find that strictly constraining the model to output a single, standardized unit (\eg, forcing all values to be in ``meters") negatively impacts its predictive accuracy. 
(2), we observe a similar degradation in performance when the target text omits the name of the object being measured (\eg, providing only the numerical value ``1.5 meters" rather than a contextual phrase like ``the table is 1.5 meters wide").}
Based on these findings, we design a ground-truth generation strategy to create more natural, ``human-like" descriptions. This strategy has two main components:
\begin{enumerate}
    \item \textbf{Contextual Phrasing:} All measurement-related descriptions are formed in a sentence with addition context, \ie, explicitly includes the name of the object and related measurement type (\eg, length, width, height), rather than being a standalone numerical value of the answer.
    For example, for the question ``\textit{What is the height of the cargo ship at left near the port terminal scene?}'', the contextualized answer ``\textit{The height of the cargo ship at left near the port terminal scene is about 43 meters}'' performs better than the raw response ``\textit{43 meters}''.
    
    \item \textbf{Dynamic Unit Selection:} Instead of a fixed unit, we implement a dynamic unit selection mechanism that adapts to the magnitude of the value. This method converts the base metric value into a set of plausible units (\eg, meters, centimeters, feet, inches) and stochastically selects one based on human-like preferences. 
    For example, values less than 1 meter are randomly converted to ``centimeters" (with a higher probability) or ``inches". Values greater than 1 meter are typically expressed in ``meters" but are also stochastically converted to ``feet". This approach ensures the model is exposed to a diverse and realistic range of textual formats, which our experiments show leads to better performance.
\end{enumerate}

\subsubsection{3D Spatial Measuring Data Generation}

To generate the data for 3D spatial measuring, we first define the object dimensions (length, width, height), which is very crucial for 3D scanning dataset, and then extract them.
Herein we mainly focus on CA-1M and ScanNet.

\vspace{+1mm}
\noindent \textbf{Data Normalization Across Datasets. }
A critical issue we notice during data processing is the inconsistent definition of bounding box dimensions across datasets. 
We find that the 3D bounding box scales provided by CA-1M~\cite{lazarow2024cubify} follow a \textbf{[width, height, length]} order. 
In contrast, the ScanNet annotations provided by EmbodiedScan adopt a \textbf{[width, length, height]} order. 
To resolve this, we implement a normalization step to unify all bounding box data into a single, consistent format before extracting measurements.

\vspace{+1mm}
\noindent \textbf{Semantic Dimension Definitions.}
\highlight{A simple extraction of [width, height, length] dimensions from dataset's raw annotations is insufficient, as these are arbitrary and do not align with human concepts of ``length" or ``width". 
We thus establish a consistent, semantically-grounded definition for object dimensions based on their canonical orientation (\eg, their ``front", "``side", and ``bottom" faces).}
Our definitions are as follows:

\begin{itemize}
    \item \textbf{Length:} Defined as the length of the edge where the object's \textbf{front and bottom face} intersect.
    
    \item \textbf{Width:} Defined as the length of the edge where the object's \textbf{side and bottom face} intersect.
    
    \item \textbf{Height:} Defined as the length of the edge where the object's \textbf{front and side face} intersect (\ie, the vertical edge).
\end{itemize}

\noindent This semantic-based definition, shown in Figure \ref{suppfig: dimension_definition}, ensures that a query for ``\textit{how wide is the cabinet}" consistently refers to the same geometric property, regardless of the object's orientation in the scene.

\subsubsection{3D Spatial Referring Data Generation}
\label{suppsubsec: 3d referring data}

For 3D spatial referring tasks, we generate data that pairs multi-step descriptions with precise 3D surface points. Our approach is inspired by previous work in 2D spatial referring~\cite{zhou2025roborefer}, which typically associates a 2D pixel coordinate $(x, y)$ with a phrase like ``\textit{the corner of the leftmost table}''.
A key limitation of 2D referring data is that a 2D point represents an infinite ray in 3D space, lacking specific 3D localization. To overcome this, we build upon the 2D referring data by incorporating depth information. 
For a given 2D referring point $(x, y)$ that indicates a specific object, we also extract its corresponding depth value $d$ from the depth map. 
This critical step ensures that the ground-truth target is not an ambiguous ray but a precise 3D coordinate $(x, y, d)$.
This 3D point is guaranteed to lie on the visible surface of the target object, providing an unambiguous 3D grounding for the language description.


\subsection{Object-centric Spatial Tracing Generation}
\label{suppsubsec: object_centric_tracing}

Since static 3D scanning datasets (\eg, CA-1M, ScanNet) lack ground-truth manipulation trajectories to exect spatial traces, we develop a simulation-based generation pipeline to synthesize high-quality object-centric spatial traces. 
This pipeline transforms static 3D scene graphs with occupancy information into simulated manipulation trajectories through a structured process: initialization, planning, and reasoning.

\subsubsection{Scene Initialization \& Task Formulation}
\label{suppsubsubsec: task_definition}

Before trace generation, we must standardize the environment and rigorously define the roles of objects within the scene to ensure plausible task formulation.

\vspace{+1mm}
\noindent \textbf{Gravity Alignment and Data Validation.}
Raw 3D scans often exhibit arbitrary coordinate orientations. 
To support physics-based reasoning (\eg, ``stacking''), we first standardize the scene geometry using a gravity alignment matrix $R_{gravity}$. 
For every object $O_i$ with corner coordinates $P_i \in \mathbb{R}^{8 \times 3}$, we apply the transformation $P'_i = P_i R_{gravity}^T$.
Following alignment, we perform strict geometric validation. 
Objects with degenerate bounding boxes (fewer than 4 valid corners) or containing numerical anomalies (NaN/Inf values) are filtered out to prevent simulation instability.

\vspace{+1mm}
\noindent \textbf{Object Role Assignment.}
\highlight{A critical challenge in data generation is balancing \textit{semantic richness} for VQA with \textit{physical completeness} for simulation.}
As detailed in Sec.~\ref{suppsubsec: 3D scanning data}, our filtering pipeline produces a subset of ``\textit{High-Quality Objects}'' with rich VLM-generated captions, while the original raw dataset contains all geometric instances but lacks detailed descriptions, especially in CA-1M (more details can be found in prior work~\cite{zhou2025roborefer}). To address this, we assign roles as follows:

\begin{itemize}
    \item \textbf{Moving Objects ($O_{src}$) \& Via Objects ($O_{via}$):} Selected exclusively from the \textit{High-Quality Object Subset}. This is crucial for generating unambiguous referring expressions in VQA tasks. For instance, in a scene with multiple apples, raw category labels are insufficient to distinguish the target. We rely on the rich `dense' and `spatial' captions (\eg, ``the apple on the right'') available in the high-quality subset to ensure the generated instruction uniquely identifies the specific object (\eg, ``Move the \textit{apple on the right} around...''), rather than a generic and ambiguous ``Move the apple''.
    
    \item \textbf{Reference Objects ($O_{ref}$):} A subset of the High-Quality Objects that satisfy stability constraints. We identify $O_{ref}$ by verifying they are resting on a supporting platform (\eg, a table or shelf) using a vertical overlap heuristic. This ensures that the destination region near $O_{ref}$ is physically reachable and capable of supporting $O_{src}$.
    
    \item \textbf{Obstacles ($O_{obs}$):} Sourced from the \textit{Original Raw Dataset} (containing all annotated instances). While some objects may lack captions and are excluded from being $O_{src}$ or $O_{ref}$, they still physically exist in the scene. To ensure collision-free motion planning, our motion planner considers the union of all raw instances as the obstacle set, preventing the generated trace from hallucinating paths through uncaptioned objects.
    
\end{itemize}

\begin{figure*}[t]
  \centering
  \includegraphics[width=\linewidth]{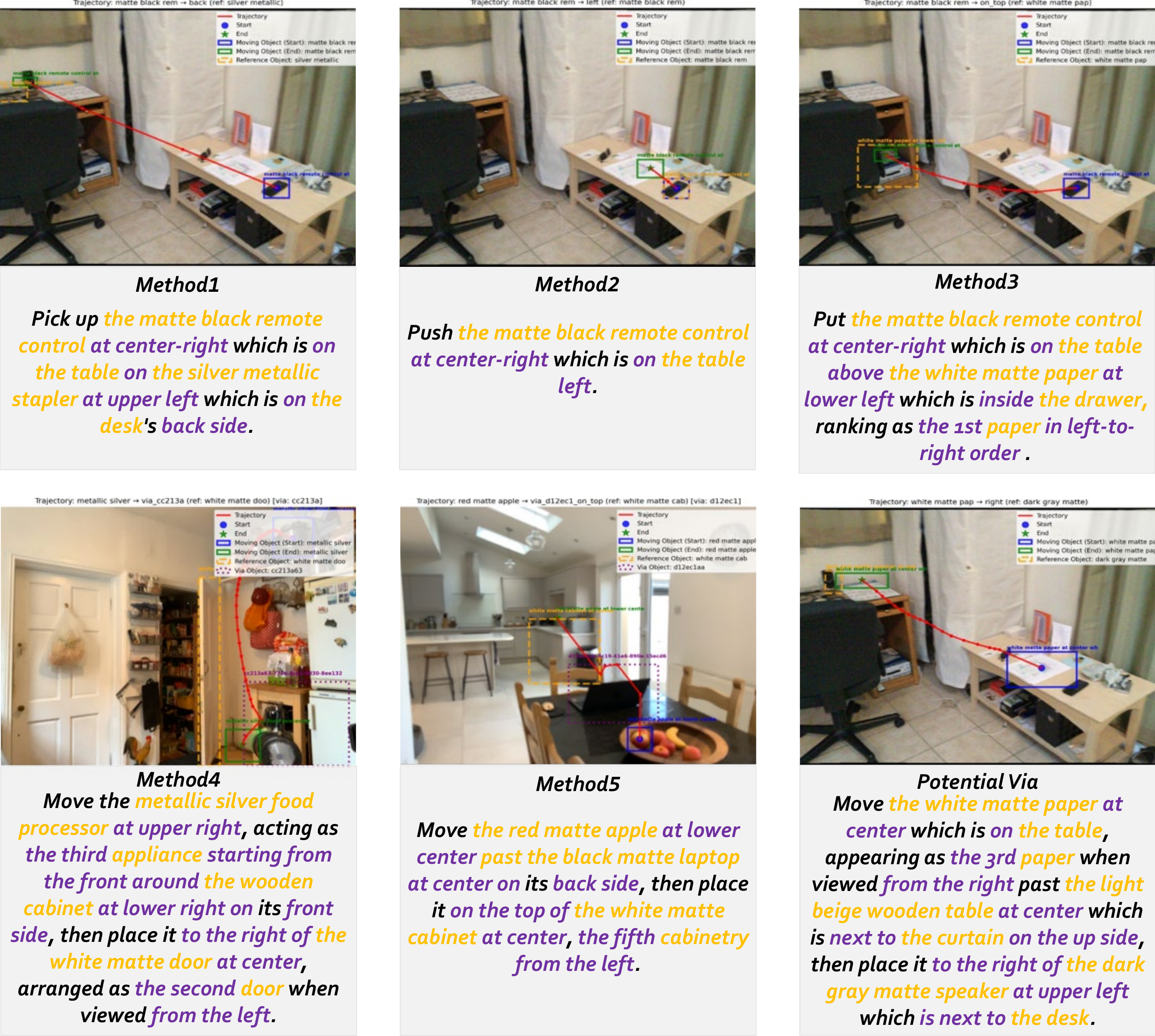} 
  \caption{\textbf{Visual Taxonomy of Manipulation Primitives.} We define five distinct primitives ranging from simple relative placement (Method 1) to complex active obstacle bypass and stacking (Methods 4--5). Additionally, we retroactively identify ``Potential Via Objects'' (bottom) for standard traces to enrich spatial descriptions in cluttered scenes.}
  \label{fig:supp_manipulation_taxonomy}
\end{figure*}

\vspace{+1mm}
\noindent \textbf{Taxonomy of Manipulation Primitives.}
To improve spatial reasoning capabilities, we define five distinct manipulation primitives (Methods 1--5). Additionally, we introduce a mechanism to identify ``Potential Via Objects'' for standard tasks to enrich spatial trace descriptions, shown in Fig.~\ref {fig:supp_manipulation_taxonomy}.

\begin{itemize}
    \item \textbf{Method 1 (Place Relative):} Moves $O_{src}$ to a spatial relation (\eg, left, right) relative to $O_{ref}$.
    \newline \textit{Instruction Example:} ``Place the \{source\_obj\} to the \{endpoint\_direction\} of the \{reference\_obj\}.''
    
    \item \textbf{Method 2 (Directional Move):} Moves $O_{src}$ by a specific distance towards a cardinal direction.
    \newline \textit{Instruction Example:} ``Move the \{source\_obj\} toward the \{endpoint\_direction\}.''
    
    \item \textbf{Method 3 (Stacking):} Places $O_{src}$ \textit{on top of} $O_{ref}$, enforcing surface area constraints (Area\_{ref} $\ge$ Area\_{src}) to ensure stability.
    \newline \textit{Instruction Example:} ``Place the \{source\_obj\} on top of the \{reference\_obj\}.''
    
    \item \textbf{Method 4 (Active Bypass \& Place):} A high-difficulty primitive where the planner explicitly identifies an obstacle ($O_{via} \in \text{High-Quality Subset}$) blocking the direct path and generates a trace to move around it before reaching the target.
    \newline \textit{Instruction Example:} ``Move the \{source\_obj\} around the \{via\_obj\} on its \{via\_direction\} side, then place it to the \{endpoint\_direction\} of the \{reference\_obj\}.''
    
    \item \textbf{Method 5 (Active Bypass \& Stack):} Combines obstacle avoidance with a final stacking operation.
    \newline \textit{Instruction Example:} ``Move the \{source\_obj\} around the \{via\_obj\} on its \{via\_direction\} side, then place it on top of the \{reference\_obj\}.''
\end{itemize}

\noindent Note on \highlight{\textit{\textbf{Potential Via Objects}}} (Method 1--3): Even for standard primitives (Methods 1--3), \highlight{the generated spatial trace may coincidentally pass close to other objects. 
We implement a \textit{\textbf{Retroactive Via Discovery}} logic} to identify these objects from the \textit{High-Quality Subset} and \highlight{add the spatial constraints along the trace path into original spatially constrained instructions.} If a spatial trace passes within a proximity threshold of an object $O_k$, $O_k$ is labeled as a ``Potential Via Object,'' adding implicit spatial constraints to the VQA annotations (e.g., ``Move \{source\_obj\} around \{via\_obj\} on its \{via\_direction\} side, then place it...'').

\subsubsection{Simulation-based Spatial Trace Generation}
\label{suppsubsubsec: simulation_planning}

Based on the initialized scene and defined tasks, we employ a rigorous motion planning pipeline in continuous 3D space ($\mathcal{C} \subset \mathbb{R}^3$). 
\highlight{Unlike rasterized grid searches (\eg, A*), our approach utilizes sampling-based planning to handle the high-resolution geometry of scanned environments, ensuring feasibility and plausibility.}

\vspace{+1mm}
\noindent \textbf{Region-based Endpoint Selection.}
While Sec.~\ref{suppsubsubsec: task_definition} defines valid destination \textit{regions}, the motion planner requires a specific goal configuration $p_{goal}$. 
\highlight{To enhance robustness against clutter, we implement a \textit{Centroid-outward Polar Sampling Strategy} within the candidate region $\mathcal{R}_{dest}$}.
Instead of uniform random sampling, we discretize $\mathcal{R}_{dest}$ into a structured polar grid centered at its geometric centroid. We define a set of concentric radii $\mathcal{R} = \{0, 0.03, 0.06, 0.10, 0.15, 0.20\}$ (meters) and angular steps. 
The pipeline iteratively evaluates candidate points from the inner rings outwards. For each candidate $p_{cand}$, we apply a height adjustment based on the target object's dimensions relative to the platform surface ($y_{target} = y_{platform} + h_{obj}/2 + \delta_{safety}$) and perform a static collision check. The first valid sample is selected as $p_{goal}$. This strategy implicitly maximizes the safety margin by preferring locations furthest from region boundaries and obstacles.

\vspace{+1mm}
\noindent \textbf{RRT* Planning with Dual-modal Escape.}
We utilize the RRT* (Rapidly-exploring Random Tree Star) algorithm to generate collision-free traces. The planner grows a tree from the start configuration $p_{start}$ towards $p_{goal}$ with a goal-biasing probability of $0.25$ and a step size of $0.05m$. 
Crucially, to optimize trace quality, we employ a rewiring radius of $0.25m$, allowing the algorithm to asymptotically converge towards the optimal path by restructuring the tree topology within the local neighborhood.

\vspace{+1mm}
\noindent \highlight{\textit{Escape Strategy for Constrained Starts.} A common failure mode in simulation is when the source object's initial pose slightly intersects with the environment (soft collision) due to meshing noise or scanning artifacts, which locks the standard planner. To address this, we implement a heuristic \textbf{Dual-modal Escape Mechanism}}:
\begin{enumerate}
    \item \textbf{Visual Opening Analysis:} If $p_{start}$ is in collision, we first analyze the depth map along 6 cardinal directions (up, down, left, right, front, back). We compute an ``Opening Score'' for each direction by ray-casting into the depth buffer; directions with longer free-space rays are prioritized.
    
    \item \textbf{Geometric Push Fallback:} If visual analysis is inconclusive (\eg, occluded camera), we fall back to a geometric approach. We calculate the penetration depth of the object's AABB against all overlapping obstacles and determine a ``Push Vector'' that minimizes the translation required to resolve the collision.
    
    \item \textbf{Recovery:} The planner moves the object along the optimal escape vector until it reaches free space $p_{free}$ (max distance $0.6m$). The escape segment $p_{start} \rightarrow p_{free}$ is then prepended to the main RRT* plan.
\end{enumerate}

\vspace{+1mm}
\noindent \textbf{Hierarchical Collision Detection.}
Precise collision checking is paramount for generating non-penetrating traces. We reject simple sphere-based approximations in favor of a two-phase hierarchical collision detection system:

\begin{itemize}
    \item \textbf{Broad Phase (AABB):} We first filter potential colliders using Axis-Aligned Bounding Boxes (AABB). Objects with non-overlapping AABBs are immediately discarded, significantly reducing the computational overhead given the large number of objects in scanned scenes.
    
    \item \textbf{Narrow Phase (OBB via SAT):} For objects passing the broad phase, we perform precise intersection tests. Since scanned objects are often oriented arbitrarily (not axis-aligned), AABBs are overly conservative and prevent valid close-proximity manipulations. We construct Oriented Bounding Boxes (OBB) for the moving object and obstacles by performing Principal Component Analysis (PCA) on their vertices to extract the three principal orthogonal axes. We then employ the Separating Axis Theorem (SAT) to test for intersection across 15 potential separating axes (3 axes from object A, 3 from B, and 9 cross-products). This rigorous check allows the planner to generate tight, realistic traces that graze valid obstacles without false positives.
\end{itemize}

\subsubsection{Advanced Spatial Reasoning Logic}
\label{suppsubsubsec: advanced_reasoning}

Beyond basic collision avoidance, our pipeline incorporates advanced spatial reasoning modules to generate and interpret complex, multi-step spatial tracing data. This includes proactive planning for obstacle bypass (Methods 4--5) and retroactive analysis of spatial relations for standard traces (Methods 1--3).

\vspace{+1mm}
\noindent \textbf{\highlight{Active Bypass Planning} (Methods 4 \& 5).}
To synthesize training data that necessitates reasoning about intermediate constraints, we implement a heuristic-guided bypass planner. This module forces the robot to deviate from the greedy trace to navigate around a specific ``Via Object'' ($O_{via}$).

\begin{enumerate}
    \item \textbf{Obstacle Identification:} We first compute a hypothetical direct trace using the standard RRT* planner. The system identifies potential blocking objects by calculating the segment-to-point distance for all scene obstacles. If an object $O_{block}$ lies within a collision margin ($<0.05m$) of the direct path, it is designated as the target Via Object.
    
    \item \textbf{Multi-directional Candidate Generation:} Unlike simple heuristics that always bypass ``to the right'', our system evaluates 6 cardinal bypass directions relative to $O_{via}$'s geometry: \textit{left, right, front, back, up, down}. For each direction $d$, we define a valid via-region $\mathcal{V}_d$. The center of this region is calculated as:
    \begin{equation}
        c_{via} = c_{obs} + (r_{obs} + r_{src} + \delta_{margin}) \cdot \vec{n}_d
    \end{equation}
    where $c_{obs}$ is the obstacle center, $\vec{n}_d$ is the normal vector for direction $d$, and $r_{obs}, r_{src}$ are the radii of the obstacle and moving object, respectively. This explicitly ensures kinematic clearance.
    
    \item \textbf{Heuristic Scoring for Naturalness:} To select the most human-like bypass spatial trace, we score each candidate region using a composite cost function $J$. Based on our empirical tuning, the cost is defined as:
    \begin{equation}
        \footnotesize
        J = 1.0 \cdot L_{total} + 0.3 \cdot P_{angle} + 2.0 \cdot P_{backtrack} + 0.2 \cdot P_{lateral}
    \end{equation}
    Here, $L_{total}$ is the total trace length. $P_{angle}$ penalizes sharp turns (cosine similarity). $P_{backtrack}$ is a binary penalty (weight 2.0) that strictly discourages candidates requiring movement opposite to the goal vector. $P_{lateral}$ penalizes excessive deviation from the main axis. The candidate minimizing $J$ is selected as the optimal via-point $p_{via}$.
    
    \item \textbf{Two-stage Planning \& Fusion:} The planner executes two independent RRT* searches: $\tau_{1}: p_{start} \to p_{via}$ and $\tau_{2}: p_{via} \to p_{goal}$. The resulting paths are concatenated. Crucially, during the smoothing phase (Catmull-Rom), we enforce a \textit{via-point constraint}: any optimization that moves the spatial trace further than $0.12m$ from $p_{via}$ is rejected, preserving the intentional bypass behavior.
\end{enumerate}

\vspace{+1mm}
\noindent \textbf{\highlight{Retroactive Via Discovery} (Methods 1--3).}
Standard manipulation tasks (\eg., ``Place A next to B'') often implicitly involve spatial constraints relative to other objects in cluttered scenes. To capture this rich semantic information, we implement a retroactive discovery module.
\begin{enumerate}
    \item \textbf{Proximity Analysis:} For a generated spatial trace $\mathcal{T}$, we compute the minimum Euclidean distance to all High-Quality objects in the scene. We calculate the distance to the object's \textit{surface} rather than its center: $d_{surf} = ||p_{traj} - c_{obj}|| - r_{obj}$. Objects satisfying $d_{surf} < 0.15m$ are flagged as Potential Via Objects.
    
    \item \textbf{Relative Direction Classification:} For each flagged object, we determine the spatial relationship of the bypass. We identify the closest point $p_{close}$ on the trace and calculate the offset vector $\vec{v}_{off} = c_{obj} - p_{close}$. By determining the dominant component of $\vec{v}_{off}$ orthogonal to the trace's instantaneous velocity vector, we classify the bypass direction into natural language labels (\eg, ``passing on the left'', ``passing above'').
    
    \item \textbf{Instruction Enrichment:} These discovered relations are injected into the VQA generation pipeline, transforming simple instructions into spatially dense descriptions (\eg, from ``Move mug to plate'' to ``Move mug to plate, \textit{passing to the right of the bottle}''). This supervision signal is critical for training RoboTracer to attend to collision boundaries and intermediate spatial context.
\end{enumerate}

\subsubsection{Trace Refinement \& Quality Assurance}
\label{suppsubsubsec: refinement_qa}

\highlight{The raw traces generated from the motion planner are geometrically valid in 3D but may lack the visual alignment and naturalness required for VLM training. We implement a comprehensive refinement pipeline followed by strict quality assurance protocols.}

\vspace{+1mm}
\noindent \textbf{Geometric Smoothing \& Compression.}
To mitigate the robotic artifacts of RRT*, we apply Catmull-Rom Spline smoothing ($\alpha=0.5$) to generate fluid, human-like curves. Subsequently, we utilize the Ramer-Douglas-Peucker (RDP) algorithm to downsample the dense path into a sparse sequence of keypoints ($N \le 8$), reducing token complexity while preserving critical motion geometry.

\vspace{+1mm}
\noindent \textbf{Physics-aware Grounding (End-Point Refinement).}
A common simulation artifact is ``floating placement,'' where the trace ends at the object's geometric center, leaving it visually suspended. We implement a depth-guided correction: by projecting the endpoint onto the image plane and querying the ground-truth depth map, we iteratively descend the trace's final point along the gravity vector until it makes contact with the physical surface (\ie, projected depth matches sensor depth). This ensures a physically grounded placement.

\vspace{+1mm}
\noindent \textbf{Visual Alignment Correction (Start-Point Refinement).}
While 3D planning operates on the object's geometric center, simply projecting this 3D center into the 2D image may result in a starting point that falls outside the object's visual mask (\eg, for C-shaped objects or due to perspective distortion). To ensure precise visual grounding:
\begin{enumerate}
    \item We retrieve the ground-truth 2D segmentation mask (RLE-encoded) of the source object.
    \item We calculate the visual centroid $(u_{mask}, v_{mask})$ of the mask's largest connected component.
    \item We override the 2D coordinates of the trace's start point with $(u_{mask}, v_{mask})$, while retaining the original metric depth $d_{start}$.
\end{enumerate}
This alignment guarantees that the visual trace originates perceptually from the object's body, eliminating ``off-target'' supervision signals.

\vspace{+1mm}
\noindent \textbf{Rigorous Quality Control.}
To filter out hallucinations and low-quality samples, we enforce a set of validation criteria. A trace is discarded if it fails any of the following checks:
\begin{itemize}
    \item \textbf{Occlusion Ratio:} We compute the visibility of every interpolated waypoint against the scene's depth buffer. Traces with $> 30\%$ occlusion are rejected to ensure the trace is visually trackable.
    \item \textbf{Field-of-View Constraint:} All keypoints, especially the start and end, must project strictly within the camera frame boundaries.
    \item \textbf{Volume-Adaptive Dynamics:} Fixed minimum-length thresholds fail to account for scale variance. We apply a dynamic threshold $L_{min} = L_{base} \cdot \sqrt[3]{V_{obj}}$. This requires large objects (\eg, laptops) to move significant distances to constitute a valid action, while allowing subtle manipulations for small objects (\eg, keys).
    \item \textbf{Semantic Movability:} We filter out immovable fixtures (\eg, ``floor'', ``wall'', ``countertop'') using a semantic blocklist, ensuring the model focuses solely on interactable entities.
\end{itemize}

\subsubsection{VQA Data Generation}
\label{suppsubsubsec: vqa_synthesis}

The final stage transforms the refined metric-grounded geometric-aware traces into multimodal instruction-following pairs. To foster robust spatial reasoning, we synthesize instructions with varying levels of granularity and generate three distinct types (\eg, 2D, 3D, and Lifting) for each trace.

\vspace{+1mm}
\noindent \textbf{Instruction Generation.}
For each generated trace, we synthesize a language instruction using a template-based approach. A critical feature of our pipeline is the injection of Metric Scale Awareness.
As shown in Listing~\ref{lst:instruction_templates}, our template pool is designed such that approximately \textbf{20\%} of the templates include explicit metric placeholders (\texttt{\{distance:.3f\}m}).
During generation, we calculate the ground-truth trace displacement $L_{traj}$ and populate these slots.
This design forces the model to not only understand relative directions (\eg, ``to the right'') but also correlate visual magnitudes with precise numerical values (\eg, ``0.25m'').

\begin{lstlisting}[basicstyle=\ttfamily\footnotesize, backgroundcolor=\color{myblue!50}, caption={Templates for Metric-aware Instruction Generation.}, captionpos=t, breaklines=true, label={lst:instruction_templates}]
INSTRUCTION_TEMPLATES = {
    "method1_place_relative": [
        # Metric-aware templates (~20%)
        "Move the {source_obj} to a position {distance:.3f}m to the {endpoint_direction} of the {reference_obj}.",
        "Pick up the {source_obj} and move it to a position {distance:.3f}m to the {endpoint_direction} of the {reference_obj}.",
        # Standard templates
        "Place the {source_obj} to the {endpoint_direction} of the {reference_obj}.",
        "Pick up the {source_obj} on the {reference_obj}'s {endpoint_direction} side.",
    ],
    "method2_directional_move": [
        # Metric-aware templates
        "Move the {source_obj} {distance:.3f}m in the {endpoint_direction} direction.",
        "Pick up the {source_obj} and move it {distance:.3f}m toward the {endpoint_direction}.",
        # Standard templates
        "Push the {source_obj} toward the {endpoint_direction}.",
        "Slide the {source_obj} toward {endpoint_direction}.",
    ],
    "method3_stacking": [
        # Stacking specific templates
        "Place the {source_obj} on top of the {reference_obj}.",
        "Stack the {source_obj} on the {reference_obj}.",
        "Put the {source_obj} above the {reference_obj}.",
        # Vertical relative positioning
        "Move the {source_obj} onto the {reference_obj}.",
        "Set the {source_obj} on the {reference_obj}.",
    ],
    "method4_bypass_place": [
        # Implicit constraints via 'Via Object'
        "Move the {source_obj} around the {via_obj} on its {via_direction} side, then place it to the {endpoint_direction} of the {reference_obj}.",
        "Pick up the {source_obj} around the {via_obj} from the {via_direction} side, then position it to the {endpoint_direction} of the {reference_obj}.",
    ],
    "method5_bypass_stack": [
        # Complex composite task: Bypass + Stacking
        "Move the {source_obj} around the {via_obj} on its {via_direction} side, then place it on top of the {reference_obj}.",
        "Pick up the {source_obj} around the {via_obj} from the {via_direction} side, then place it on the {reference_obj}.",
    ],
    "potential_via_enrichment": [
        # Retroactive description for standard tasks
        "Move the {source_obj} around the {via_obj} on its {via_direction} side, then {final_action}.",
        "Pick up the {source_obj}, passing to the {via_direction} of the {via_obj}, then {final_action}.",
    ]
}
\end{lstlisting}

\vspace{+1mm}
\noindent \textbf{Multi-task VQA Formatting.}
To fully exploit the generated data, we construct three distinct VQA tasks for every single trace sample. We utilize specific prompt templates (sourced from the ``Droid'' template set in our codebase) to format the queries. This multi-task formulation encourages the model to learn consistent representations across 2D and 3D spaces.

\vspace{+1mm}
\noindent \textbf{Type 1: 2D Visual Trace.}
This task supervises the model to ground instructions into the 2D image plane. The output is a sequence of 2D coordinates $\tau_{2D} = \{(u_i, v_i)\}_{i=1}^N$.

\begin{lstlisting}[basicstyle=\ttfamily\footnotesize, backgroundcolor=\color{myblue!50}, caption={Prompts for 2D Visual Tracing.}, captionpos=t, breaklines=true, label={lst:prompts_2d}]
PROMPTS_2D = [
    "Please predict 2D object-centric waypoints to complete the task successfully. The task is \"<instruction>\". Your answer should be formatted as a tuple, i.e. [(x, y)], where the tuple contains the x and y coordinates of a point satisfying the conditions above.",
    "Point the 2D object-centric waypoints for the task \"<instruction>\". Your answer should be formatted as a tuple, i.e. [(x, y)].",
    "You are currently a robot performing robotic manipulation tasks. The task instruction is: \"<instruction>\". Use 2D points to mark the manipulated object-centric waypoints...",
    "Please predict 2D object-centric visual trace to complete the task successfully. The task is \"<instruction>\". Your answer should be formatted as a tuple, i.e. [(x, y)]."
]
\end{lstlisting}

\vspace{+1mm}
\noindent \textbf{Type 2: 3D Spatial Trace.}
This is the core task, requiring the model to infer depth and 3D structure from monocular input. The output is a sequence of 3D coordinates $\tau_{3D} = \{(u_i, v_i, d_i)\}_{i=1}^N$, where $d_i$ is the absolute metric depth.

\begin{lstlisting}[basicstyle=\ttfamily\footnotesize, backgroundcolor=\color{myblue!50}, caption={Prompts for 3D Spatial Tracing.}, captionpos=t, breaklines=true, label={lst:prompts_3d}]
PROMPTS_3D = [
    "Please predict 3D object-centric waypoints to complete the task successfully. The task is \"<instruction>\". Your answer should be formatted as a list of tuples, i.e., [(x1, y1, d1), (x2, y2, d2), ...], where each tuple contains the x and y coordinates and the depth of the point.",
    "Point the 3D object-centric visual trace for the task \"<instruction>\". Your answer should be formatted as a list of tuples, i.e., [(x1, y1, d1), ...].",
    "You are currently a robot performing robotic manipulation tasks. The task instruction is: \"<instruction>\". Use 3D points to mark the manipulated object-centric waypoints to guide the robot...",
]
\end{lstlisting}

\vspace{+1mm}
\noindent \textbf{Type 3: 2D-to-3D Trace Lifting.}
This task isolates the geometric reasoning capability. The model is provided with the ground-truth 2D trace in the text prompt and must ``lift'' it into 3D space. This effectively trains the model to perform trace-conditioned depth estimation.

\begin{lstlisting}[basicstyle=\ttfamily\footnotesize, backgroundcolor=\color{myblue!50}, caption={Prompts for 2D-to-3D Trace Lifting.}, captionpos=t, breaklines=true, label={lst:prompts_lift}]
PROMPTS_LIFT = [
    "Please lift the 2D object-centric waypoints to 3D object-centric waypoints to complete the task successfully. The task is \"<instruction>\". The 2D waypoints is <trace>. Your answer should be formatted as a list of tuples, i.e., [(x1, y1, d1), ...].",
    "Lift the 2D object-centric visual trace to 3D object-centric visual trace for the task \"<instruction>\". The 2D visual trace is <trace>. Your answer should be formatted as a list of tuples, i.e., [(x1, y1, d1), ...].",
    "Please lift the 2D object-centric visual trace to 3D object-centric visual trace to complete the task successfully. The task is \"<instruction>\". The 2D visual trace is <trace>. Your answer should be formatted as a list of tuples..."
]
\end{lstlisting}

\vspace{+1mm}
\noindent \textbf{Trace Representation.}
For all tasks, the spatial trace is represented as a sequence of discrete tokens. We normalize the 2D pixel coordinates $(u, v)$ to the range $[0, 1000]$ and keep the depth $d$ in absolute meters. This hybrid representation allows the model to leverage its pre-trained 2D visual grounding capabilities while learning precise 3D metric structures.

\begin{figure*}
\centering
\includegraphics[width=\linewidth]{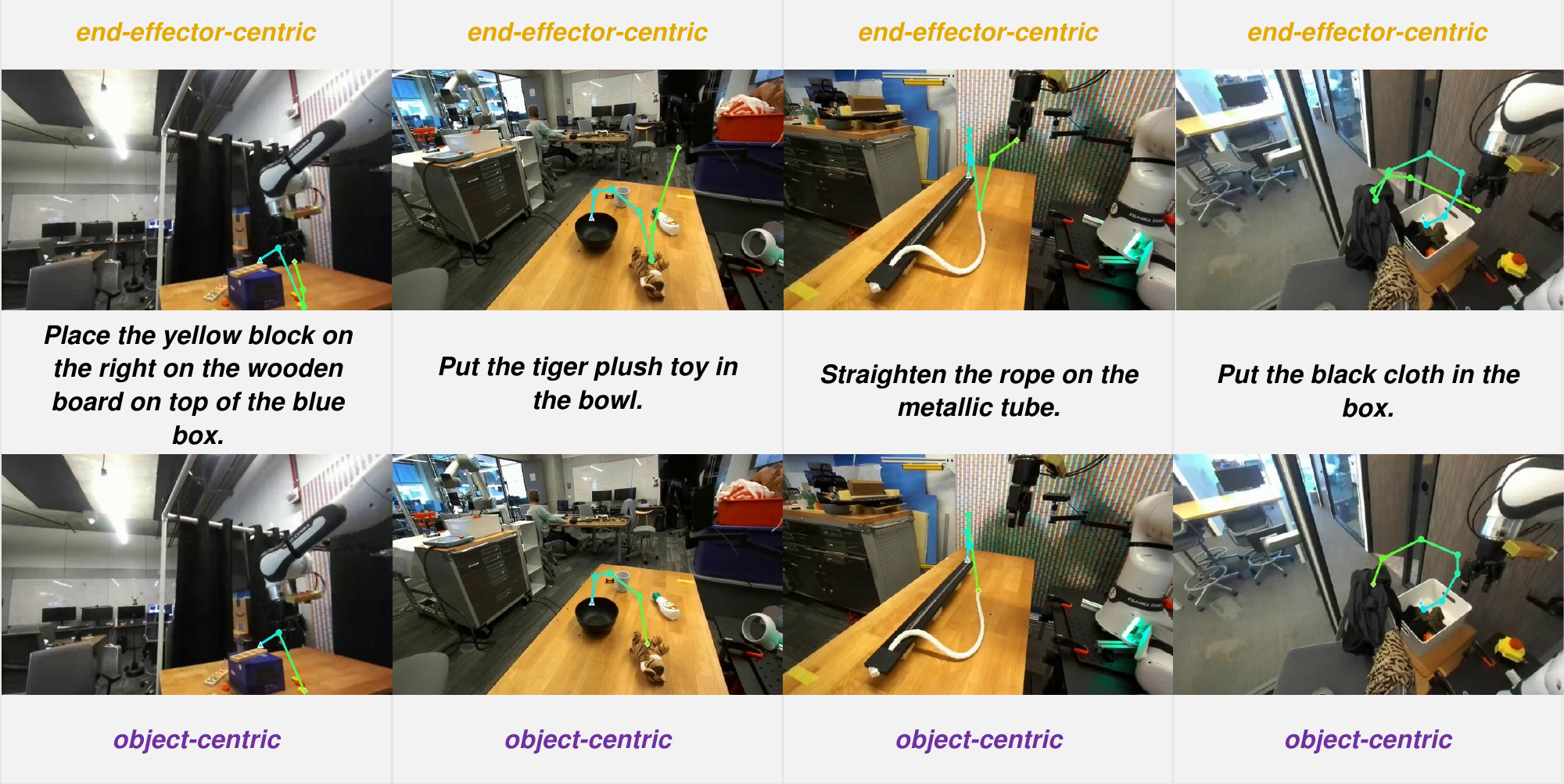}
   \caption{Extract end-effector-centric and object-centric traces from Droid’s original data by projecting gripper positions into a base frame; downsample traces using RDP and remove tasks with too many points after downsampling.
   }
\label{fig: droid_eefobj}
\end{figure*}

\begin{figure*}
\centering
\includegraphics[width=1\linewidth]{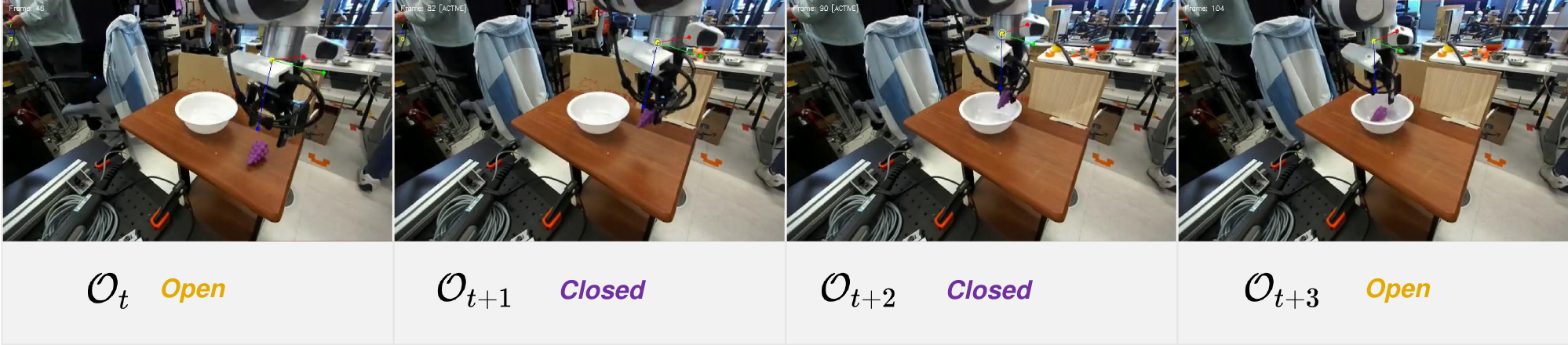}
\vspace{-7mm}
   \caption{Visualization of the end-effector coordinate frame. We obtain the gripper position (blue point) by extending the end-effector's local $z$-axis and record both the spatial trace of the gripper and its open/closed states.
   }
\label{suppfig: droid_closed_open}
\vspace{-3mm}
\end{figure*}

\subsection{Manipulation Video}
\label{suppsubsec: manipulation video}

While 3D scans enable object-centric tracing, they lack physically plausible manipulations.
We further leverage manipulation video datasets, either real-world videos or simulation videos, to provide spatial traces aligned with the embodied manipulation in tabletop settings.

\begin{figure*}
\centering
\includegraphics[width=1\linewidth]{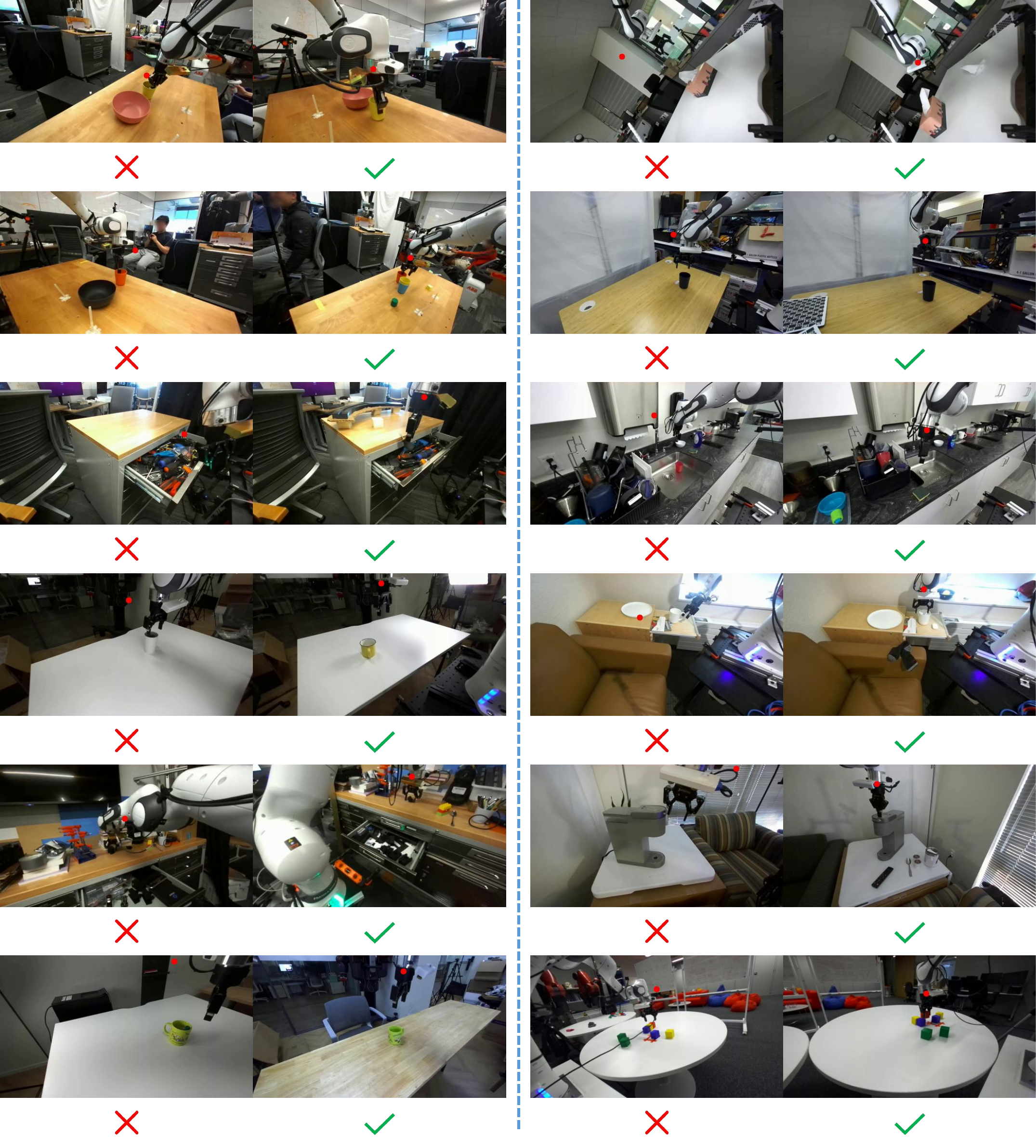}
\vspace{-7mm}
   \caption{Preview images from the Droid raw dataset with the end-effector projected as a red dot. Each row shows a pair of images from the same scene: the left image corresponds to incorrect camera extrinsics, and the right image corresponds to correct extrinsics.
   }
\label{suppfig: supply_droid2}
\end{figure*}

\subsubsection{Droid Data Processing}
\label{droid}
The Droid dataset~\cite{khazatsky2024droid} provides two versions: RLDS-format version and raw version containing raw stereo videos and metadata. 
Since the RLDS version does not allow extraction of camera intrinsics or depth information, we use the raw version. 
Ignoring the wrist cameras, the Droid raw dataset contains 117k stereo videos stored in the original SVO format. 
We extract the camera intrinsics and depth frames from these SVO files using the ZED SDK. 
\highlight{The Droid dataset contains a substantial number of samples with incorrect extrinsics (see Fig.\ref{suppfig: supply_droid2}), which manifest as the projected end-effector positions not aligning with the actual end-effector locations in the images.}
After processing the Droid Raw dataset, we obtain \textbf{20.5k} unique manipulation trajectories and generate a total of \textbf{46.8k }end-effector-centric QAs and \textbf{58.4k} object-centric QAs based on templates, covering both 2D and 3D trace generation tasks (see Fig.~\ref{fig: droid_eefobj}, \ref{fig: droid_2d3d}). The process is as follows:

\begin{figure*}
\centering
\includegraphics[width=\linewidth]{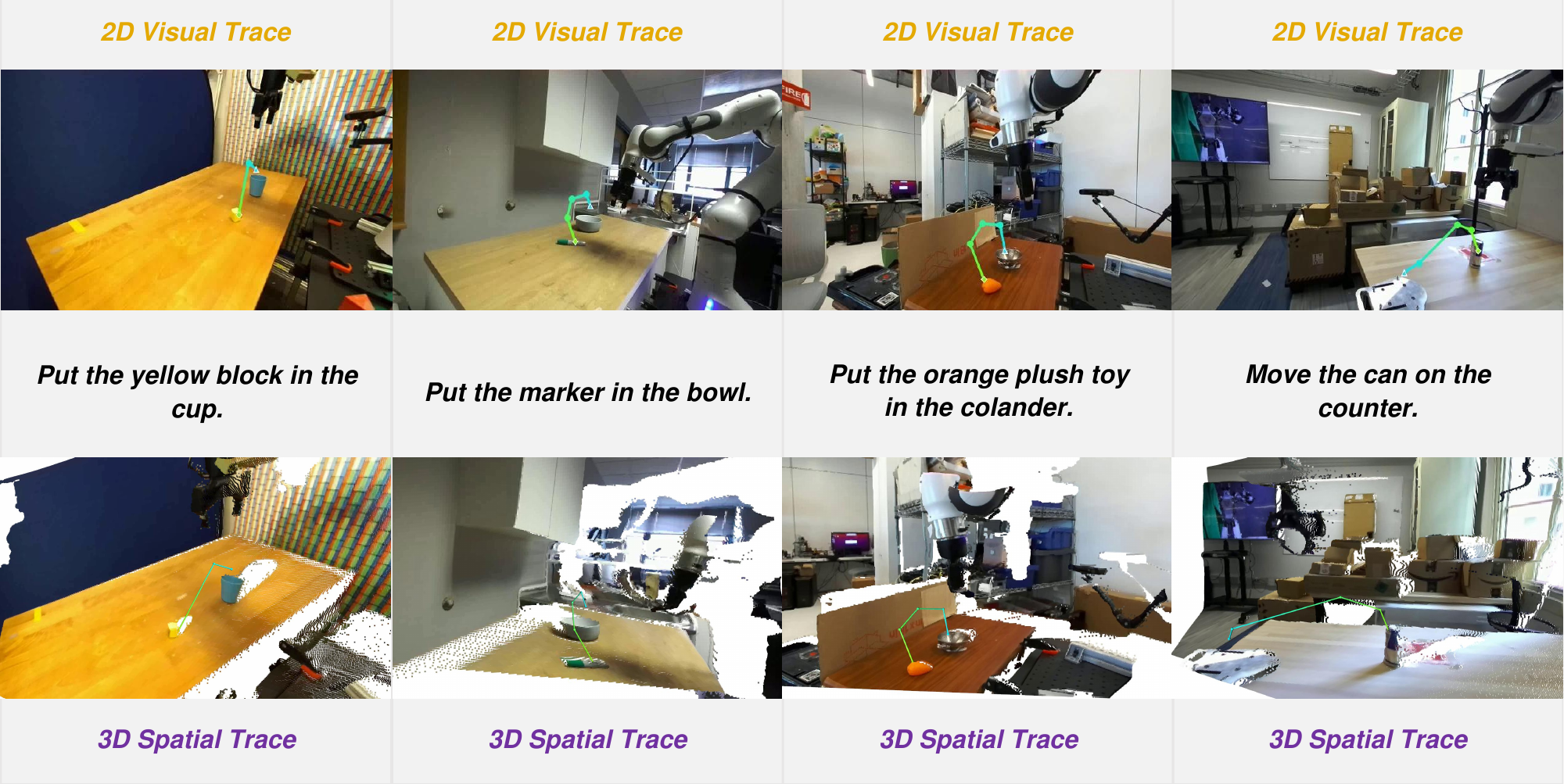}
   \caption{Droid's 2D and 3D trace visualizations under the object-centric setting. The 3D traces are visualized in the point-cloud space.
   }
\label{fig: droid_2d3d}
\end{figure*}

\vspace{+1mm}
\noindent \textbf{Raw Data Process.}
The original dataset is filtered through four criteria, resulting in $80\text{k}$ valid SVO files.
The criteria include:  \textbf{(1)} Presence of metadata; \textbf{(2)} Availability of language annotations; \textbf{(3)} Complete H5 file provided; \textbf{(4)} Presence of the SVO file. 
For each valid SVO file, camera intrinsics and extrinsics, as well as depth frames, are extracted using the ZED SDK.

\begin{figure*}
\centering
\includegraphics[width=1\linewidth]{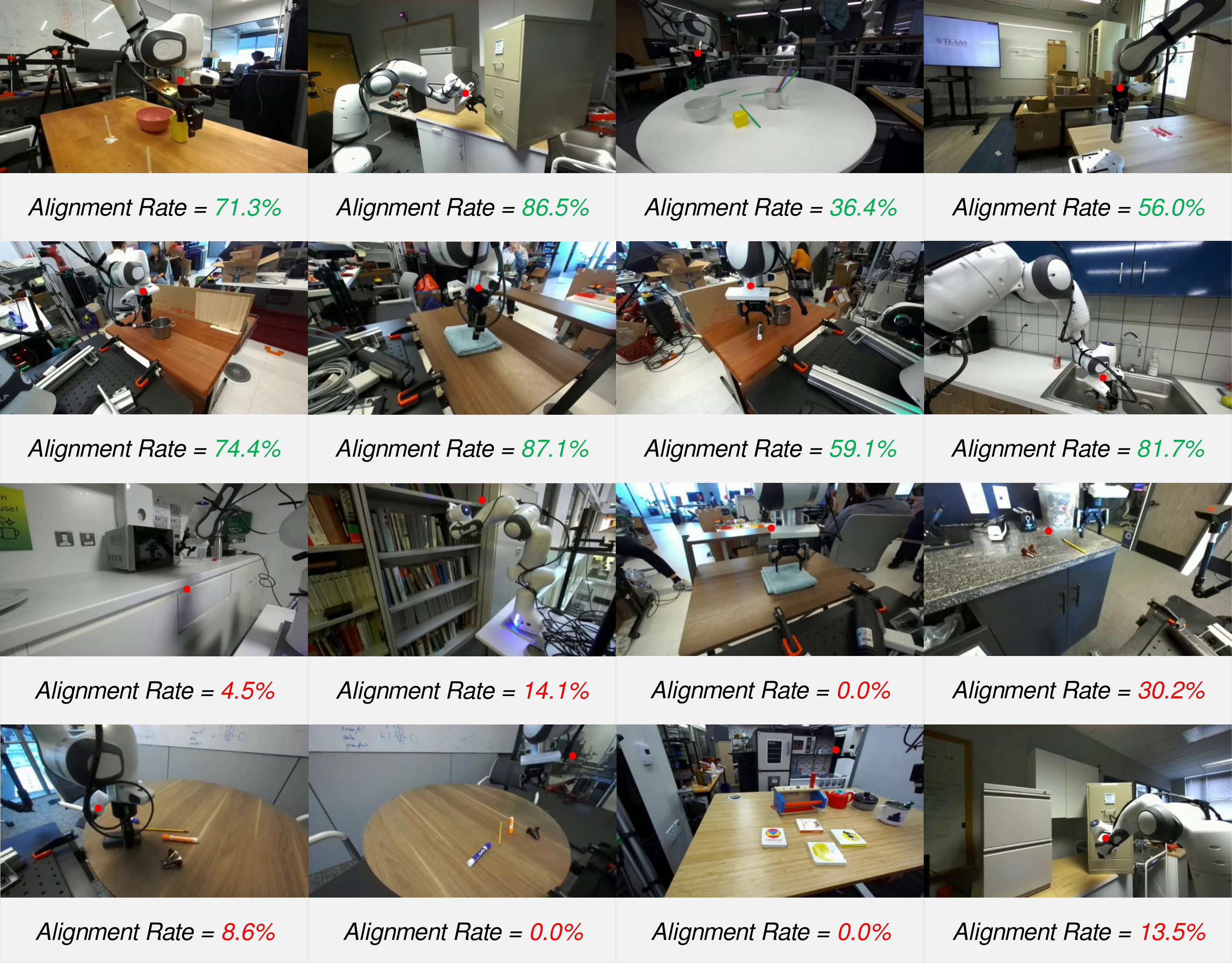}
   \caption{DROID depth-alignment results. We verify camera extrinsics by projecting the end-effector into the image and comparing pixel depth with the camera-frame end-effector $z$ value; extrinsics are accepted if over $\frac{1}{3}$ frames have a depth difference under 5 cm.
   }
\label{fig: droid_depth_check}
\end{figure*}

\vspace{+1mm}
\noindent \textbf{Extrinsic Validation.}
To verify the correctness of the camera extrinsics, we apply a strict depth-alignment check, which also filters out some occluded frames. The procedure is as follows:
\begin{enumerate}
        \item Project the end-effector coordinates into the corresponding image frame using the camera intrinsics $K$ and extrinsics $(R, t)$:
        $$\mathbf{u} = K \left(R \mathbf{p}_{eef} + t \right),$$
         where $\mathbf{p}_{eef}$ is the end-effector position in world coordinates, and $\mathbf{u}$ is the projected pixel coordinate.
         \item Compare the depth value at the projected pixel $D(\mathbf{u})$ with the end-effector's $z$-coordinate in the camera frame, $z_{eef}^{cam}$. The projection is considered consistent if:
        $$|D(\mathbf{u}) - z_{eef}^{cam}| < 5\,\text{cm}.$$
    Only frames where the projected pixel lies within the image bounds are considered.
    \item If the fraction of frames satisfying the above consistency exceeds $\frac{1}{3}$ of the total tested frames, the extrinsics are deemed valid. This procedure also filters out frames in which the end-effector is occluded by the robot arm. Fig.~\ref{fig: droid_depth_check} presents several examples illustrating the correctness of our method. 
\end{enumerate}

\vspace{+1mm}
\noindent \textbf{End-Effector Position Definition and Task Segmentation.}
We define the gripper position along the end-effector coordinate system's $z$-axis at a distance of $0.15\,\text{m}$ from the base of the end-effector (See Fig.~\ref{suppfig: droid_closed_open}). Since Droid does not provide pre-segmented tasks, we perform segmentation based on gripper closure events:
(1) Detect intervals where the gripper is closed.
(2) Retain only videos containing a single gripper-closure interval.
(3) Assign the video instruction directly as the description of the corresponding motion segment.


\vspace{+1mm}
\noindent \textbf{Trace Extraction and Downsampling.}
For each selected video, we extract end-effector-centric and object-centric traces as follows:
(1) Let the gripper closure interval be $[F_s, F_e)$. Define a base frame: $F_b = \max(0, F_s- 60)$.
(2) Project the gripper positions within $[F_s, F_e)$ into the base frame to obtain the object trace.
(3) Project the gripper positions within $[F_b, F_e)$ into the base frame to obtain the end-effector trace.
(4) Downsample each trace using the RDP algorithm and remove tasks with excessive points remaining after RDP downsampling. The visualization of the extracted results is shown in Fig.~\ref{fig: droid_eefobj},~\ref{fig: droid_2d3d}.

\begin{figure*}[t]
    \centering
    \includegraphics[width=\linewidth]{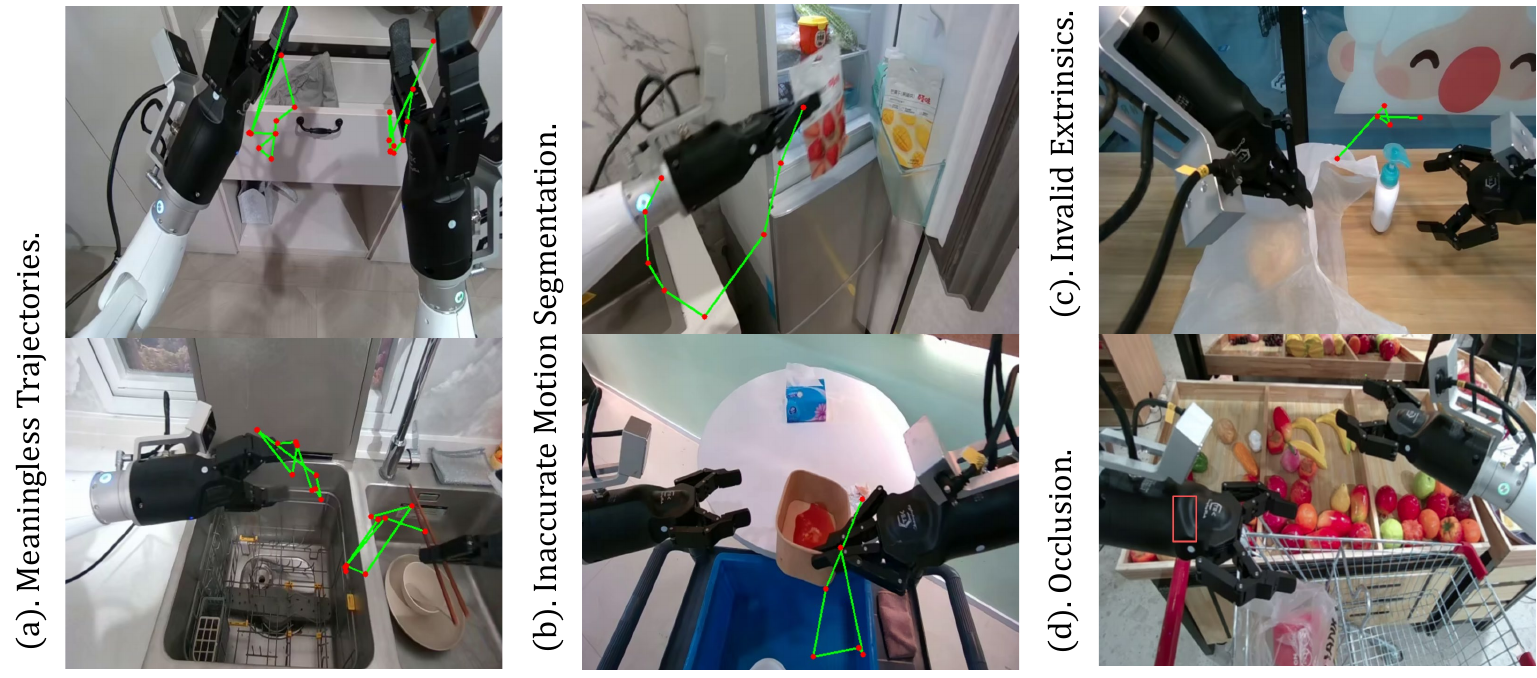}
    \caption{Common issues in the AgiBot dataset. 
(a) \textbf{Meaningless Traces.} For certain dual-arm collaborative tasks, spatial traces alone cannot accurately describe the motion, making them difficult to interpret even for humans. 
(b) \textbf{Inaccurate Motion Segmentation.} The original segmentation is often temporally misaligned; for example, trace recording continues even after the object has already been placed on the platform or inserted into the container. 
(c) \textbf{Invalid Extrinsics.} Incorrect camera extrinsics lead to inaccurate trace projections. 
(d) \textbf{Occlusion.} The manipulated object (highlighted in the red box) is occluded in the image, rendering the resulting trace meaningless.}

    \label{suppfig: agibot common issues}
\end{figure*}

\subsubsection{AgiBot Data Processing}

We use the AgiBot World Beta dataset~\cite{contributors2024agibotworldrepo} to extract spatial traces from dual-arm robotic manipulation. 
\highlight{However, the raw data contains several issues as shown in Fig.~\ref{suppfig: agibot common issues}, including invalid extrinsics, inaccurate motion segmentation, and severe occlusions, which should be resolved prior to trace extraction.}
We detail our full preprocessing pipeline below.

\vspace{+1mm}
\noindent \textbf{Tasks Filtering.}
We begin with a coarse filtering stage to retain only demonstrations that can meaningfully support spatial trace supervision. 
In Fig.~\ref{suppfig: agibot common issues}, demonstrations exhibiting global robot-based motion while the camera extrinsics remain unchanged implicitly introduce incorrect camera poses; such sequences are manually removed. 
AgiBot additionally contains many samples with extrinsic errors, which are later detected using depth-alignment validation.

\vspace{+1mm}
\noindent \textbf{Extrinsics Validation via Depth Alignment.}
Unlike Droid, AgiBot uses a head-mounted RGB-D camera whose depth maps encode near-range geometry, including the robot arm, as zero depth. To continue using depth consistency for extrinsics validation, we extend the definition of aligned depth:
$$D(\mathbf{u})=0 \text { or }\left|D(\mathbf{u})-z_{e e f}^{c a m}\right|<0.05 \mathrm{~m} .$$

\begin{figure*}
    \centering
    \includegraphics[width=1\linewidth]{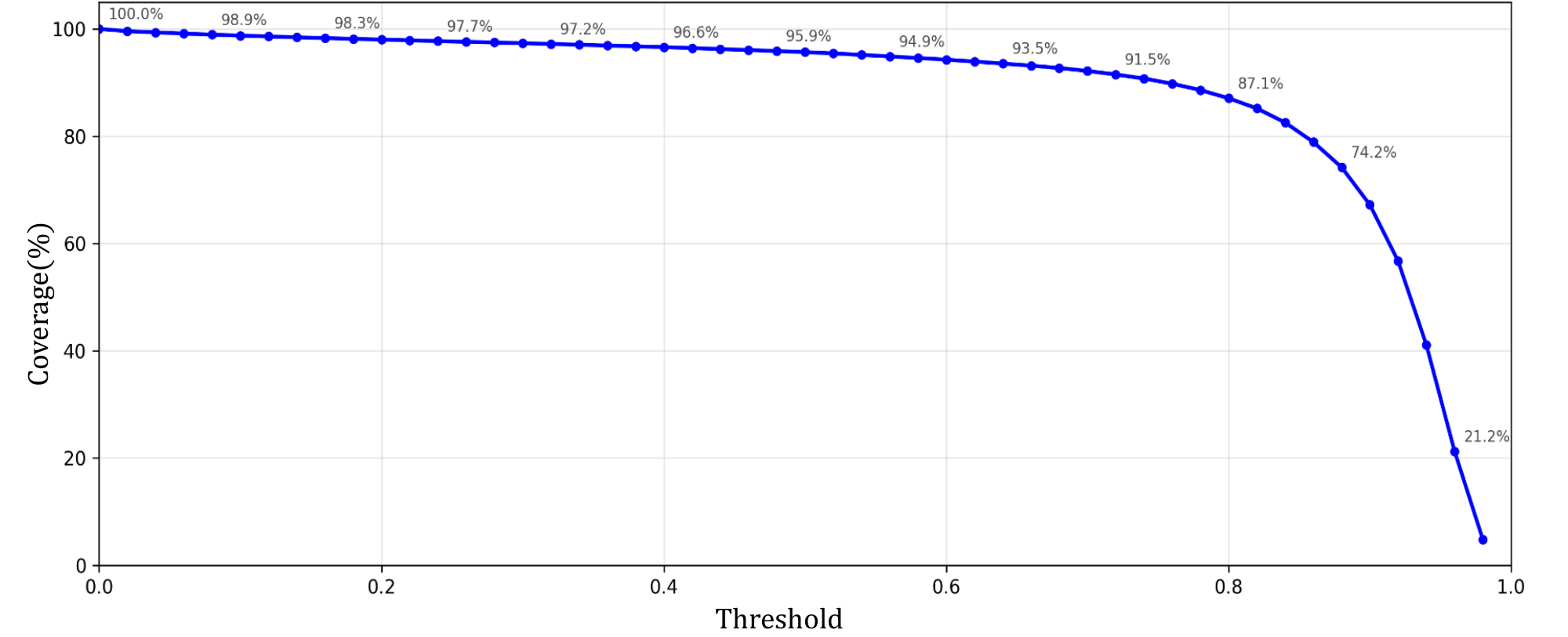}
    \caption{Proportion of retained samples under different threshold values. A sample is considered to have correct extrinsics if the proportion of aligned frames exceeds a given threshold.}
    \label{suppfig: agibotfig1}
\end{figure*}

\begin{figure*}
    \centering
    \includegraphics[width=1\linewidth]{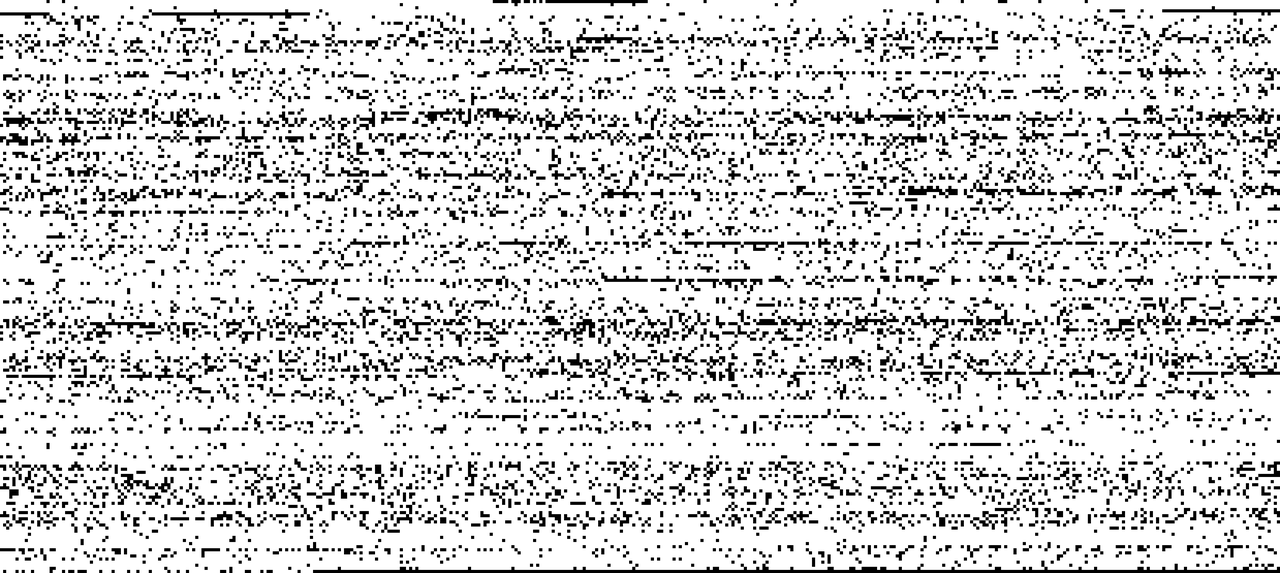}
    \caption{Visualizing all samples as a 2D grid (black = misaligned). Misaligned frames form spatially contiguous clusters, consistent with the observation that extrinsic failures occur in temporally coherent segments under fixed scene configurations.}
    \label{suppfig: agibotfig2}
\end{figure*}

\noindent A video is accepted if the proportion of aligned frames exceeds a threshold. The coverage of retained samples under different choices of the threshold in Fig.~\ref{suppfig: agibotfig1}. We use $0.83$ as the threshold in practice. Fig.~\ref{suppfig: agibotfig2} visualizes all samples as a 2D grid (black = misaligned) and reveals that misaligned frames appear in spatially contiguous clusters, matching manual observations that extrinsic failures occur in temporally coherent segments under fixed scene configurations.


\vspace{+1mm}
\noindent \textbf{Improved Motion Segmentation.}
The original AgiBot annotations decompose tasks into atomic labels (\eg, \emph{Pick}, \emph{Place}), but these labels are often temporally inaccurate. For instance, the \emph{Place} interval frequently extends well after the object has reached the target surface. Such coarse segmentation prevents the reliable construction of object-centric traces. We redesign the segmentation to obtain precise end-effector- and object-centric traces.

\begin{itemize}
    \item \textbf{Merging Pick--Place Pairs:} 
We scan the annotation and merge each \emph{Pick} with its subsequent \emph{Place} segment into a unified Pick-and-Place (P\&P) subtask.
We define the gripper position as the point located at
    $z = 0.15 \ \text{m}$
along the end-effector's local $z$-axis. 
All traces are expressed in the coordinate frame of the first frame of the subtask, which we refer to as the \emph{base frame}.
\item \textbf{Trace Extraction Rules:} For each subtask with exactly one gripper-closure interval, we extract traces as follows:

\begin{itemize}
    \item \textbf{Pick subtasks:}
    \begin{itemize}
        \item If only one arm closes, we record the end-effector-centric trace from the subtask start to the closure-interval start.
        \item If both arms close and both move, the sample is discarded (ambiguous).
        \item If both arms close but only one moves, we use the moving arm only.
    \end{itemize}

    \item \textbf{Pick-and-Place subtasks:}
    \begin{itemize}
        \item If only one arm closes, we record the end-effector-centric trace from the subtask start to the closure-interval end, and the object-centric trace in the closure-interval.

        \item If both arms close but only one moves, we use the moving arm following the same rule.
    \end{itemize}

    \item \textbf{Other subtasks:}
    \begin{itemize}
        \item If exactly one arm contains a closure interval, we record the end-effector-centric trace from the subtask start $\rightarrow$ closure-interval end.
        \item If both arms close but only one arm moves, we use the moving arm.
    \end{itemize}
\end{itemize}

This redesigned segmentation enables the construction of both end-effector-centric and object-centric traces across a wide variety of manipulation behaviors.
\end{itemize}

\paragraph{Occlusion Detection.}
Head-camera viewpoints introduce significant occlusion, especially when the robot arm blocks the manipulated object. To filter such cases, we evaluate gripper visibility at a designated \emph{check frame}:

\begin{itemize}
    \item For Pick and P\&P subtasks: use the start of the closure interval.
    \item For other subtasks: use the end of the closure interval or the subtask end.
\end{itemize}

We project the gripper position at the check frame into the base frame and inspect depth values within a radius-30 pixel neighborhood. If more than $60\%$ of these depth values are zero, the sample is marked as occluded and discarded.

\paragraph{QA Generation.}
We apply RDP simplification to all extracted traces to reduce redundancy while preserving geometric structure. Traces that contain out-of-bounds projected points or retain an excessively large number of points after RDP compression are discarded. Finally, for all valid traces, we instantiate trace-description templates according to their types (end-effector--centric or object-centric) to construct three categories of QA pairs, yielding a total of \textbf{977.5k} samples.

\subsection{Simulation Data}
\label{suppsubsec: simulation data}

\begin{table*}[t]
\centering
\caption{Statistics of the collected simulation dataset. We utilize the modified RoboTwin environment to generate spatial traces across 16 manipulation tasks. The table details the number of episodes, total traces (including both end-effector–centric and object–centric traces), and the data collection time in hours (using 8 $\times$ NVIDIA H100 GPUs).}

\label{supptab:sim_tasks}
\small
\resizebox{\textwidth}{!}{
\begin{tabular}{@{}lccc|lccc@{}}
\toprule
Task Name               & Episodes & Traces & Collection Time (h) 
& Task Name                & Episodes & Traces & Collection Time (h) \\ \midrule
Click Bell              & 8827 & 17654 & 10.8  & Move Playingcard Away & 8104  & 16208 & 14.3 \\
Click Alarmclock        & 8222 & 16444 & 4.2   & Move Stapler Pad       & 12340 & 24680 & 27.2 \\
Blocks Ranking Size     & 8863 & 17726 & 41.6  & Open Laptop            & 3932  & 7864  & 20.3 \\
Blocks Ranking RGB      & 9135 & 18270 & 64.2  & Place A2B Left         & 7329  & 14658 & 16.0 \\
Handover Block          & 4234 & 8468  & 16.4  & Place A2B Right        & 6437  & 12874 & 16.2 \\
Handover Mic            & 7562 & 15124 & 27.9  & Place Bread Basket     & 4337  & 8674  & 17.2 \\
Hanging Mug             & 2577 & 5154  & 17.8  & Place Bread Skillet    & 1370  & 2740  & 6.9  \\
Move Can Pot            & 9275 & 18550 & 28.9  & Place Burger Fries     & 6005  & 12010 & 28.8 \\ \bottomrule
\end{tabular}
}
\end{table*}

We utilize RoboTwin 2.0~\cite{chen2025robotwin} to collect high-precision spatial traces in a simulation environment. 
\highlight{While the original RoboTwin data collection pipeline captures only end-effector positions, we modified the simulator to support our specific spatial trace requirements.}
Our modifications include: 
\textbf{(1)} Implementation of multi-GPU parallel simulation to accelerate data collection; 
\textbf{(2)} Extended tracking capabilities to record traces for both grippers and manipulated objects; 
\textbf{(3)} Refined instruction annotation, allowing a single execution frame to be mapped to multiple fine-grained sub-task instructions; 
\textbf{(4)} Incorporation of spatial referring expressions to localize objects, thereby enhancing the model's spatial understanding.
\highlight{We re-engineered 16 original RoboTwin tasks to generate the necessary spatial traces.} The statistics of the collected simulation data are listed in Tab.~\ref{supptab:sim_tasks}. The processing pipeline is as follows:

\vspace{+1mm}
\noindent \textbf{Coordinate Mapping.}
We define the functional grasp point by applying an offset of $0.14\,\text{m}$ along the $x$-axis of the robot's end-effector coordinate system to align with the actual gripper center. 
Using the camera intrinsic and extrinsic parameters, we project the 3D world coordinates of the grasp point into 2D pixel coordinates and depth values within the camera frame.

\vspace{+1mm}
\noindent \textbf{Task Segmentation and Trace Collection.}
During task definition, we annotate every simulation frame with sub-task labels. A single recorded frame may correspond to multiple active sub-tasks. 
To ensure geometric consistency, traces within a specific sub-task are projected into the coordinate system of that sub-task's start frame. The extraction process involves two steps:
\begin{enumerate}
    \item \textbf{Subject Identification:} We determine the active arm based on the instantaneous velocities of the left and right end-effectors. Similarly, the currently manipulated object is identified by calculating the velocities of all objects in the scene.
    \item \textbf{Trace Recording:} The motion of the active arm is recorded as the end-effector-centric trace, while the position history of the manipulated object constitutes the object-centric trace.
\end{enumerate}

\vspace{+1mm}
\noindent \textbf{Spatial Referring Generation.}
To enhance the model's capability in spatial reasoning and object grounding, we augment the simulation environment by algorithmically generating cluttered scenes. 
For each task, we instantiate random distractor objects around the target object within a defined radius. These distractors are sampled from a diverse asset library and include both distinct categories and identical categories to the target, necessitating fine-grained discrimination.
We develop a rule-based engine to synthesize unique referring expressions for the target object based on its spatial relationship with the environment. The engine parses the scene graph and selects the most concise unambiguous description from the following four categories:
\begin{itemize}
    \item \textbf{Object-to-Object Spatial Relations:} Describes the target's position relative to a unique reference object (\eg, \textit{``in front of''}, \textit{``behind''}, \textit{``to the left/right of''}, \textit{``next to''}).
    \item \textbf{Proximity Comparisons:} Describes the target based on its distance to reference objects or camera viewpoint (\eg, \textit{``closest to the [reference]''}, \textit{``farthest from the camera''}).
    \item \textbf{Ordinal Positioning:} Identifies the target based on its sorting order along the Cartesian axes (\eg, \textit{``leftmost''}, \textit{``rightmost''}, \textit{``frontmost''}, \textit{``the second from the left''}).
    \item \textbf{Attribute Comparisons:} Distinguishes the target from same-category distractors based on physical properties such as scale (\eg, \textit{``the largest''}, \textit{``the smallest''}).
\end{itemize}

\vspace{+1mm}
\noindent \textbf{QA Generation.}
Finally, after applying the RDP algorithm to downsample the traces, we obtain \textbf{443.1k} unique spatial traces (see Fig.~\ref{suppfig: robotwin_1}-~\ref{suppfig: robotwin_4}). Using template-based generation, we further produce \textbf{1.329M} QA pairs, including \textbf{914.1k }end-effector–centric and \textbf{415.1k }object-centric samples.

\section{{\bname} Details}
\label{suppsec: implementation_of_benchmark}

The {\bname} is a manually annotated, object-centric evaluation suite. 
To the best of our knowledge, it is the first benchmark designed to evaluate 3D spatial trace prediction for object-centric tasks. The data is sourced from complex indoor scenes in ScanNet~\cite{dai2017scannet} and CA-1M~\cite{lazarow2024cubify}, which \highlight{include precise geometry annotations (\eg, absolute depth maps, camera geometry, and 3D occupancy).}

\subsection{Benchmark Composition and Data}
Each scene in the benchmark is comprehensive and provides the following components:
\begin{itemize}
    \item \textbf{Visual Observation:} The source image and its corresponding absolute depth map.
    \item \textbf{Camera Parameters:} Full intrinsic camera parameters (\eg, intrinsic and extrinsic).
    \item \textbf{Object Definition:} A 2D mask identifying the specific object to be moved.
    \item \textbf{Target Destination:} A 3D bounding box specifying the object's final intended location.
    \item \textbf{Reference Trace:} A \textit{feasible} 3D object-centric movement path, represented as a series of 3D coordinates. We emphasize that this path is a reference, not a unique ground-truth; many other valid, collision-free paths to the destination may exist.
\end{itemize}

\noindent This data structure allows for rich 3D analysis. \highlight{Using the depth map and camera intrinsics, we can reconstruct a 3D occupancy map of the environment. Similarly, we can isolate the point cloud of the target object by combining its 2D mask with the depth map.}
Although {\bname} is an inherent 3D benchmark, its components can be projected into the 2D image plane. \highlight{We can project the 3D spatial trace and 3D destination box to create 2D paths and 2D boxes, enabling a rigorous evaluation of model performance on analogous 2D tasks.}

\subsection{Benchmark Statistics}
The benchmark consists of 100 manually annotated scenes. The distribution of data sources and task categories is summarized in Tab.~\ref{tab:benchmark_stats}.

\begin{table}[h]
\centering
\caption{Statistics of the {\bname} Benchmark.}
\label{tab:benchmark_stats}
\begin{tabular}{llr}
\toprule
\textbf{Criterion} & \textbf{Category} & \textbf{Count} \\
\midrule
\multirow{2}{*}{Data Source} & CA1M & 51 \\
 & ScanNet & 49 \\
\midrule
\multirow{2}{*}{Task Category} & Pick \& Place & 82 \\
 & Push \& Pull & 18 \\
\midrule
\textbf{Total} & & \textbf{100} \\
\bottomrule
\end{tabular}
\end{table}

\noindent The language prompts have an overall average length of 21.73 words. \highlight{As defined in our dataset, a ``step" refers to the number of spatial relations and associated objects within a single prompt.} As shown in Tab.~\ref{tab:step_stats}, there is a clear correlation: as the ``step'' count (\ie, task complexity) increases, the average word count of the prompt also increases, reflecting the need for more descriptive language.

\begin{table}[h]
\centering
\small
\caption{Benchmark Prompt Statistics by ``Step" Complexity.}
\label{tab:step_stats}
\begin{tabular}{ccc}
\toprule
\textbf{Step Count} & \textbf{Number of Prompts} & \textbf{Average Word Count} \\
\midrule
Step 2 & 7 & 10.14 \\
Step 3 & 16 & 14.81 \\
Step 4 & 16 & 17.19 \\
Step 5 & 28 & 22.29 \\
Step 6 & 21 & 27.48 \\
Step 7 & 7 & 30.86 \\
Step 8 & 5 & 34.60 \\
\bottomrule
\end{tabular}
\end{table}

\section{{\mname} Details}
\label{suppsec: implementation details}

\subsection{Architecture}
\label{suppsubsec: architecture}

We adopt NVILA~\cite{liu2024nvila} as the base model, including a visual encoder, an LLM, and a multimodal projector.
We further incorporate a universal spatial encoder and a scale decoder into the base model to complete our architecture.

\vspace{+1mm}
\noindent\textbf{Visual Encoder.} 
The image encoder in the base model is siglip-so400m-patch14-448~\cite{zhai2023sigmoid}, supporting $448 \times 448$ resolution for richer visual details.
Rather than simply resizing the image to a fixed resolution and producing the same number of tokens, this image encoder processes inputs at dynamic resolutions, yielding more visual tokens from higher-resolution images via finer patch division.
This enables fine-grained vision-language understanding, crucial for tasks like point prediction that require detailed perception beyond VQA.
%
%

\vspace{+1mm}
\noindent\textbf{Large Language Model.} 
We adopt the Qwen2 LLM backbone from NVILA~\cite{liu2024nvila}, which has been fully fine-tuned with extensive data during supervised training. This endows the model with rich visual knowledge, facilitating downstream 3D spatial understanding and reasoning tasks.

\vspace{+1mm}
\noindent\textbf{Universal Spatial Encoder.}
We leverage the encoder of the universal feed-forward metric 3D reconstruction model, MapAnything~\cite{keetha2025mapanything}, as our universal spatial encoder by removing the task-specific DPT and pose heads.
During training, the encoder is kept frozen to preserve its pretrained representations.
It provides strong 3D metric-grounded priors and flexibly accepts various geometric inputs (\eg, absolute depth, camera geometry). Incorporating more geometric cues leads to more accurate spatial representations, enhancing the VLM's 3D spatial awareness.

\vspace{+1mm}
\noindent \textbf{Scale Decoder.}
We introduce an MLP-based scale decoder that maps the hidden embedding of a special token, ``\texttt{<SCALE>}'', to a precise metric scale factor representing the ratio between predicted and real-world scale.
The decoder is initialized from the scale head of MapAnything~\cite{keetha2025mapanything} and is trainable throughout our training process.
By supervising the predicted scale with a regression loss, we enhance the model’s metric awareness, going beyond standard classification-based next-token prediction.

\vspace{+1mm}
\noindent \textbf{Multi-Modal Projector.} 
To align multi-modal representations (\eg, image to language, spatial representations to language, language to scale), we use linear connectors, the same as NVILA~\cite{liu2024nvila}, which is better than Q-Former, to allow the LLM to focus on fine-grained visual understanding and improve generalization. 
%

\subsection{Training Data}
\label{suppsubsec: training data}
Here we highlight the training data used at each stage, including the number of samples per dataset and the overall total.
See Tab.~\ref{supptab: dataset} for details.

\vspace{+1mm}
\noindent \textbf{SFT stage.}
Specifically, in the first step of the SFT stage, \ie, Metric Alignment, we not only train a spatial projector to align spatial representations and language space, but also train a scale projector with the scale decoder to predict the metric scale factor, using the {\dname} (RGB+$\mathcal{X}$) dataset with $4.5$M samples, where $\mathcal{X}$ indicates arbitrary combinations of geometric annotations (\eg, absolute depth map, camera geometry).
%
%
%
In the second step, \ie, Metric Enhancement, we use both {\dname} (RGB) and {\dname} (RGB+$\mathcal{X}$) datasets, yielding $9$M samples to enhance 3D spatial understanding and reasoning (\eg, \textit{3D spatial referring, measuring, tracing}).
To further improve instruction-following and basic capabilities, we incorporate auxiliary datasets: $965$k samples from instruction-tuned data (LLaVA-1.5~\cite{liu2024improved}, LRV~\cite{liu2024mitigating}), $321$k from referring datasets (RefCOCO/+/g~\cite{yu2016modeling}), $3$k from ShareRobotBench~\cite{ray2024sat} benchmark training sets, and $127$k from EmbSpatial~\cite{du2024embspatial} benchmark training sets.
These additions help bridge distribution gaps between {\dname} and benchmark-style queries. 
The total number of samples used in this stage reaches $14$M.

\vspace{+1mm}
\noindent \textbf{RFT Stage.}
In the RFT stage, we train the model using {\dname} data annotated with detailed reasoning processes, including key intermediate steps (\eg, \textit{3D spatial referring, measuring, scale}) and final answers.
To ensure both training efficiency and effective learning, we use moderately difficult samples (typically involving $4-6$ reasoning steps), resulting in a $120$k-sample dataset.

\begin{table*}[t]
\caption{
Details about the training datasets used in the SFT and RFT stages.
M.A. and M.E denote the Metric Alignment and Metric Enhancement step in the SFT stage, respectively.}
\tiny
\centering
\begin{tabular}{l|l|l}
\toprule                          

Stage & Categories & Datasets \\
\midrule
SFT (M.A) & Spatial & {\dname} (RGB+$\mathcal{X}$) \\
\midrule
\multirow{3}{*}{SFT (M.E)} & Spatial & {\dname} (RGB), {\dname} (RGB+$\mathcal{X}$), ShareRobotBench~\cite{ji2025robobrain}, EmbSpatial~\cite{du2024embspatial} \\
 & General & COCO~\cite{lin2014microsoft}, GQA~\cite{hudson2019gqa}, OCR-VQA~\cite{mishra2019ocr}, TextVQA~\cite{singh2019towards}, VG~\cite{krishna2017visual}, LRV~\cite{liu2024mitigating} \\
 & REC & RefCOCO/+/g~\cite{yu2016modeling} \\
\midrule
RFT & Spatial & {\dname} (RGB+$\mathcal{X}$) w/ Reasoning Processing \\

\bottomrule[1pt]
\end{tabular}
\label{supptab: dataset}
\end{table*}

\subsection{SFT Training Details}
\label{suppsubsec: SFT training details}

We formulate the SFT training stage as follows: given a dataset $\mathcal{D}$ consisting of samples in the form of triplets ($\mathcal{O}$, $\mathcal{Q}$, $\mathcal{A}$), where $\mathcal{O}$ is a sensor image with geometry information (either RGB or RGB+$\mathcal{X}$), $\mathcal{Q}$ is a textual question, and $\mathcal{A}$ is the corresponding answer.
The answer $\mathcal{A}$ may be a direct response (\eg, a sequence of 3D point coordinates) or include intermediate reasoning steps (\eg, key perceptual evidence followed by the final answer, such as \textit{3D spatial referring and measuring}). 
The training objective is to maximize the likelihood of generating the answer given the input pair ($\mathcal{Q}$, $\mathcal{A}$):

{
\begin{equation}
\mathcal{L}_{\mathrm{SFT}}=-\mathbb{E}_{(\mathcal{O}, \mathcal{Q}, \mathcal{A})\sim\mathcal{D}}\sum_{t=1}^T\log\pi_\theta(y_t\mid \mathcal{O}, \mathcal{Q}, y_{<t}),
\end{equation}
}

\noindent where $\pi_\theta$ is the model’s token distribution. 
The output model $\pi_\mathrm{SFT}$ serves as the initialization for the next RFT stage, ensuring a robust foundation for reinforcement learning.
To be specific, our SFT consists of two steps. 
In the first step, Metric Alignment, only the spatial projector, scale projector, and scale decoder are updated by using the {\dname}~(RGB+$\mathcal{X}$). We employ a maximum learning rate of 1e-4, a weight decay of 0, and a warm-up ratio of 0.03. 
The 2B variant is trained with a batch size of 7 per GPU, and the 8B variant with 3, both for one epoch.
In the second step of Metric Enhancement, we fine-tune all model parameters without the universal spatial encoder using the datasets described in Sec.~\ref{suppsubsec: training data}.
Training is conducted for one epoch with a maximum learning rate of 5e-5. We use a batch size of 6 per GPU for the 2B model and 2 for the 8B model. 
Other hyperparameters follow those in the first step.
For more details, please refer to NVILA~\cite{liu2024nvila} settings during alignment and SFT.

\subsection{RFT Training Details}
\label{suppsubsec: RFT training details}

During the RFT stage, we refine $\pi_\mathrm{SFT}$ via GRPO~\cite{shao2024deepseekmath}, a reinforcement learning method designed for efficiency and scalability. 
Unlike PPO~\cite{schulman2017proximal}, which relies on a costly value network, GRPO estimates relative advantages by comparing intra-group rewards, reducing computation, and simplifying optimization. 
This makes it well-suited for reasoning-intensive spatial tracing tasks.
In detail, we modify R1-V~\cite{zhang2025r1} to support our 3D-aware architecture.
Training is conducted for two epochs with a batch size of 1 per GPU and 8 outputs in GRPO.
For details about hyperparameters, see R1-V~\cite{zhang2025r1}.

\subsubsection{Sampling Action Groups}
Given an input state \( s = (\mathcal{O}, \mathcal{Q}) \), where \( \mathcal{O} \) denotes the visual encoding of the RGB or RGB+$\mathcal{X}$ observation and \( \mathcal{Q} \) the textual encoding of the question, GRPO samples a set of actions \( \{a_1, a_2, \dots, a_N\} \) from the current policy \( \pi_\theta \), initialized from \( \pi_{\mathrm{SFT}} \).
The sampling process is:

{
\begin{equation}
a_i\sim\pi_\theta(a\mid \mathcal{O}, \mathcal{Q}),\quad\mathrm{for~}i=1,2,\ldots,N
\tag{2}
\end{equation}
}

\noindent This strategy ensures diverse responses, promoting exploration and preventing premature convergence.

\subsubsection{Reward Design and Policy Update}
\label{suppsubsubsection: reward design and policy update}
Each sampled action $a_i$ is assigned a reward $ R (a_i) $ based on verifiable criteria, yielding a reward set $ r_1, r_2, \ldots, r_N$. 
For spatial tracing tasks, $R(a_i)$ integrates three outcome-based and our proposed two process-based components.
The outcome-based reward functions are defined as follows: 

\vspace{+1mm}
\noindent \textbf{Outcome Format Reward $R_{OF}$.}
This component ensures structured and interpretable outputs by requiring the model to a predefined format: reasoning within ``$\texttt{<think>}\ldots\texttt{</think>}$'' and the final answer in ``$\texttt{<answer>}\ldots\texttt{</answer>}$''. A reward is assigned 1 for strict compliance, 0 otherwise.

\vspace{+1mm}
\noindent \textbf{Point Reward ($R_P$).}
This component primarily evaluates the consistency between the predicted trace \(\hat{\tau} = \{\hat{p_t}\}_{t=1}^{T}\)’s start and end points \((p_1, p_T)\) and those of the ground-truth trace \(\tau\) from the annotations of {\dname}.
This process can be formulated as:

\begin{align}
    &R_P = \tfrac{1}{2}\bigl[f(p_1, \hat{p}_1) + f(p_T, \hat{p}_T)\bigr],
    \\
    &f(p, p') = \max\bigl(0,\; 1 - \|p - p'\|_2^2\bigr)
\end{align}

\vspace{+1mm}
\noindent \textbf{Trace Reward ($R_P$).}
This component measures the consistency between the predicted trace \(\hat{\tau}\) and the ground-truth trace \(\tau\), using a distance metric \(d(\tau, \hat{\tau})\) (\eg, Dynamic Time Warping).
This process can be formulated as:
\[
d(\tau, \hat{\tau}): R_{\text{T}} = \max\bigl(0,\; 1 - d(\tau, \hat{\tau})\bigr)
\] 

\noindent All \((u,v,d)\) points are normalized to the interval \([0,1]\), with the depth dimension scaled by the maximum scene depth.


\vspace{+1mm}
\noindent Nevertheless, the outcome-based rewards described above are metric-agnostic and thus fail to provide explicit supervision for the crucial perceptual evidence (\eg, \textit{3D spatial measuring and referring}) involved in trace generation. 
Accordingly, we need process rewards to more effectively guide the model.
Howeever, most process-based reward mechanisms rely on a Process Reward Model (PRM), typically a fine-tuned large language model (LLM) or vision–language model (VLM) responsible for providing feedback. 
However, integrating such an approach into our framework presents two main challenges. 
\textbf{(1)} Since LLMs cannot process images, they are unable to verify whether the predicted coordinates correspond to the intended object. 
\textbf{(2)} Although VLMs combine visual and textual inputs, prior work~\cite{majumdar2024openeqa} indicates that they may fail to exhibit precise visual understanding when interpreting textual coordinates or metric estimation. Because accurate verification of predicted coordinates or metric approximation is essential for reward assignment, additional or specialized methods must be employed to ensure reliable feedback.


\vspace{+1mm}
\noindent \highlight{To address this issue, we introduce a metric-sensitive rule-based process reward for spatial tracing, which eliminates the need for a Process Reward Model.} 
Specifically, our method directly evaluates crucial intermediate perceptual steps by leveraging the ground-truth stepwise annotations available in {\dname}. This approach differs from most existing methods that focus on process-based rewards~\cite{khalifa2025process, liu2025can}, which typically assume strictly sequential reasoning and rely on dedicated Process Reward Models (PRMs) for evaluating intermediate outputs.
\highlight{In contrast, we employ \textit{metric-sensitive} rule-based process reward functions to evaluate intermediate perceptual outcomes in an \textit{order-invariant} fashion.}
Our key insight is twofold:
\begin{itemize}
    \item \highlight{\textit{(1) Metric-sensitivity}}: Different spatial attributes (\eg, 3D spatial referring requires 3D points with depth information, while 3D spatial measuring demands precise numerical predictions, including scale) all involve metric scale and require distinct reward formulations due to their inherently different representations.

    \item \highlight{\textit{(2) Order-invariance}}: The reasoning process in spatial tracing is not strictly sequential; for instance, identifying the placement position of the keyboard or the mouse first does not affect the final interpretation of ``the free area between the keyboard and the mouse''.
\end{itemize}

\noindent By using this rule-based yet flexible strategy, we overcome the limitations of relying on rigid sequential processes, instead allowing for more robust and adaptable spatial reasoning. We have two process-based reward functions:


\vspace{+1mm}
\noindent \textbf{Process Format Reward $R_{PF}$.}
Similar to the Outcome Format Reward strategy, this component enforces a structured and interpretable reasoning process, thereby facilitating accurate reward computation. In particular, the model is required to produce outputs in the following format:

\begin{equation}
\footnotesize
    \texttt{[Perception Type] [Target Object]: [Value]}
\end{equation}

\noindent where ``\texttt{Perception Type}'' must be one of three categories: ``\texttt{Referring}'', ``\texttt{Measuring}'', or ``\texttt{Scale}''.
The ``\texttt{Target Object}'' corresponds to a uniquely identifiable entity (\eg, ``\texttt{the second leftmost cup}'' or ``\texttt{the second large cup from left to right}''). 
The “\texttt{Value}” depends on the selected ``\texttt{Perception Type}'':

\begin{itemize}
  \item For ``\texttt{Referring}'', the value should be a normalized 2D coordinate with absolute depth of the form \texttt{[(u, v, d)]}, where both \(u\) and \(v\) lie in the interval \([0, 1000]\), and \(d\) rounded to three decimal places.  
  \item For ``\texttt{Measuring}'', the value represents a scalar indicating an object’s length, width, or height, rounded to three decimal places, expressed in potentially different but appropriate units, as discussed in Sec.~\ref{suppsubsubsec: Human-Like Measuring Descriptions Generation}.

  \item For ``\texttt{Scale}'', the value is a scalar representing the scale ratio between the predicted scene and the real-world scale, rounded to three decimal places.  
\end{itemize}

\noindent Below are examples to illustrate the expected format:

\begin{itemize}
  \item \texttt{[Position] [the second largest cup]}: [(245, 147, 1.837)]
  \item \texttt{[Measuring] [the height of the second largest cup]}: 20 centimeters
  \item \texttt{[Scale] [Scene]}: 2.342
\end{itemize}

\vspace{+1mm}
\noindent \textbf{Accuracy Reward $R_{Acc}$.}
The reward is computed only for steps annotated as key steps in {\dname}. 
In detail, we use regex matching to determine whether the ``\texttt{Target Object}'' in the current process format appears in the key-step annotations. 
If not, the step receives no reward.
Since the model has already undergone a cold-start phase in SFT, it can interpret instructions and identify relevant target objects. 
Thus, a failed match implies that the model cannot accurately refer to the object linguistically, and no reward is assigned.
For each perception type, we apply a specific metric to compute the reward:

\begin{itemize}
    \item ``\texttt{Referring}'': For 3D point modeling \((u, v, d)\), we evaluate each component separately. For the 2D coordinates \((u, v)\), we compute the L1 distance between the predicted and ground-truth points. If the error is within $10$\% of the image’s longer side, a reward of $0.5$ is assigned; otherwise, $0$. For the depth \(d\), if the predicted value falls within $\pm30$\% of the ground truth, a reward of $0.5$ is given; otherwise, $0$.

    \item ``\texttt{Measuring}'': If the predicted value falls within $\pm30$\% of the ground truth, the reward is $1$; otherwise, $0$.
    
    \item ``\texttt{Scale}'': If the predicted value falls within $\pm30$\% of the ground truth, the reward is $1$; otherwise, $0$.
\end{itemize}

\noindent We prioritize the correctness of the final outcome over intermediate steps. 
To prevent reward accumulation from multi-step processes, we scale the process reward by $0.25$. The final reward function is defined as:
\begin{align}
    r_i = R_{OF}(a_i) + R_{P}(a_i) + \alpha R_{PF}(a_i) + \alpha R_{Acc}(a_i)
\end{align}


\noindent where $\alpha$ is set to $0.25$.
By normalizing the rewards within the sampled group, we obtain the set of relative advantages \(\{A_1, A_2,\ldots, A_N\}\) defined as

\begin{align}
A_i = \frac{r_i - \text{mean}(\{r_j\})}{\text{std}(\{r_j\})},
\end{align}

\noindent which measures how each reward compares to the mean in units of standard deviation. We then update the policy based on these advantages, reinforcing actions with higher relative advantages while reducing the likelihood of those deemed less effective.
To ensure stable reinforcement learning, the update is further constrained by minimizing the KL divergence between the updated policy and its reference counterpart, thereby promoting incremental and controlled policy adjustments.

\section{Experimental Setting and Details}
\label{suppsec: experimental setting and details}
\subsection{Experiments Compute Resources}
\label{suppsubsec: compute resources}

We conduct experiments on an H100 GPU cluster, with each node equipped with 8 GPUs.

\vspace{+1mm}
\noindent \textbf{Object-centric Spatial Trace Generation from 3D Scanning Data.}
Given that the trace generation pipeline relies heavily on computationally intensive geometric operations (\eg, RRT* planning, OBB-SAT collision detection, and visibility checks) rather than neural network inference, we utilize high-performance CPU parallelization without GPU acceleration.
The generation and refinement pipeline is executed on a single high-end server equipped with dual Intel(R) Xeon(R) Platinum 8468 CPUs (totaling 96 physical cores and 192 threads).
We configure the pipeline to utilize 180 concurrent processes to maximize throughput.
The most time-consuming components are the RRT* obstacle-aware planning, where we set a high iteration count to ensure the generation of complex, high-difficulty traces, and the rendering of visualization assets for quality verification.
Consequently, the complete generation, cleaning, and visualization process takes approximately 32 hours for CA-1M and 6 hours for ScanNet.

\vspace{+1mm}
\noindent \textbf{Real-world Manipulation Video.}
For Droid, we use the ZED SDK to extract depth and camera intrinsics on a single NVIDIA H100 GPU, which takes 46 hours in total.
For the data processing pipelines of Droid and AgiBot, we use 64-way parallel processing, which takes 5 hours and 27 hours, respectively.
These processing times include the generation of visualization images and videos.

\vspace{+1mm}
\noindent \textbf{Simulation Manipulation Video and Evaluation.}
The data collection time for the RoboTwin simulation is reported in Tab.~\ref{supptab:sim_tasks}.
For the RoboTwin simulation evaluation of our model, we use 8 NVIDIA H100 GPUs, with a total runtime of 36 hours.
We further observe that RoboTwin executes significantly slower on H100 GPUs compared to RTX 4090 GPUs.

\vspace{+1mm}
\noindent \textbf{Metric Alignment in SFT.}
The process is conducted on 8 nodes over 20 hours for 2B variants and 8 nodes over 50 hours for 8B variants. Both variants training use ZeRO-3.

\vspace{+1mm}
\noindent \textbf{Metric Enhancement in SFT.}
The process is conducted on 8 nodes over 2 days for 2B variants and 8 nodes over nearly 1 week for 8B variants. Both variants training use ZeRO-3.

\vspace{+1mm}
\noindent \textbf{Spatial Tracing in RFT.}
The process is conducted on 1 node over 3 days for 2B variants.
However, our model is over twice as slow as other Qwen 2.5-VL-based methods~\cite{zhang2025r1, shen2025vlm}, mainly because they process only a single RGB image during training and can leverage vLLM for group inference acceleration. In contrast, our method requires RGB+$\mathcal{X}$ inputs and modifies the original NVILA architecture, making it incompatible with vLLM or SGLang acceleration.

\subsection{Spatial Understanding Benchmarks}
\label{suppsubsec: spatial understanding benchmarks}

We evaluate several publicly available \textit{spatial understanding} benchmarks, strictly adhering to their official evaluation protocols. 
These benchmarks include CV-Bench~\cite{tong2024cambrian} (covering 2D Spatial Relation, 3D Depth Order, and 3D Distance), the BLINK~\cite{fu2024blink} validation set (Spatial Relation, Relative Depth), RoboSpatial~\cite{song2024robospatial} (configuration), and EmbSpatial~\cite{du2024embspatial}. 
We omit tasks that are not directly related to spatial understanding—such as the 2D Counting task in CV-Bench and the Art Style or IQ Test tasks in BLINK—from our analysis. 
As all the selected benchmarks employ multiple-choice formats, we report accuracy as the evaluation metric.
We compare three categories of models in our experiments:
\begin{enumerate}
    \item Proprietary VLMs (\eg, Gemini-2.5-Pro~\cite{team2023gemini}), which have demonstrated strong spatial perception, as discussed in Gemini-Robotics~\cite{team2025gemini}.

    \item Open-source VLMs trained on general VQA datasets (\eg, NVILA~\cite{liu2024nvila}, Qwen3-VL~\cite{qwen3technicalreport}).

    \item Spatial specialist models trained specifically on spatially relevant datasets, such as SpatialBot~\cite{cai2024spatialbot} and RoboBrain 2.0~\cite{team2025robobrain}, which offer fundamental spatial understanding capabilities.
\end{enumerate}

\subsection{Spatial Measuring Benchmarks}
\label{suppsubsec: spatial measuring benchmarks}

We evaluate our model's spatial measurement capabilities on three distinct benchmarks: QSpatial~\cite{liao2024reasoningpathsreferenceobjects} (Plus, ScanNet) and MSMU~\cite{chen2025sdvlm} bench. 
To determine the success of a single question-answering instance, we adopt the evaluation protocol established in the original QSpatial paper. 
First, both the model's predicted value and the ground-truth answer are converted to centimeters (cm). An answer is considered successful if the ratio of the predicted value to the ground-truth value falls within the range of [0.5, 2.0]. 
Formally, let $v_{pred}$ be the predicted value in cm and $v_{gt}$ be the ground-truth value in cm. The prediction is marked as successful if:
$ 0.5 \le \frac{v_{pred}}{v_{gt}} \le 2.0 $.

\subsection{2D Spatial Referring Benchmarks}
\label{suppsubsec: spatial referring benchmarks}

We strictly follow the official evaluation protocols of several publicly available \textit{2D spatial referring} benchmarks, including Where2Place~\cite{yuan2024robopoint}, RoboSpatial~\cite{song2024robospatial}, and RefSpatial-Bench~\cite{zhou2025roborefer}. 
These benchmarks gauge a model’s ability to predict one or more points given an image and a textual instruction. 
The primary evaluation metric is the average success rate, which quantifies the fraction of predicted points that lie within the ground-truth mask for each sample.
We compare two categories of models in our experiments:
\begin{enumerate}
    \item General VLMs trained on general VQA datasets (\eg, Gemini-2.5-Pro~\cite{team2023gemini}, Qwen3-VL~\cite{qwen3technicalreport}).

    \item Referring specialist models trained specifically on spatial referring/grounding datasets, such as Molmo~\cite{deitke2025molmo} and RoboPoint~\cite{yuan2024robopoint}, which offer fundamental 2D point referring capabilities.
\end{enumerate}

\subsection{2D Visual Trace Benchmarks}
\label{suppsubsec: 2D visual tracing benchmarks}

We evaluate our method on public 2D visual trace benchmarks, including ShareRobotBench-V~\cite{ji2025robobrain} and VABench-V~\cite{yuan2025seeing}.
ShareRobotBench-V focuses on end-effector-centric 2D visual traces, while VABench-V emphasizes object-centric 2D visual traces.
Unlike 3D spatial traces, 2D visual traces lack depth information and cannot assess collision-free properties.
Therefore, these benchmarks evaluate the similarity between predicted traces and ground-truth references using metrics such as Discrete Fréchet Distance, Hausdorff Distance, and Root Mean Square Error, where lower values indicate better performance.
We compare two categories of models in our experiments:
\begin{enumerate}
    \item General VLMs trained on general VQA datasets (\eg, Qwen3-VL~\cite{qwen3technicalreport}).

    \item 2D visual trace specialist models trained specifically on tracing datasets, such as MolmoAct~\cite{lee2025molmoact} and Embodied-R1~\cite{yuan2025embodied}, which offer fundamental 2D visual trace generation capabilities.
\end{enumerate}

\subsection{Spatial Tracing Benchmarks}
\label{suppsubsec: spatial tracing benchmarks}

We evaluate model performance in both 2D and 3D for spatial tracing tasks.

\subsubsection{2D Evaluation}
For 2D evaluation, all 3D ground-truth attributes (\eg, generated 3d spatial trace, ground-truth 3D bounding box for object's final intended location) are projected into the 2D image plane. We use the following metrics:

\begin{itemize}
    \item \textbf{2D Start Success:} The predicted starting point must be located in the ground-truth 2D mask of the target object.
    \item \textbf{2D End Success:} At least one of the final three (3) predicted points in the projected 2D trace must be located inside the projected 2D bounding box of the destination.
\end{itemize}

\subsubsection{3D Evaluation}
Given the high difficulty of the 3D task, we adopt slightly more lenient spatial thresholds:
\begin{itemize}
    \item \textbf{3D Start Success:} The predicted 3D starting point must be in a 20cm distance of the target object's point cloud.
    \item \textbf{3D End Success:} At least one of the final three (3) predicted 3D points must be within a 20cm distance of the 3D destination bounding box.
    \item \textbf{Overall 3D Success:} This composite metric requires a task to first achieve both \textbf{3D Start Success} and \textbf{3D End Success}. If it passes, we then simulate the movement of the object's point cloud along the predicted trace. The path is considered successful if, during this movement, no more than 20\% of the object's points intersect with the environmental 3D occupancy map (\ie, it is collision-free).
\end{itemize}

\subsubsection{Baseline Performance}

This benchmark proves to be highly challenging. Even with these metrics, advanced baselines like Gemini-2.5-Pro and Qwen-3VL achieve an Overall 3D Success rate of less than 10\%. In contrast, our model demonstrates significantly stronger performance, exceeding a 30\% success rate.

\subsection{Simulation Evaluation}
\label{suppsubsec: simulation evaluation}

We evaluate the capability of {\mname} to support embodied tasks through spatial trace generation in the RoboTwin simulation environment. Specifically, we select 19 tasks whose behaviors can be intuitively specified using spatial traces--12 tasks have variants that appear in {\dname}, while 7 tasks are entirely novel. It is worth noting that the end-to-end model~\cite{ACT,DP,DP3,RDT,Pi0} results on these tasks are taken from the RoboTwin 2.0 online leaderboard \footnote{\href{https://robotwin-platform.github.io/leaderboard}{https://robotwin-platform.github.io/leaderboard}}  (November 2025). All end-to-end models on the leaderboard are trained or fine-tuned for a single specific task setting and are evaluated in the same task environment. In contrast, the simulation data in {\dname} were collected in newly designed scenes with our own instructions. 

Our evaluation environment configuration follows that of the RoboTwin 2.0 VLA benchmark (demo\_randomized). For \textit{Click Bell} and \textit{Click Alarmclock}, we generate end-effector–centric traces, and the gripper executes the predicted trace after closing.
 For all remaining tasks, we generate object-centric traces: the gripper grasps the object corresponding to the starting point of the trace, follows the model-generated trace, and places the object at the final point.
 Both grasping and placing operations are implemented using the RoboTwin API.
 For complex multi-stage tasks, we decompose the instruction into several segments and execute them sequentially. The detailed evaluation results are provided in Tab.~\ref{supptab: robotwin eval}, and the visualization of the evaluation process are provided in Fig.~\ref{suppfig: robotwin_eval1}-\ref{suppfig: robotwin_eval6}. \highlight{Under the traces predicted by our model, all tasks can be executed with high success rates, demonstrating both the strong spatial understanding capability of our model and its effectiveness in supporting embodied tasks.}

We construct the prompts used for RoboTwin evaluation following the template below:

\begin{lstlisting}[basicstyle=\ttfamily\footnotesize, backgroundcolor=\color{myblue!50}, caption={Evaluation Prompts.}, captionpos=t, breaklines=true, label={lst:prompts_robotwin}]
Please predict 3D \{object-centric | end-effector-centric\} visual trace to complete the task successfully. The task is "<instruction>". Your answer should be formatted as a list of tuples, i.e., [(x1, y1, d1), (x2, y2, d2), ...], where each tuple contains the x and y coordinates and the depth of the point.
]
\end{lstlisting}

\begin{table*}
\caption{Performance on RoboTwin hard tasks. We report the success rate (\%) compared to end-to-end and VLM-based models. 
Gray rows indicate the task variants that are not present in {\dname}.}
\label{supptab: robotwin eval}
\centering
\tiny
\begin{tabular}{l|ccccc|cc|c}
\toprule
\multirow{2}{*}{Task} & \multicolumn{5}{c|}{\cellcolor{myred}End-to-End Policy} & \multicolumn{2}{c|}{\cellcolor{mygreen}Vision-Language Model} & \cellcolor{myblue}Ours \\ 
& \cellcolor{myred}ACT & \cellcolor{myred}DP & \cellcolor{myred}DP3 & \cellcolor{myred}RDT & \cellcolor{myred}$\pi_0$ & \cellcolor{mygreen}Qwen3-VL-8B & \cellcolor{mygreen}Gemini-2.5-Pro & \cellcolor{myblue}{\mname-2B} \\ 
\midrule
Click Bell             & 3 & 0 & 0  & 9  & 3  & 0 & 0 & \textbf{96} \\
Click Alarmclock       & 4 & 5 & 14 & 12 & 11 & 0 & 0 & \textbf{79} \\
Blocks Ranking Size    & 0 & 0 & 0  & 0  & 1  & 0 & 0 & \textbf{89} \\
Blocks Ranking RGB     & 0 & 0 & 0  & 0  & 5  & 0 & 0 & \textbf{96} \\
Move Can Pot           & 4 & 0 & 6  & 12 & 21 & 0 & 0 & \textbf{27} \\
Move Playingcard Away  & 0 & 0 & 3  & 11 & 22 & 0 & 5 & \textbf{94} \\
Move Stapler Pad       & 0 & 0 & 0  & 0  & 2  & 0 & 0 & \textbf{18} \\
Place A2B Left         & 0 & 0 & 2  & 1  & 1  & 0 & 2 & \textbf{84} \\
Place A2B Right        & 0 & 0 & 0  & 1  & 6  & 0 & 1 & \textbf{93} \\
Place Bread Basket     & 0 & 0 & 1  & 2  & 4  & 0 & 0 & \textbf{82} \\
Place Bread Skillet    & 0 & 0 & 0  & 1  & 1  & 0 & 0 & \textbf{48} \\
Place Burger Fries     & 0 & 0 & 18 & 27 & 4  & 0 & 0 & \textbf{99} \\

\rowcolor{gray!10}
Place Container Plate  & 1 & 0 & 1  & 17 & 45 & 0 & 0 & \textbf{52} \\
\rowcolor{gray!10}
Place Empty Cup        & 0 & 0 & 1  & 7  & 11 & 0 & 0 & \textbf{85} \\
\rowcolor{gray!10}
Place Fan              & 0 & 0 & 1  & 2  & 10 & 0 & 2 & \textbf{66} \\
\rowcolor{gray!10}
Place Mouse Pad        & 0 & 0 & 1  & 0  & 1  & 0 & 0 & \textbf{30} \\
\rowcolor{gray!10}
Place Object Stand     & 0 & 0 & 0  & 5  & 11 & 0 & 0 & \textbf{38} \\
\rowcolor{gray!10}
Stack Blocks Two       & 0 & 0 & 0  & 2  & 1  & 0 & 0 & \textbf{33} \\
\rowcolor{gray!10}
Stamp Seal             & 0 & 0 & 0  & 0  & 4  & 0 & 0 & \textbf{7}  \\ 
\bottomrule
\end{tabular}
\end{table*}

\subsection{Real-world Evaluation}
\label{suppsubsec: real-world evaluation}

\subsubsection{UR5 Manipulation}

We show the demo for UR5 \footnote{\href{https://www.universal-robots.com/products/ur5e/}{https://www.universal-robots.com/products/ur5e/}} Manipulation with human disturbance.
In this case, {\mname} runs at 1.5Hz. 
Significant shifts in the predicted end-effector-centric spatial trace endpoint, particularly at the 2D pixel level  $(u, v)$, can trigger motion interruption and re-planning.
Specifically, for grasping, the 2D pixel level $(u, v)$ of the predicted end-effector-centric spatial trace endpoint is fed into SAM2~\cite{ravi2024sam} to generate a segmentation mask, which filters the target object's point cloud from the scene captured by a third-person Intel RealSense L515~\footnote{\href{https://www.intelrealsense.com/lidar-camera-l515/}{https://www.intelrealsense.com/lidar-camera-l515/}} depth camera. 
The extracted point cloud is input to AnyGrasp~\cite{fang2023anygrasp} to predict a grasp pose in the camera coordinate frame. 
Using an eye-to-hand calibration method, the grasp pose is transformed into the UR5 robot's base frame for execution.
For final placement, the 2D pixel level $(u, v)$ of the predicted end-effector-centric spatial trace endpoint is also converted to 3D coordinates using the camera's intrinsic parameters and depth data. 
The 3D point is then transformed into the robot’s coordinate system to identify the final placement loccation.
Notably, during grasping and final placement, we do not directly use the depth predicted by the spatial trace.
\highlight{Since these points (\eg, grasp point, placement point) lie on contact surfaces (\eg, table, target object), more accurate values can be obtained using known camera intrinsics and observed depth data, enabling robust pick-and-place operations. 
In contrast, during movement, the waypoints are in mid-air without contact surfaces, so we rely on the model's predicted 3D points.
The robotic arm's end-effector is guided to sequentially reach these 3D waypoint positions for manipulation.
}

\subsubsection{G1 Humanoid Manipulation}

The main pipeline is the same as UR5 Manipulation.
For grasping, we employ a head-mounted Intel RealSense D435\footnote{\href{https://www.intelrealsense.com/depth-camera-d435/}{https://www.intelrealsense.com/depth-camera-d435/}} on the Unitree G1 humanoid to capture RGB-D images, which are processed by {\mname} (\ie, RGB images, depth maps, camera intrinsic as model input) to extract 2D target coordinate $(u, v)$ from the predicted end-effector-centric spatial trace endpoint. 
These coordinates guide SAM2~\cite{ravi2024sam} to generate a segmentation mask, which filters the third-person D435 point cloud to isolate the target object. 
The filtered point cloud is then passed to AnyGrasp~\cite{fang2023anygrasp} to predict a grasp pose in the third-person frame, which is transformed to the robot’s base frame using known camera-to-robot calibration.
%
%
%
Since the waypoints during robotic watering are in mid-air and not on a surface, 2D-to-3D projection using RGB-D cameras is not applicable. 
Therefore, the robot directly moves the end-effector to the predicted 3D spatial waypoints for manipulation.

\section{More Demonstrations}
\label{suppsec: more demonstrations}

\begin{figure*}[h]
    \centering
    
    \includegraphics[width=0.8\linewidth]{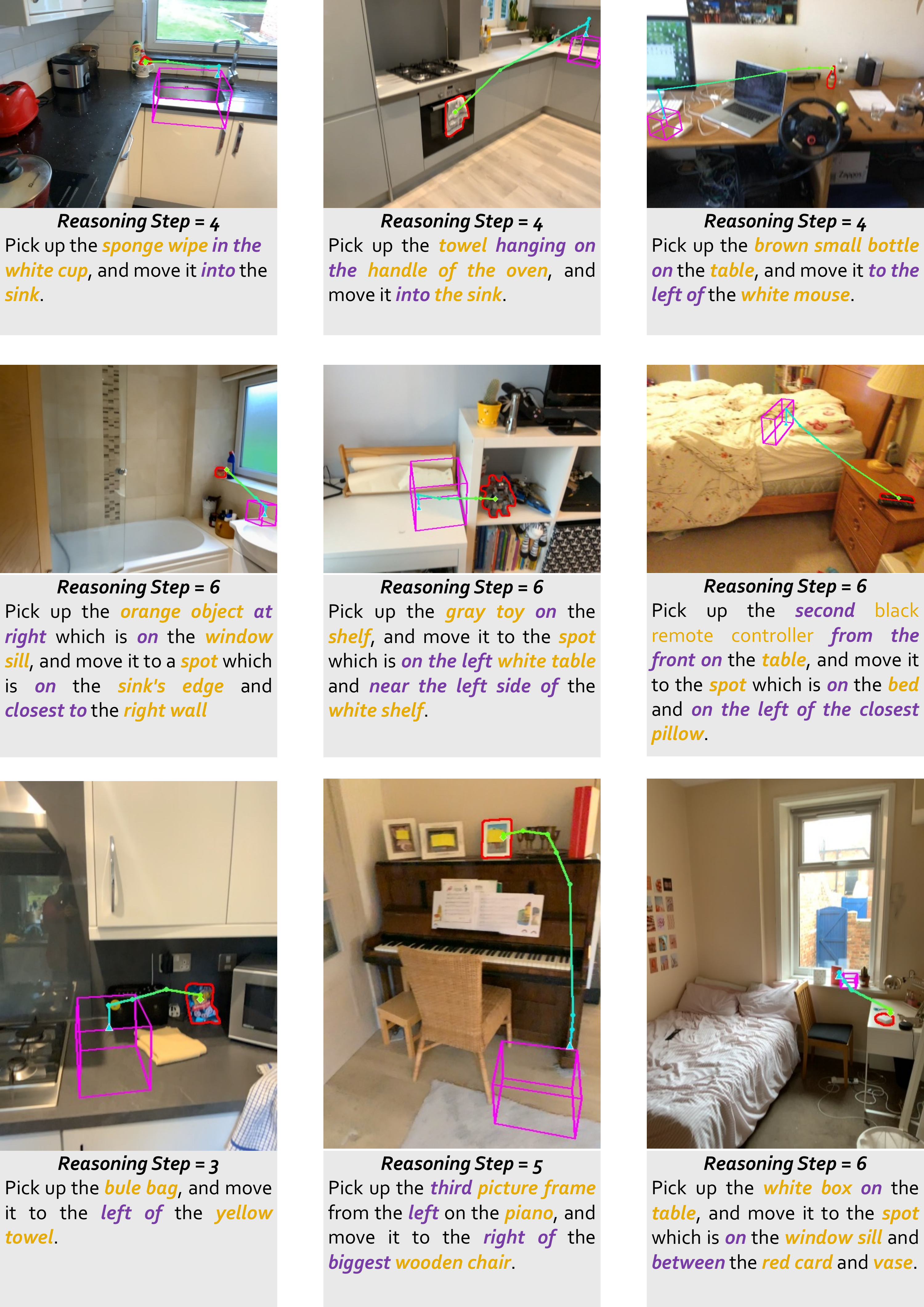}
    

    \caption{Visualization of TraceSpatial-Bench and {\mname}'s rollouts. The red mask indicates the ground-truth starting point, and the purple 3D bounding box denotes the ground-truth endpoint. We show the 2D projection of {\mname}'s predicted trace.}
    
    \label{suppfig: TraceSpatial-Bench_vis1}
\end{figure*}

\begin{figure*}[h]
    \centering
    
    \includegraphics[width=0.8\linewidth]{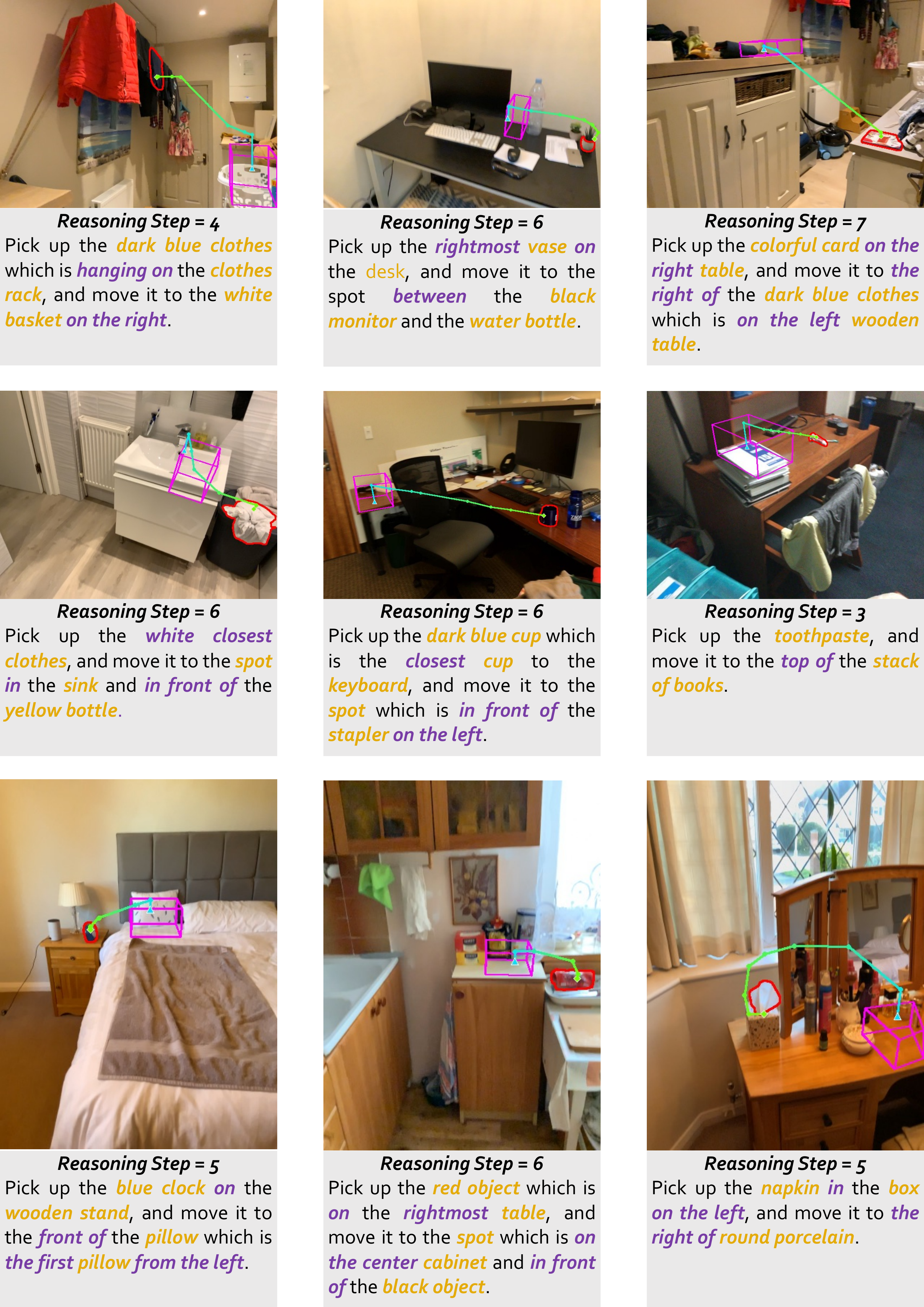}
    
    \caption{Visualization of TraceSpatial-Bench and {\mname}'s rollouts. The red mask indicates the ground-truth starting point, and the purple 3D bounding box denotes the ground-truth endpoint. We show the 2D projection of {\mname}'s predicted trace.}
    
    \label{suppfig: TraceSpatial-Bench_vis2}
\end{figure*}

\begin{figure*}[h]
    \centering
    
    \includegraphics[width=0.8\linewidth]{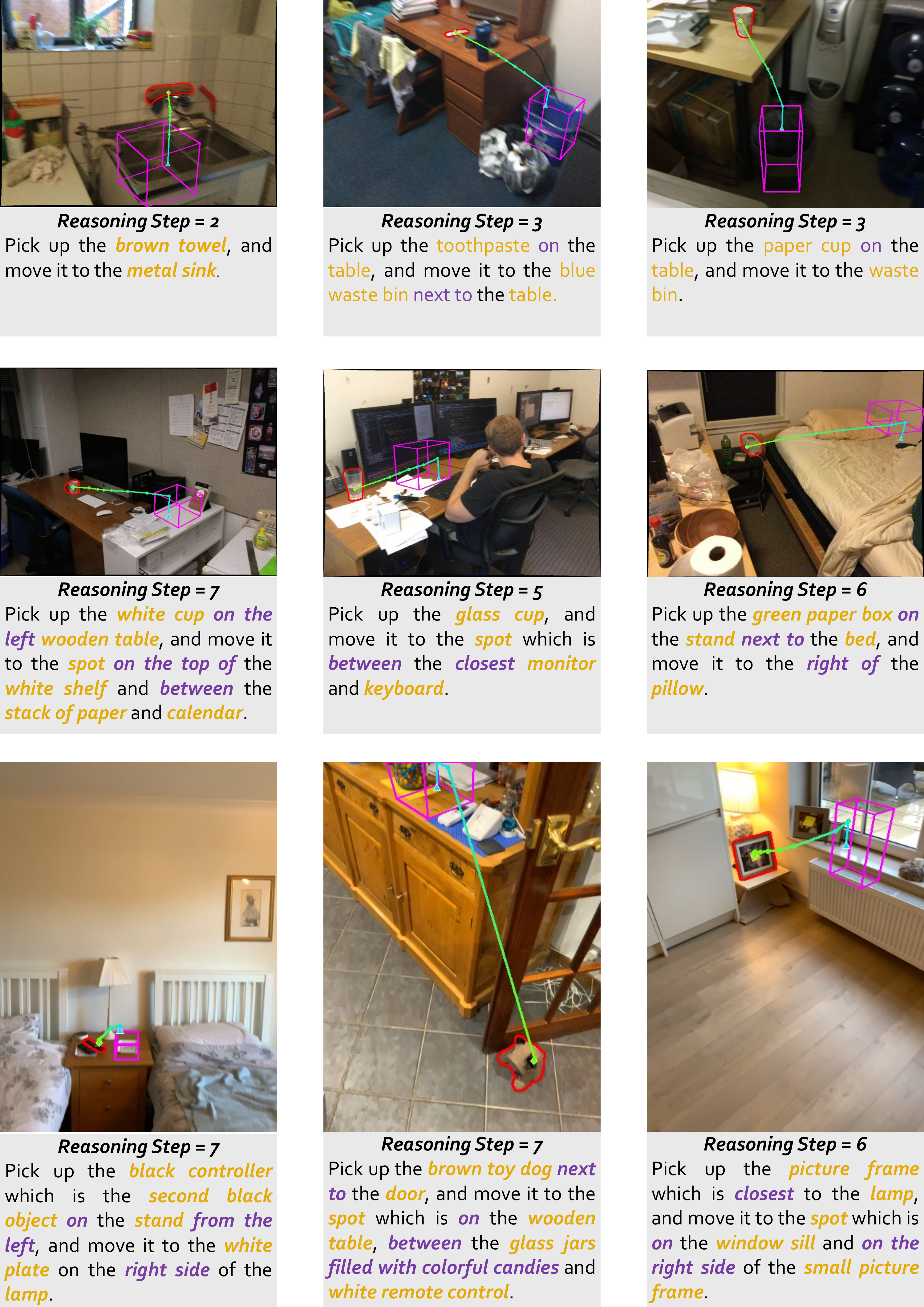}
    
    \caption{Visualization of TraceSpatial-Bench and {\mname}'s rollouts. The red mask indicates the ground-truth starting point, and the purple 3D bounding box denotes the ground-truth endpoint. We show the 2D projection of {\mname}'s predicted trace.}
    
    \label{suppfig: TraceSpatial-Bench_vis3}
\end{figure*}

\textbf{Visualization of {\bname}.}
We present examples of {\bname} and our model's rollouts in Fig.~\ref{suppfig: TraceSpatial-Bench_vis1}, \ref{suppfig: TraceSpatial-Bench_vis2}, \ref{suppfig: TraceSpatial-Bench_vis3}.

\begin{figure*}[h]
    \centering
    
    \includegraphics[width=0.8\linewidth]{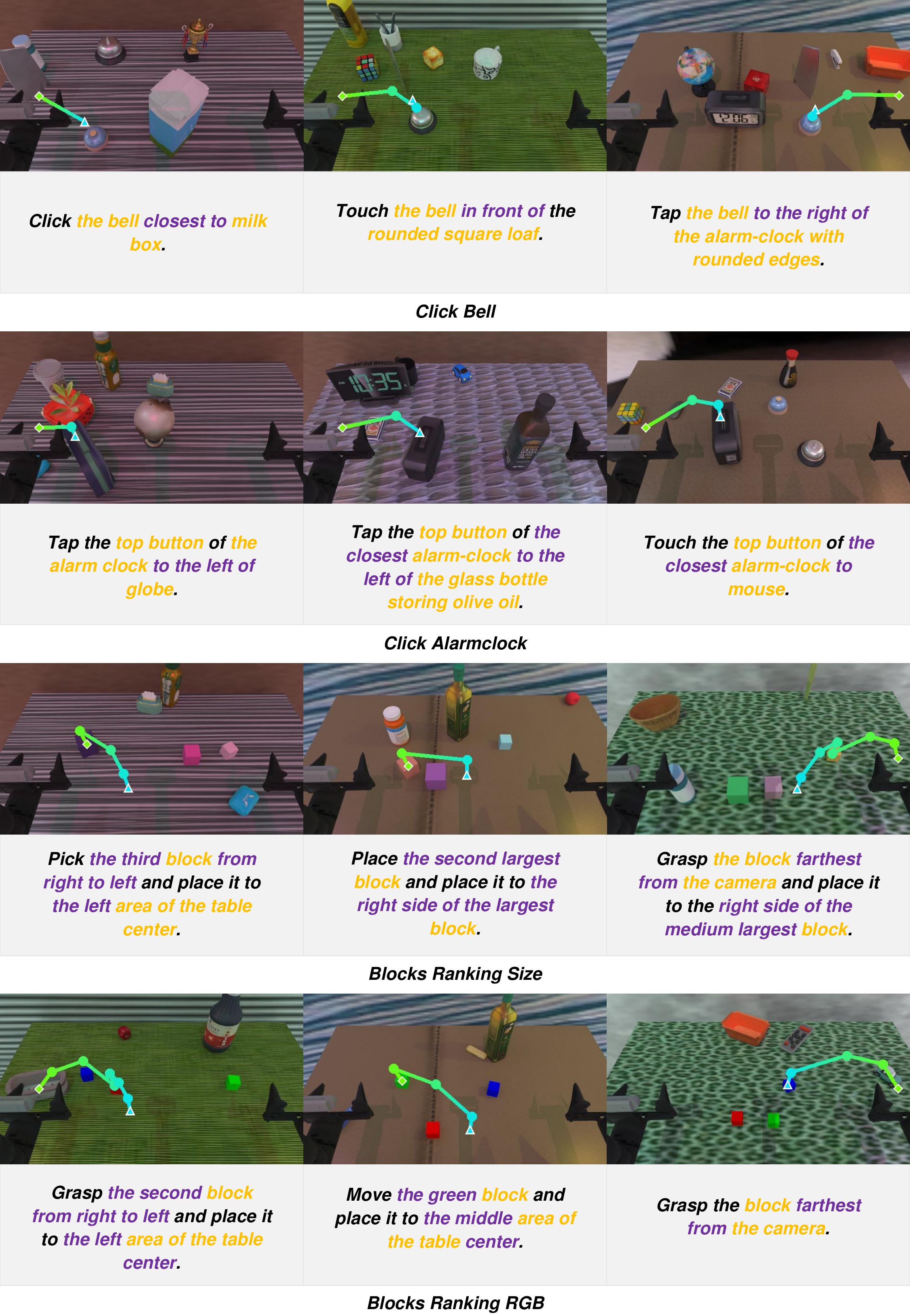}
    
    \caption{RoboTwin Data Visualization}
    
    \label{suppfig: robotwin_1}
\end{figure*}

\begin{figure*}[h]
    \centering
    
    \includegraphics[width=0.8\linewidth]{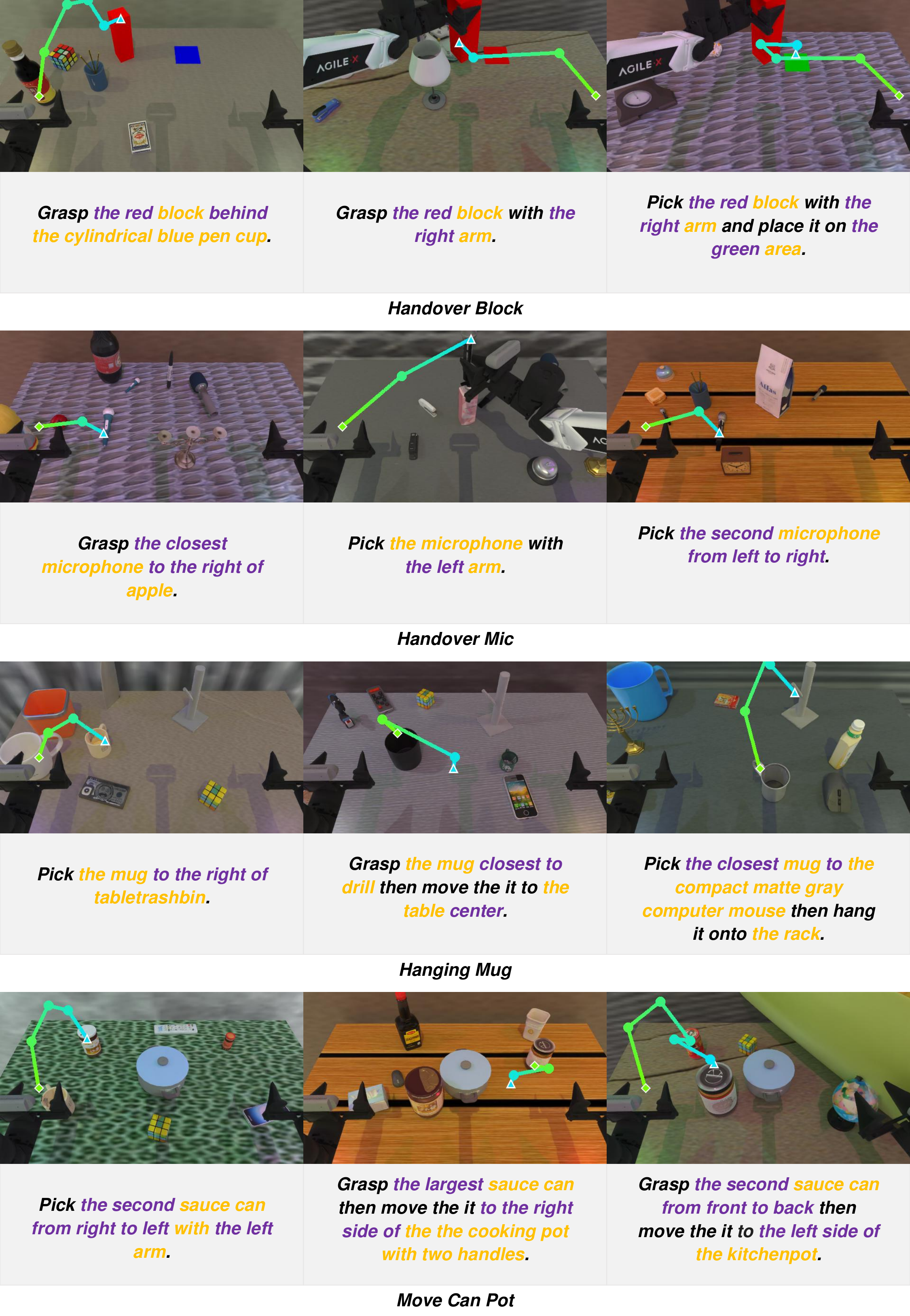}
    
    \caption{RoboTwin Data Visualization}
    
    \label{suppfig: robotwin_2}
\end{figure*}

\begin{figure*}[h]
    \centering
    
    \includegraphics[width=0.8\linewidth]{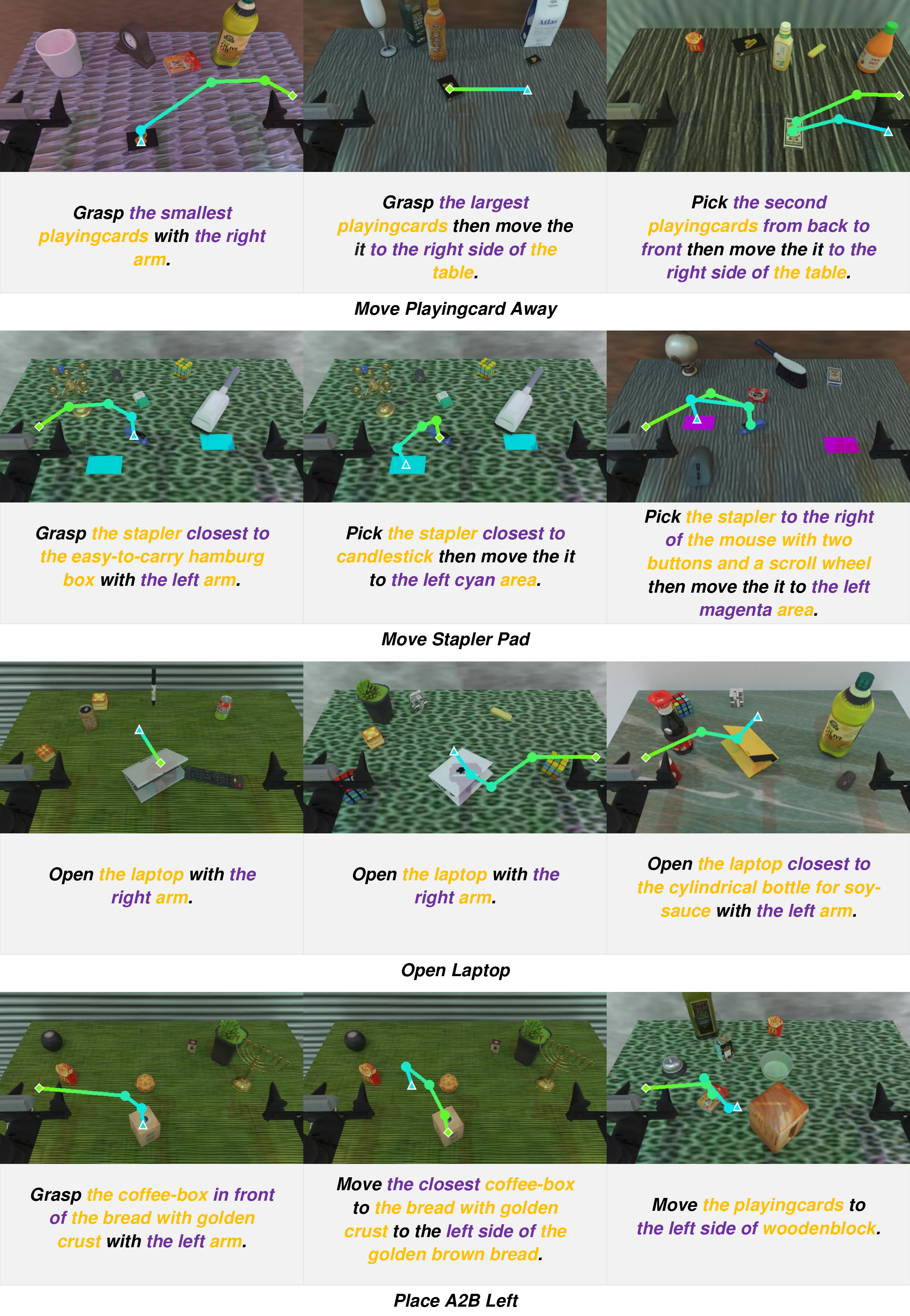}
    
    \caption{RoboTwin Data Visualization}
    
    \label{suppfig: robotwin_3}
\end{figure*}

\begin{figure*}[h]
    \centering
    
    \includegraphics[width=0.8\linewidth]{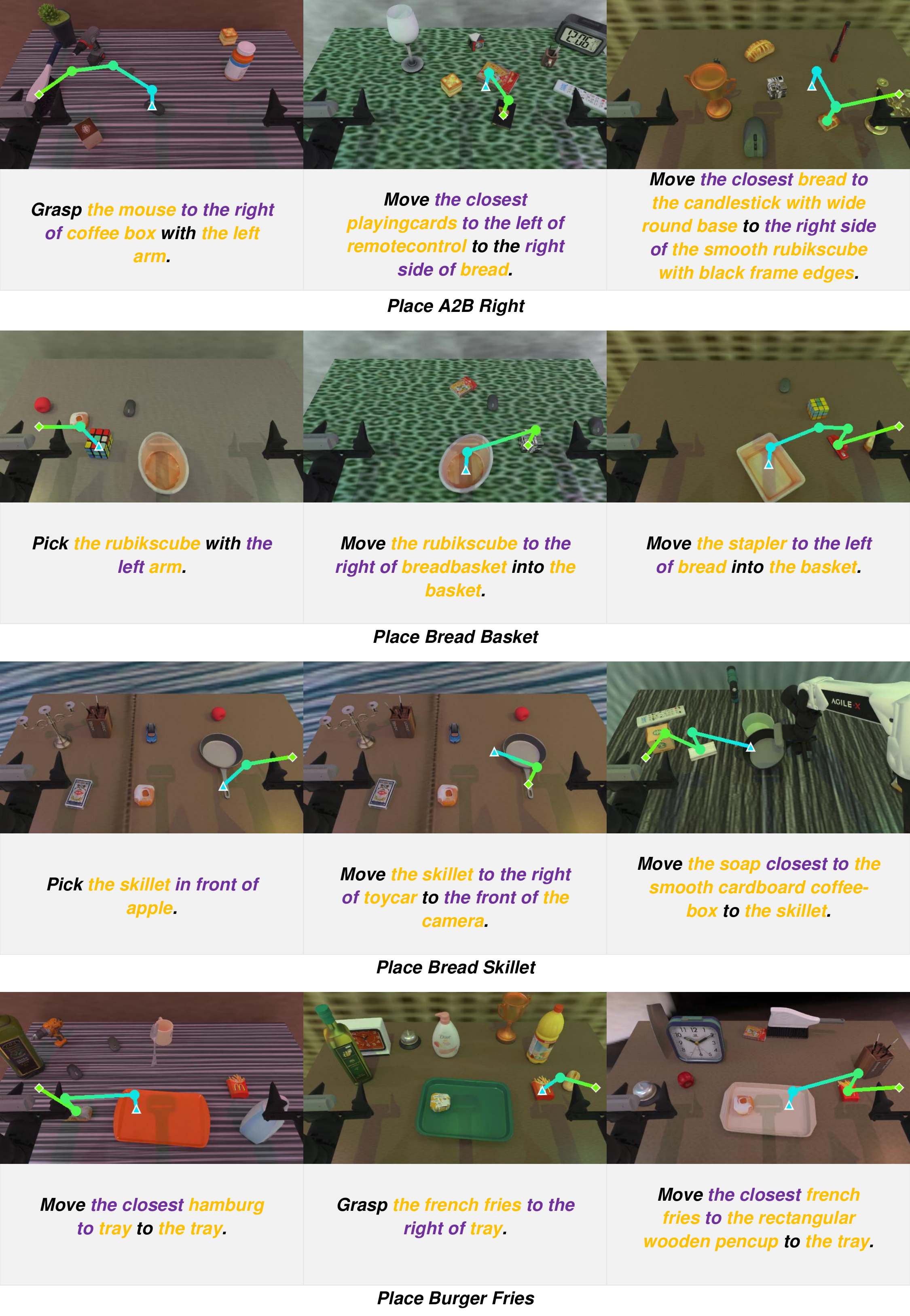}
    
    \caption{RoboTwin Data Visualization}
    
    \label{suppfig: robotwin_4}
\end{figure*}

\vspace{+1mm}
\noindent \textbf{Visualization of Simulation Data for Spatial Tracing.}
We present data examples in Fig.~\ref{suppfig: robotwin_1}, \ref{suppfig: robotwin_2}, \ref{suppfig: robotwin_3}, \ref{suppfig: robotwin_4}.



\begin{figure*}
\centering
\vspace{-7mm}
\includegraphics[width=0.8\linewidth]{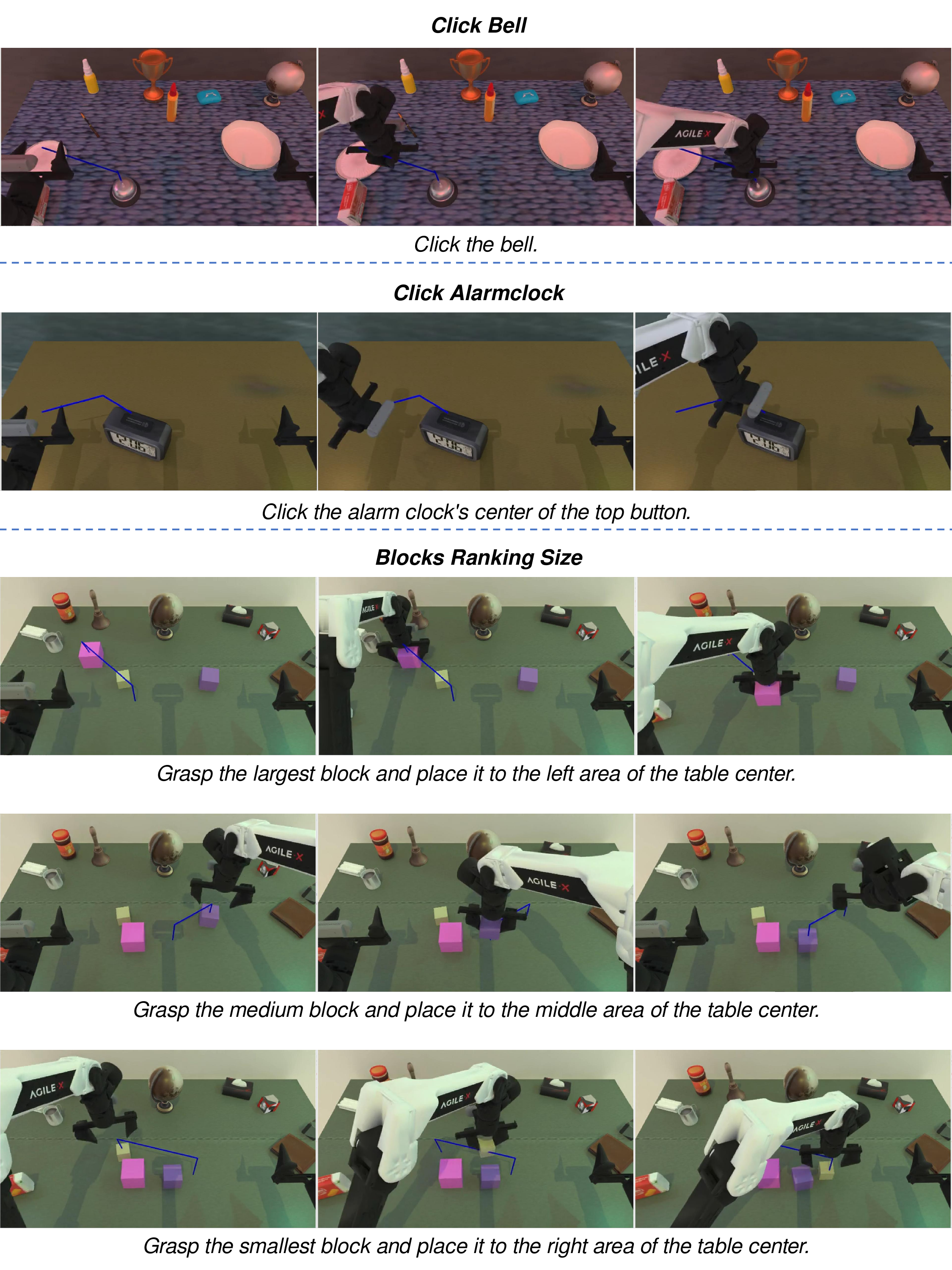}
   \caption{Visualized RoboTwin simulation evaluation process. }
\label{suppfig: robotwin_eval1}
\end{figure*}

\begin{figure*}
\centering
\vspace{-7mm}
\includegraphics[width=0.8\linewidth]{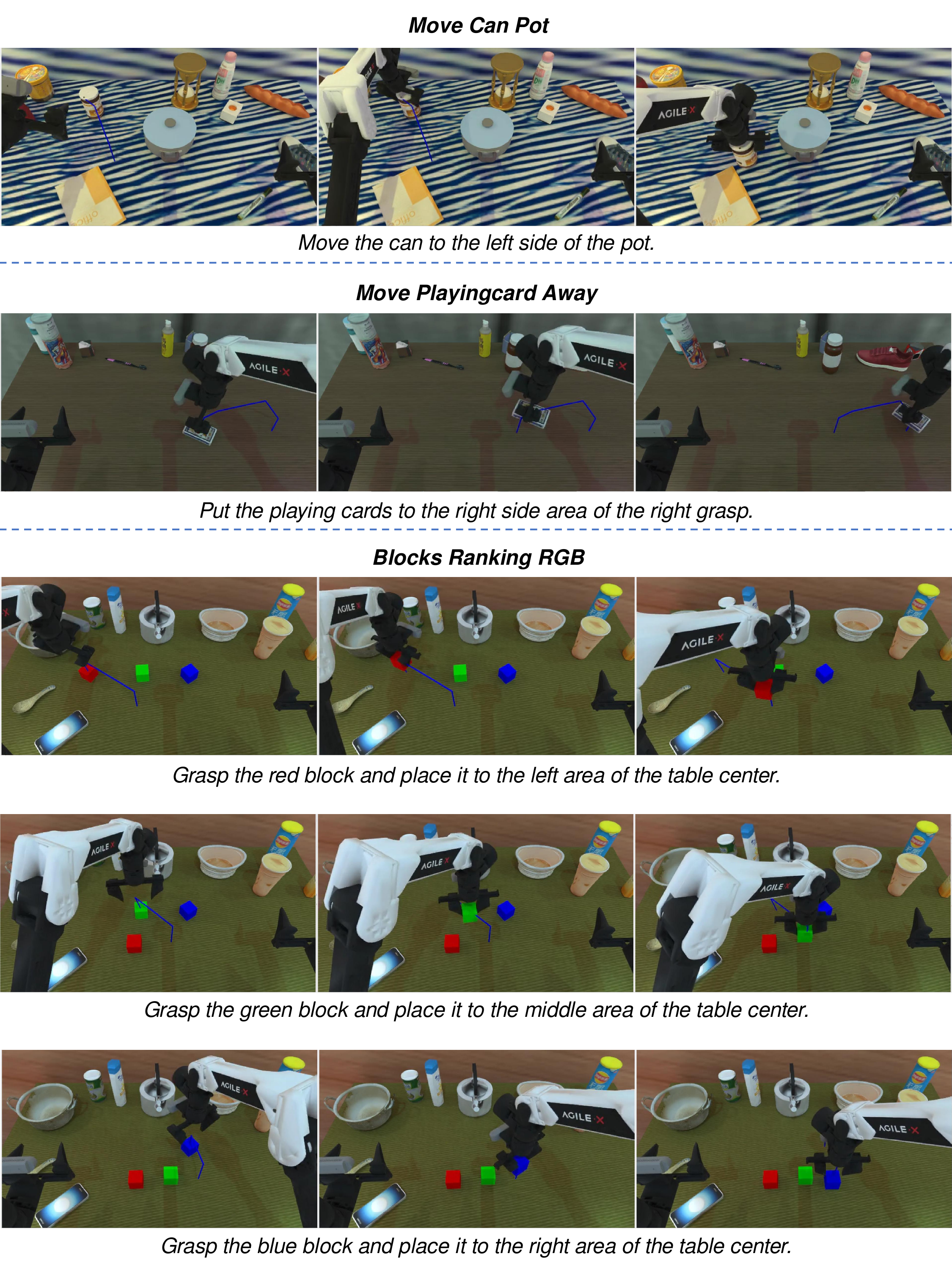}
   \caption{Visualized RoboTwin simulation evaluation process. }
\label{suppfig: robotwin_eval2}
\end{figure*}

\begin{figure*}
\centering
\vspace{-7mm}
\includegraphics[width=0.8\linewidth]{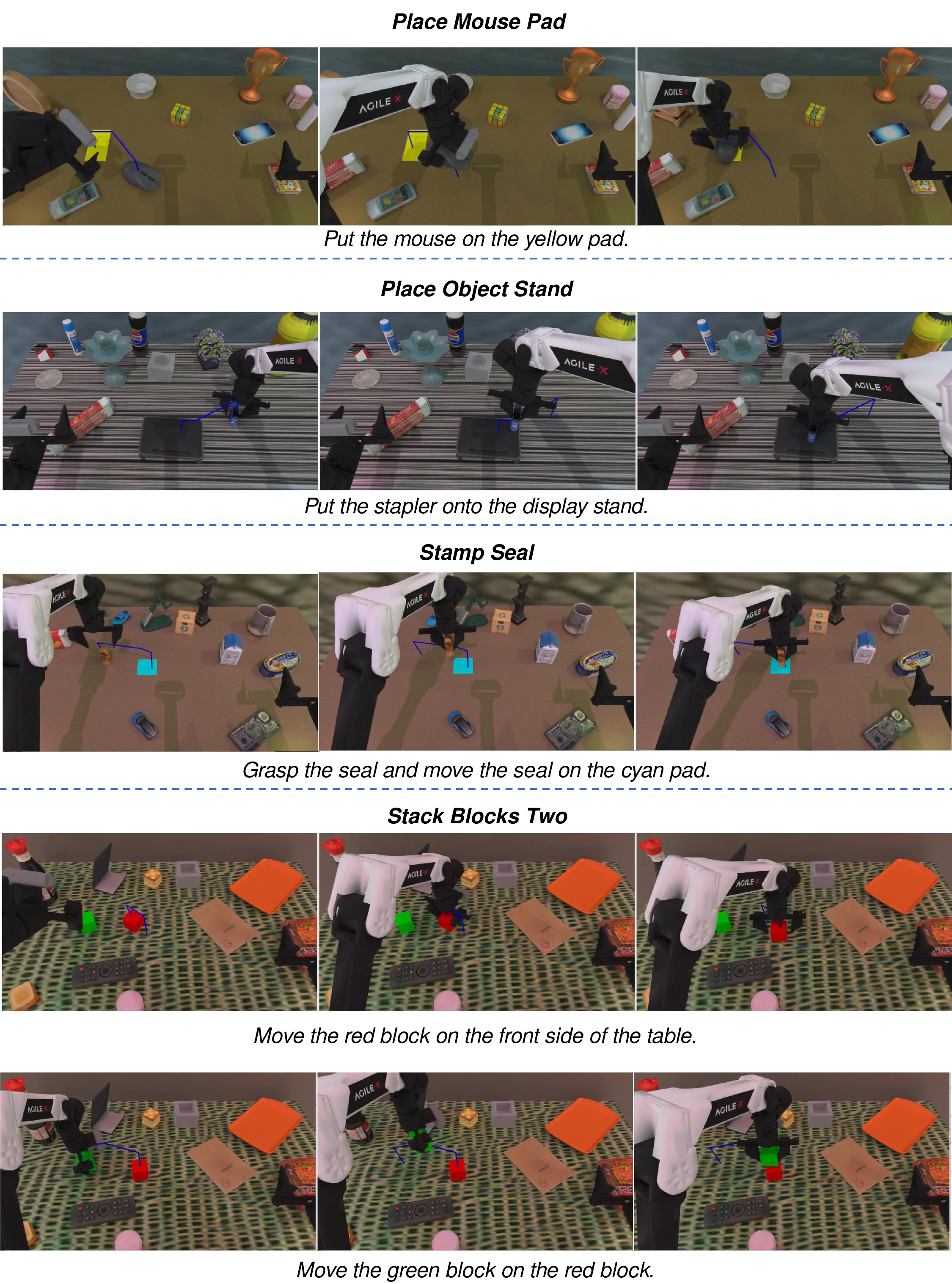}
   \caption{Visualized RoboTwin simulation evaluation process. }
\label{suppfig: robotwin_eval3}
\end{figure*}

\begin{figure*}
\centering
\vspace{-7mm}
\includegraphics[width=0.8\linewidth]{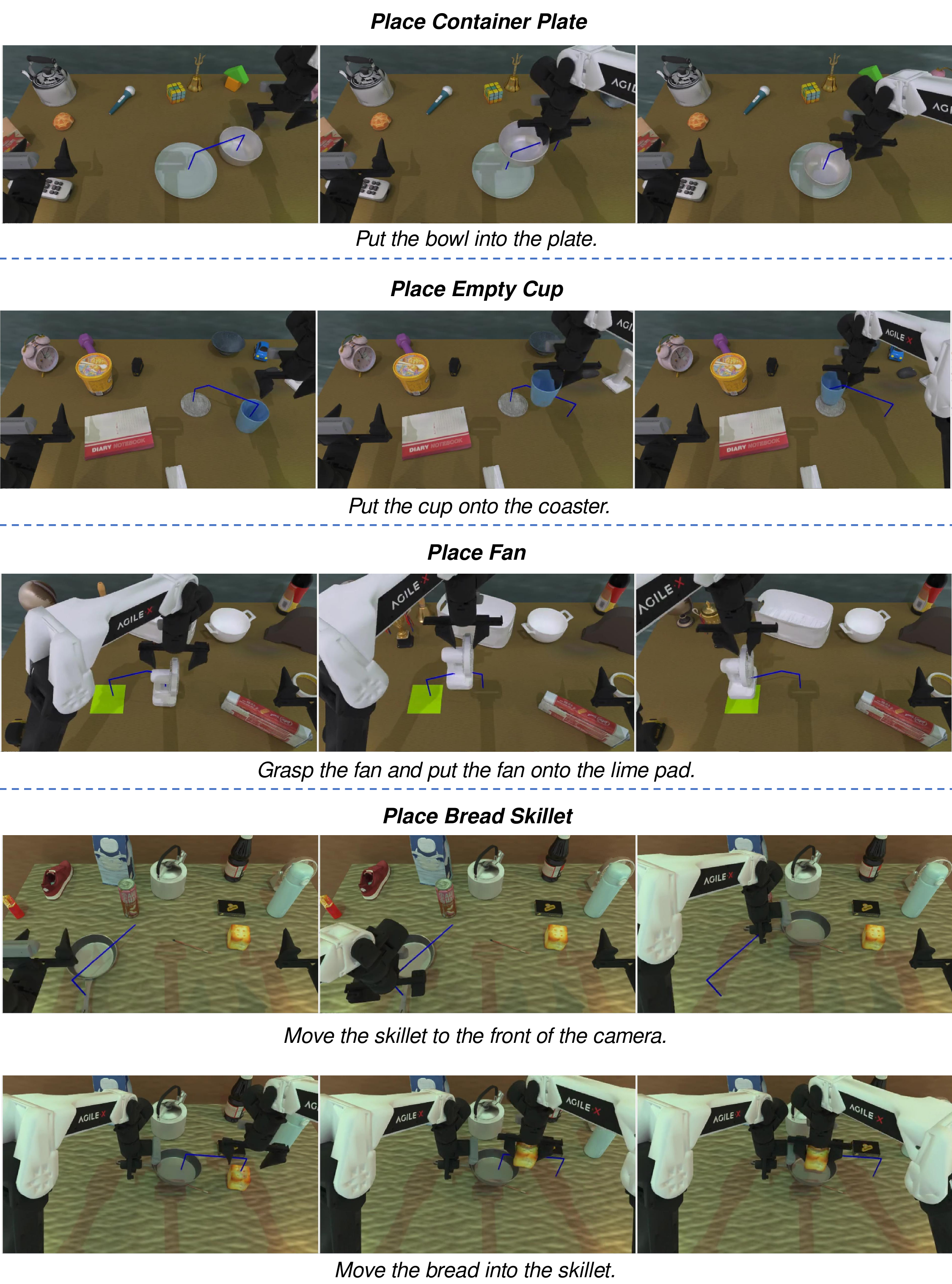}
   \caption{Visualized RoboTwin simulation evaluation process. }
\label{suppfig: robotwin_eval4}
\end{figure*}

\begin{figure*}
\centering
\vspace{-7mm}
\includegraphics[width=0.8\linewidth]{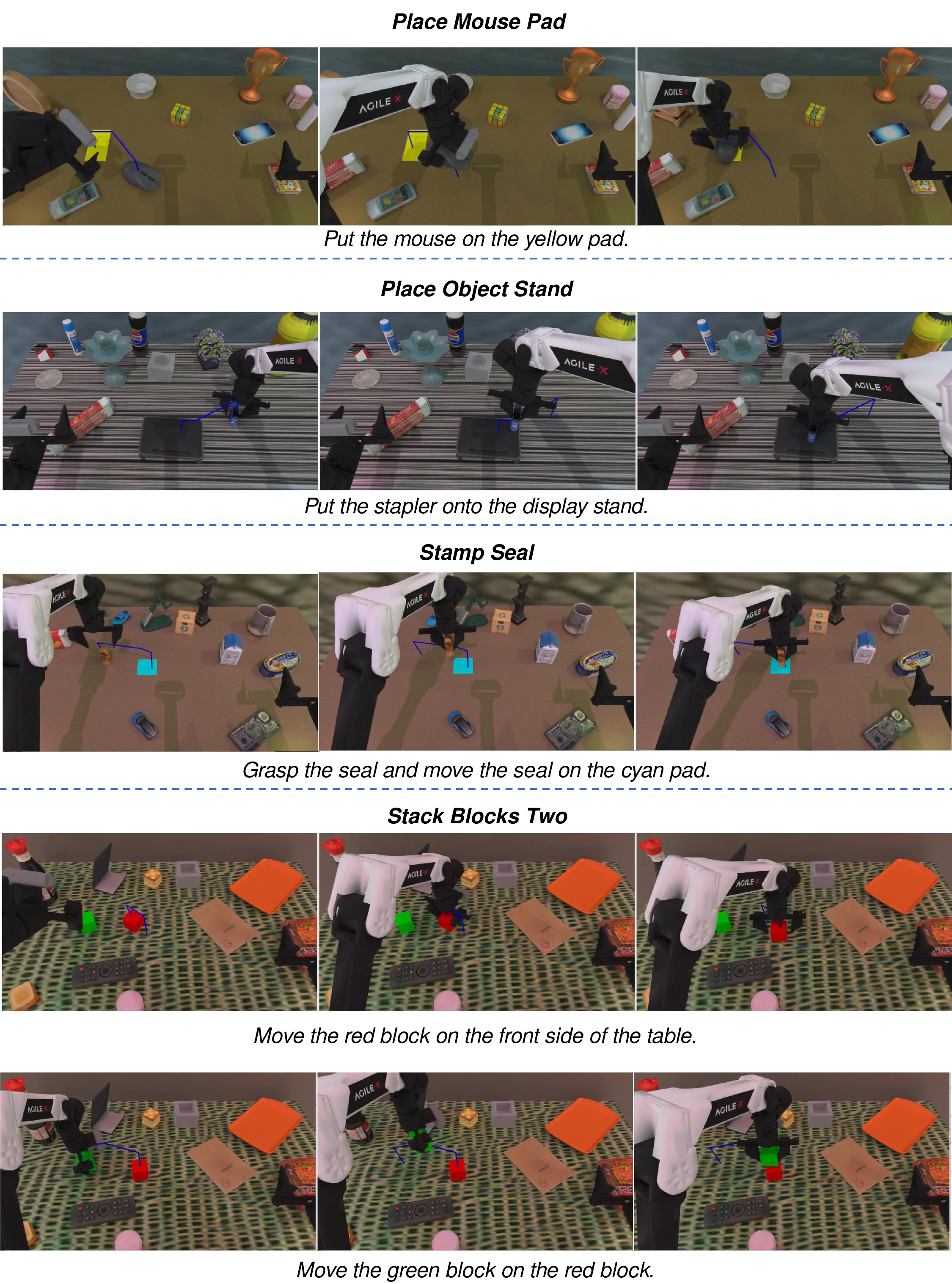}
   \caption{Visualized RoboTwin simulation evaluation process. }
\label{suppfig: robotwin_eval5}
\end{figure*}

\begin{figure*}
\centering
\vspace{-7mm}
\includegraphics[width=0.7\linewidth]{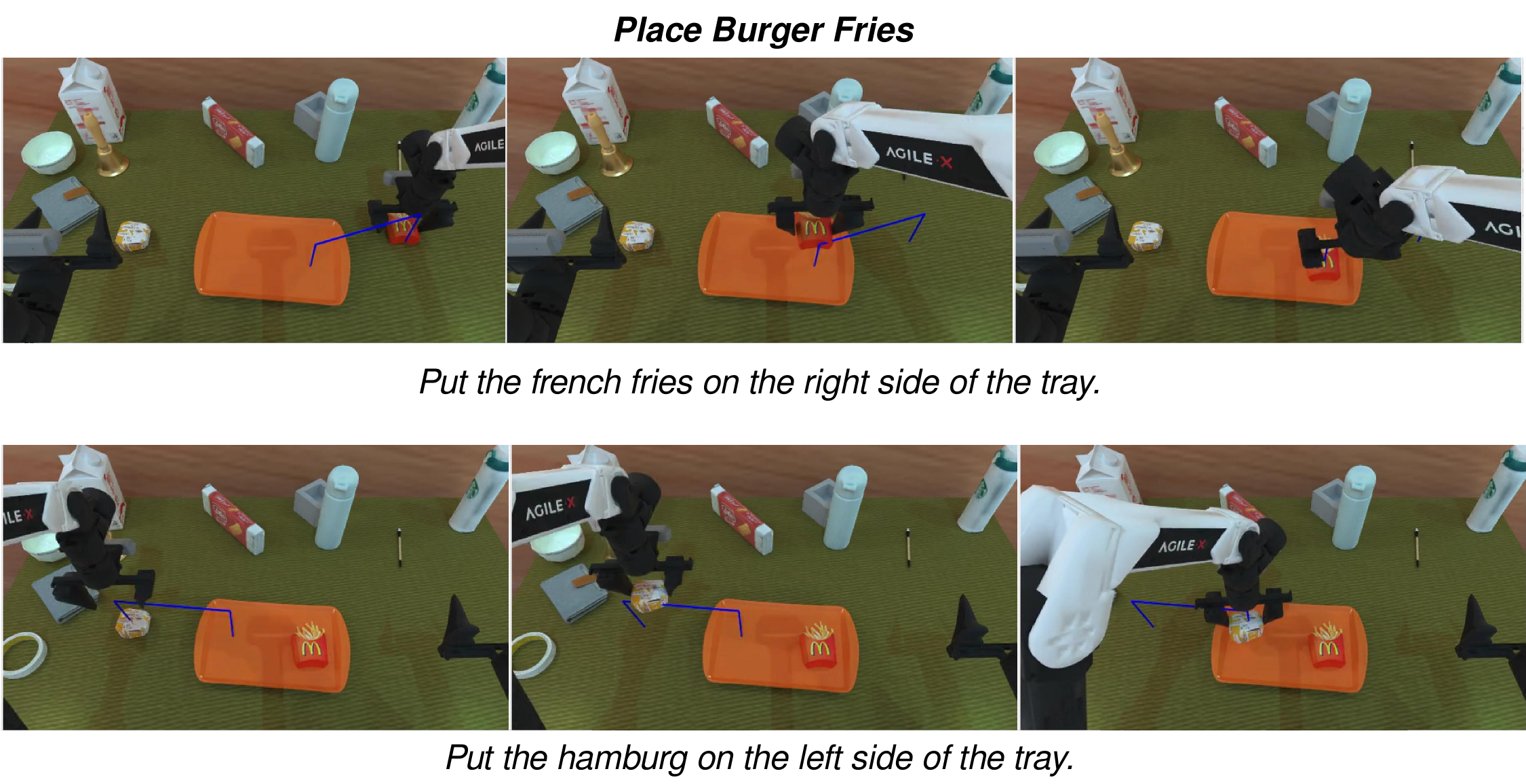}
   \caption{Visualized RoboTwin simulation evaluation process. }
\label{suppfig: robotwin_eval6}
\end{figure*}

\vspace{+1mm}
\noindent \textbf{Visualization of Simulator.}
We present example rollouts with {\mname} predictions in Fig.~\ref{suppfig: robotwin_eval1}, \ref{suppfig: robotwin_eval2}, \ref{suppfig: robotwin_eval3}, \ref{suppfig: robotwin_eval4}, \ref{suppfig: robotwin_eval5}, \ref{suppfig: robotwin_eval6}.


\section{Discussion on Limitations and Future Work}
\label{suppsec: limitation}

Despite achieving promising results, our model still has limitations.
First, the spatial tracing task requires the model to fully understand 3D space and interpret various spatial constraints from instructions. 
This demands strong capabilities in perception, reasoning, and generalization across tasks, scenes, and objects. 
As a result, a VLM-based architecture is essential.
However, VLMs often suffer from slow inference, making them less suitable for highly dynamic environments. 
Acceleration techniques such as quantization or parallel inference may help mitigate this limitation.
Second, spatial tracing requires predicting 3D positional sequences with real-world scale information (\eg, depth), which is particularly challenging for VLM-based architectures that are primarily pre-trained on 2D data.
While {\mname} makes significant efforts, such as collecting metric-grounded data during SFT, aligning the new inputs via a universal spatial encoder, supervising scale factor prediction through regression loss, using metric-sensitive function to generate rewards, \highlight{the resulting improvements in 3D metric-grounded spatial understanding, especially driven by scaling up data, are limited compared to the gains observed in 2D spatial reasoning tasks} (\eg, RoboRefer~\cite{zhou2025roborefer}).
Recent work like DepthLM~\cite{cai2025depthlm} further highlights this limitation. 
Despite processing over 30M+ images with unified camera intrinsics, DepthLM sacrifices language understanding to focus solely on depth estimation at the pixel level, achieving performance only comparable to task-specific expert models.
These results indicate that enhancing the 3D metric-grounded capabilities of VLMs, while maintaining their language understanding for reasoning, remains an open and challenging problem.
We argue that scaling up data alone yields limited benefits due to two key reasons:
\textbf{(1)} \highlight{The lack of a scene-level 3D representation that is naturally aligned with language and rich enough to encode fine-grained geometry.} Although this work attempts alignment in the initial SFT stage, the spatial encoder’s features remain poorly aligned with the language space, and supervision via next-token prediction alone may be insufficient.
\textbf{(2)} \highlight{The lack of comprehensive, scale-aware supervision signals.} This work only supervises a simple metric scale factor, but dense geometric outputs (\eg, depth maps), which are crucial for full 3D understanding, are not utilized due to computational overhead.
Overall, enabling VLMs to fully understand 3D space requires more than large-scale data.
Therefore, better input representations and richer supervision signals, especially at both the input and output levels, are a promising direction for future research to further advance spatial intelligence.

\newpage








%
%
\clearpage
\bibliographystyle{splncs04}
\bibliography{main}
\end{document}